\documentclass[11pt, a4paper]{custom_class}

\usepackage{custom_style}
\usepackage[utf8]{inputenc} 
\usepackage[T1]{fontenc}    
\usepackage{hyperref}       
\usepackage{url}            
\usepackage{booktabs}       
\usepackage{amsfonts}       
\usepackage{nicefrac}       
\usepackage{microtype}      
\usepackage{cleveref}       
\usepackage{lipsum}         
\usepackage{graphicx}
\usepackage{doi}
\usepackage{glossaries}
\usepackage{natbib}
\title{A Philosophical Introduction to Language Models \\[1ex] \subtitlefont\sc Part II: The Way Forward}

\author{
Raphaël Millière \\
Department of Philosophy \\
Macquarie University \\
\texttt{raphael.milliere@mq.edu.eu} \\
\And
Cameron Buckner \\
Philosophy Department \\
University of Houston \\
\texttt{cjbuckner@uh.edu} \\
}

\renewcommand{\headeright}{Part II}
\renewcommand{\undertitle}{}
\renewcommand{\shorttitle}{A Philosophical Introduction to Language Models}

\definecolor{glossarycolor}{RGB}{107, 0, 0}

\date{} 

\makeglossaries

\newglossaryentry{nonparametric}{
  name={nonparametric},
  description={In the context of experiments on LLMs, nonparametric methods are used for analyzing or extracting information from the trained model without the need to train additional parameters. Unlike traditional probing techniques that involve training a separate classifier (probe) to interpret the model's internal activations, nonparametric approaches seek to infer linguistic or other properties directly from the existing components of the trained model. These methods often focus on pairwise importance scores, such as attention weights or distances between token representations, to derived insights about the internal structure of the model's representations and computations}
}

\newglossaryentry{circuit}{
  name={circuit},
  description={In a trained neural network, a circuit refers to a specific configuration of interconnected model weights that collectively perform a meaningful computation or series of computations. In practice, a circuit can be understood as a chain of operations that detects certain features in the input data encoded in one layer of the network and transforms them into different -- often more complex -- features encoded in a further layer. For instance, in computer vision models, a circuit might consist of edge detectors, texture detectors, or shape recognizers, each specifically tuned to identify and process certain visual aspects of the input. In the case of language models, circuits might be specialized in recognizing specific grammatical structures, semantic relationships, or patterns in text sequences. The identification and analysis of circuits are crucial for mechanistic interpretability, as they provide a modular view of how neural networks transform inputs into outputs}
}

\newglossaryentry{gracefuldegradation}{
  name={graceful degradation},
  description={The capacity of neural networks to maintain functionality despite the impairment or loss of some components. This concept emerged from the study of how neural networks respond to interventions such as ablations, where nodes within the network are selectively disabled. Unlike abrupt failure, graceful degradation is characterized by a gradual decline in performance, illustrating the network's robustness and resilience to partial damage. This phenomenon is akin to how biological brains, when faced with injury or loss of neurons, often continue to function, albeit at a reduced efficiency. The discovery of graceful degradation in neural networks underscores their parallel with biological systems and highlights a fundamental property of distributed processing systems, where information processing and network functionality are not localized to a single node but are distributed across the network}
}

\newglossaryentry{ablation}{
  name={ablation},
  description={Ablation refers to the process of removing or disabling parts of a neural network in order to study the impact on the network's performance. It allows researchers to determine which components of a network contribute most to its functionality. As a causal intervention method, it is somewhat rudimentary and has largely been superseded by more sophisticated techniques like interchange interventions}
}

\newglossaryentry{nullspace}{
  name={nullspace},
  description={In the context of iterative nullspace projection, the nullspace of a probe contains representation directions that are not useful for that probe to make its prediction. Projecting onto the nullspace removes information the probe uses, but maintains encoding in the orthogonal directions}
}

\newglossaryentry{keyqueryvalue}{
  name={key-query-value},
  description={In the attention mechanism of a Transformer, each input word embedding is projected into three different vectors - a query vector ($Q$), a key vector ($K$), and a value vector ($V$). This is done using three separate trainable weight matrices ($W_q$, $W_k$, $W_v$) for each attention head. The purpose of these three vectors is to enable the Transformer to selectively focus on or `attend to' the most relevant parts of the input sequence when a given attention head is processing each token. At a given attention head, each token's query vector encodes the information that should be queried from other tokens, while key vectors encodes the information about each token that is relevant to the query. To determine which tokens are most relevant to each token in the input, the Transformer takes the dot product of each query vector with all the key vectors. This produces an `attention score' for every token indicating how closely it matches what the query is looking for at that particular attention head. Finally, the value vector encodes which information should be retrieved from each token, which is then weighted by the attention score. Through this mechanism, each attention head processing a given input token probes all the other tokens for relevant information (using $Q$ and $K$ vectors), and then construct a new embedding (using the $V$ vectors) that integrates that information based on relevance. This allows attention heads to selectively capture different relationships and dependencies between words, even across long distances}
}

\newglossaryentry{bigram}{
  name={bigram},
  description={A bigram is a sequence of two adjacent elements from a string, such as two consecutive words or tokens. Bigram models predict the next token based on the previous token, essentially memorizing common pairs like "United States" or "New York". In contrast, induction heads do not merely memorize bigram statistics, but implement instead a general algorithm: find previous occurrences of the current token, attend to the token that came next, and predict that same token is likely to come next again (\texttt{[A][B] ... [A] → [B]}). This computation is \textit{not} content-specific; it does not depend on the specific identities of tokens \texttt{[A]} and \texttt{[B]}. It works for arbitrary token pairs, not just memorized common bigrams. In fact, induction heads can complete repeated sequences of random tokens whose bigram statistics the model could not have possibly memorized during training}
}

\newglossaryentry{GAN}{
  name={GAN},
  description={A Generative Adversarial Network (GAN) is a machine learning architecture that consists of two networks – the \textit{generator} and the \textit{discriminator} – competing against each other to generate new, synthetic data that closely resembles real data. The generator's goal it so produce fake data (e.g., fake images) that looks as close to real data as possible, while the discriminator's goal is to distinguish between the real data and the fake data created by the generator. The generator and discriminator are trained simultaneously. The generator learns to create increasingly realistic fake data to fool the discriminator, while the discriminator continuously learns to better identify the differences between real and fake data. As the training progresses, the generator becomes better at creating realistic data, and the discriminator becomes more skilled at detecting fake data. The result of this competitive process is a generator that can create new, synthetic data that is very similar to real data. GANs have been used in various applications, such as generating realistic images}
}

\newglossaryentry{probing}{
  name={probing},
  description={Probing is a method to analyze and interpret the information encoded in artificial neural networks such as LLMs. The basic idea is to train a classifier, called a `probing classifier', to predict some property from the model's internal representations, such as part-of-speech tags or syntactic dependencies. More formally, let $f$ be the target neural network model that maps input text $x$ to some output $y$. This model generates intermediate representations $f_l(x)$ at some layer $l$. The probing classifier $g$ is a separate model (typically, a linear classifier) that maps these intermediate representations $f_l(x)$ to linguistic properties $z$. The probing classifier $g$ is trained on a labeled dataset of $\{x, z\}$ pairs to predict property $z$ from representation $f_l(x)$. The key assumption is that if the probing classifier $g$ achieves high accuracy in predicting property $z$, then the representations $f_l(x)$ must encode useful information about that linguistic property. Thus, probing aims to shed light on what kind of information is encoded in the representations learned by the target model $f$ at different layers. There are several limitations and open questions in the probing paradigm, which includes issues related to the choice of probing classifier architecture, the selection of proper control tasks and baselines for comparison, disentangling the influence of the training dataset vs the model architecture. Crucially, probing is a correlational rather than causal method; it can reveal if some information is encoded in a network, but it cannot by itself reveal whether the network actually makes use of that information for behavior. Causal claims can be established on firmer ground by complementing probing with intervational methods that modify the information encoded by the network and assess downstream effects of such interventions on model behavior}
}

\newglossaryentry{attribution}{
  name={attribution},
  description={Attribution methods refer to a set of techniques used in deep learning to determine which parts of the input data a model relies on most when making predictions or decisions. The goal is to explain what aspects of the input have the greatest influence on a trained model's outputs. Attribution methods originated in computer vision, where they have been used to highlight the portions of an image that a neural network considers most important for its predictions. For example, when classifying an image as containing a dog, an attribution method could identify that the pixels representing the dog's face were the key factor driving the classification. By visualizing these important input features, attribution methods aim to provide insight into the model's classification process. More recently, the notion of attribution has been applied to other domains such as NLP. In NLP, attribution methods can highlight the words or phrases in a text input that have the biggest impact on a language model's predictions. This allows for better interpretability of language models and can reveal whether they are basing their outputs on relevant words or irrelevant/spurious correlations. Importantly, attribution methods only identify which input features are most important for a model's predictions, but do not explain how those features are internally encoded by the model. Other interpretability techniques such as \gls{probing} and \gls{causal-intervention} methods can provide deeper insight into the model's learned representations and computations}
}

\newglossaryentry{causal-intervention}{
  name={causal intervention},
  description={Traditional methods for interpreting deep learning models, such as \gls{probing}, can only establish correlational relationships between a model's internal representations and properties of interest. Such correlational methods risk yielding `false positives' where a probe detects information that is not causally relevant to the model's behavior. In contrast, causal intervention methods aim to establish the causal role that a representation plays in a model's computation. The key idea is to directly manipulate or intervene on a model's representations and observe the effect on the model's output. If intervening to modify information about a property in the representation changes the model's predictions in a systematic way, this provides evidence that the model was relying on that information to make predictions. One popular approach known as \textit{interchange intervention} works in the following way: given two inputs, swap the model's intermediate representations for those inputs and observe whether this changes the model's final output in an interpretable way. This allows researchers to test whether the model's representations encode the kind of modular, compositional structure that would be expected if the model is performing a systematic computation to solve the task}
}

\newglossaryentry{addressable-memory}{
  name={addressable memory},
  description={An addressable memory is a critical component of a classical computer architecture that enables a computing system to store and retrieve information. It consists of a set of memory locations, each of which has a unique address. The address acts as a `handle' that allows the computational machinery to access the contents of a specific memory location directly, without having to search through the entire memory. Adressable memory enables variable binding – the process of associating a specific value with a variable, so that the value can be retrieved and used in computations by referring to the variable. In an addressable memory, the address of a memory location serves as a variable, while the contents at that memory location represent the value bound to that variable. By using the address as a symbol for the variable, the computational machinery can retrieve the value associated with the variable (stored at that memory address), and can also update the value by writing a new value to the same memory location. This architecture allows for complex data structures and computations. Notably, it enables indirect addressing, where the value stored at a memory location is itself an address pointing to another memory location. Unlike classical computer architectures, DNNs like Transformer-based LLMs do not have explicit addresses and memory locations. Yet interpretability research suggests that they develop an analogous mechanism through the \gls{residual-stream}. In a Transformer, the residual stream is the main vector that flows through the network at each token position, being iteratively updated by each layer. The residual stream is a high-dimensional vector space, and different components of the model, such as attention heads, can use different subspaces of this vector space to store and communicate information. An attention head reading and writing to a particular subspace of the residual stream is analogous to reading and writing to a particular memory address, even though distinctions between addressable subspaces are learned rather than hardcoded in the architecture}
}

\newglossaryentry{residual-stream}{
  name={residual stream},
  description={The residual stream is the central pathway through which information flows in a Transformer model. It is a vector at each token position that is iteratively updated by each layer as it passes through the model. Specifically, the residual stream starts as the sum of the token embeddings and positional embeddings. This initial vector representation is then fed into the first layer. Each layer reads from the residual stream, applies attention heads and MLP blocks to update the information, and then writes its output back into the residual stream by adding it to the previous values. This process repeats at each layer, with the residual stream accumulating information at each step. The final value of the residual stream is what gets mapped to output logits to predict the next token. Importantly, the residual stream provides a direct pathway for information to travel from any layer to any later layer, and attention heads can also route information dynamtically between residual streams across distinct token positions. The fact that attention heads and MLP layers can write and read information in subspaces of residual streams suggests that they effectively function like an \gls{addressable-memory}, which enables sophisticated mechanisms like induction head circuits}
}

\newglossaryentry{grokking}{
  name={grokking},
  description={Grokking is a phenomenon in neural networks where, after a period of overfitting the training data and performing poorly on held-out test data, the model abruptly transitions to generalizing well, with test performance rapidly improving to match the training performance. This sudden onset of generalization, which occurs after the model has already overfit, distinguishes grokking from typical learning curves. Grokking relies on the model being regularized, typically through weight decay (a technique where the model weights are gradually shrunk towards zero, favoring simpler solutions), and trained on a limited dataset; without these conditions, the rapid shift to generalization does not occur. The term `grokking' evokes the idea of the network suddenly `getting it' after a period of stagnation. Mechanistically, grokking is hypothesized to occur when the regularization pressure eventually causes the model to shift from a brittle, overfitted solution to a more parsimonious, generalizable one after the memorization becomes too costly. Grokking highlights the complex, non-monotonic generalization behavior that can arise in deep learning}
}

\begin{document}
\maketitle

\begin{abstract}
In this paper, the second of two companion pieces, we explore novel philosophical questions raised by recent progress in large language models (LLMs) that go beyond the classical debates covered in the first part. We focus particularly on issues related to interpretability, examining evidence from causal intervention methods about the nature of LLMs' internal representations and computations. We also discuss the implications of multimodal and modular extensions of LLMs, recent debates about whether such systems may meet minimal criteria for consciousness, and concerns about secrecy and reproducibility in LLM research. Finally, we discuss whether LLM-like systems may be relevant to modeling aspects of human cognition, if their architectural characteristics and learning scenario are adequately constrained.
\end{abstract}

\section{Introduction}

The maturation of connectionist models in the form of deep neural networks not only revives longstanding philosophical issues but also introduces new ones that await exploration \citep{bucknerDeepLearningPhilosophical2019}. This development notably includes the progress of large language models (LLMs) like GPT-4 \citep{openaiGPT4TechnicalReport2023}, whose impressive performance on complex linguistic and cognitive tasks raises philosophically rich questions. In a companion paper, we discussed the significance of LLMs in relation to classical problems from the philosophy of artificial intelligence, cognitive science, and linguistics \citep{millierePhilosophicalIntroductionLanguage2024}. In this paper, we turn to relatively new issues raised by the progress of LLMs. We focus in particular on three sets of questions. In section \ref{mechanistic-understanding}, we ask whether and how we can gain an understanding of the internal mechanisms of LLMs beyond their behavioral performance. In section \ref{sec:newer-philosophical-questions}, we discuss philosophical questions about achievements that may be just over the horizon of current research: focusing on multimodality, agency, consciousness, and reproducibility. Finally, in section \ref{sec:cognitive-models}, we bring these strands together by interrogating the status of LLMs as models of human and animal cognition. 

\section{Mechanistic understanding and intervention methods} \label{mechanistic-understanding}

As noted in Part I, LLMs have made rapid progress on producing human-like linguistic behavior in many different scenarios that were challenging for previous methods in artificial intelligence. They can readily produce grammatically-correct and generally semantically-coherent text. They can respond flexibly to a wide range of questions, including a variety of challenging aptitude and proficiency tests. They can even generate functional blocks of computer code in various programming languages, as well as code in visual markup languages that produce coherent images. As we also noted, however, there remain significant concerns that these achievements may have less profound explanations than they first appear. While we have argued that skeptical interpretations appealing to mere memorization cannot account for the full range of behaviors exhibited by these models, our previous analysis did not provide a positive account of the mechanisms that might enable LLMs to achieve such breakthrough performance. Providing such an account is challenging due to familiar concerns about the opacity of neural networks, which is exacerbated by LLMs' relatively novel architecture, their enormous number of adjustable parameters, and the sheer magnitude of their training data. To address these challenges, the research community is rapidly developing new methods of analysis, which we explore in this section.

\subsection{The trouble with benchmarks}

The most common way to evaluate LLMs is to pit them against a human baseline, or against each other, on \textit{benchmarks} -- standardized tests or sets of tasks designed to assess specific capabilities of these models. Benchmarks are designed to provide a quantitative assessment of model performance in various domains, facilitating a direct comparison of their abilities in a controlled and systematic way. Unfortunately, benchmarking methods are plagued by limitations that hamper their reliability and adequacy to arbitrate dispute about the capacities of LLMs. These limitations include \textit{saturation}, \textit{gamification}, \textit{contamination}, and \textit{lack of construct validity} (which, as we shall see, are not all independent concerns, but are interrelated).

New natural language processing (NLP) benchmarks tend to saturate at an accelerating pace, meaning that LLMs quickly surpass the human baseline   \citep{kielaDynabenchRethinkingBenchmarking2021,ottMappingGlobalDynamics2022}. This is particularly striking for so-called `natural language understanding' (NLU) benchmarks. Taken at face value, saturation on these benchmarks would suggest that the best-performing NLP models `understand' language better than humans themselves. However, such success is often marred by independent examples of obvious failures, suggesting that performance saturation is not reliable evidence that LLMs actually surpass humans on the cognitive ability or aptitude (the `target construct') that the benchmark was designed to assess.\footnote{In the case of NLU, the target construct -- language `understanding' -- is itself ill-defined. See our discussion of semantic competence in Part I \citep{millierePhilosophicalIntroductionLanguage2024} and our discussion of construct validity below.}

Several factors explain benchmark saturation. One such factor is the gamification of benchmarks. Intense competition for best model performance on leaderboards -- also known as `SOTA' (state-of-the-art) chasing -- ultimately undermines the very purpose of behavioral evaluation. It results in models optimized to improve scores on proxy metrics targeted at benchmarks rather than the attribute the benchmark was designed to assess. This can be seen as an example of Goodhart's Law, commonly stated as the idea that when a measure becomes a target, it thereby ceases to be a good measure \citep{goodhart1975problems}. In the context of benchmark creation, statistical relationships identified from empirical observation and used to construct proxy metrics measured by test items tend to break down when actively exploited and optimized \citep{manheimCategorizingVariantsGoodhart2018}. This results in models that may overperform on a benchmark, yet lack the capacity supposed to be measured by that benchmark, at least to the degree suggested by their performance.\footnote{See \cite{gururanganAnnotationArtifactsNatural2018} for an example of how artifacts in benchmark construction can amplify the divergence between proxy metrics and target construct as models saturate the benchmark.}

Another phenomenon that contributes to benchmark saturation is data contamination. This occurs when test data -- including benchmark items and their solutions -- leak into the training data, either by chance or by design. With LLMs trained on internet-scaled data, contamination is increasingly difficult to avoid because benchmarks are widely reproduced and discussed online \citep{openaiGPT4TechnicalReport2023}. Detecting contamination is a challenge because simple variations of test items, such as paraphrases, can easily evade common detection measures based on string matching \citep{yangRethinkingBenchmarkContamination2023}. This calls into question the significance of some LLMs' reported performance on certain tests. For example, there is evidence that GPT-4's training data was contaminated with solutions to problems from the competitive programming contest Codeforces, such that it performs much worse than reported on old or recent problems that did not make it into its training data \citep{robertsDataContaminationLens2023}.

A broader concern with LLM benchmarks pertains to the construct validity of their target. Construct validity refers to the degree to which a test accurately measures the theoretical construct it purports to assess.  Standardized tests designed for humans are (ideally) meaningful because they are conceived to measure some underlying skill or capacity, such that good test performance should generalize to relevant real-world situations. However, establishing construct validity for LLM benchmarks is more challenging. Constructs targeted by common benchmarks, such as `understanding' or `reasoning,' are often abstract, multifaceted, and implicitly defined with reference to human psychology. Simply using test items validated for human subjects will not do, not only because of the potential for contamination, but also because validity is conditional upon background theoretical assumptions about test subjects. In other words, background assumptions about human cognition on which human-centric tests are premised can impact their applicability to LLMs. As a result, an LLM achieving good performance on a validated test designed to measure a capacity $\phi$ in humans may not constitute adequate evidence that the LLM has $\phi$.

As an example, consider the contentious case of Theory of Mind (ToM) -- the cognitive ability to understand and attribute mental states to oneself and others, and to recognize that these mental states may differ from one's own. Using classic verbal false belief tasks adapted from the developmental psychology literature on ToM, \cite{kosinskiTheoryMindMight2023} found that GPT-3 performed comparably to nine-year-old children. However, \cite{ullmanLargeLanguageModels2023} complicates this picture by showing that introducing minor conceptual variations on these test items, while preserving the requirement for false belief inference, significantly degrades LLM performance. This disparity suggests that GPT-3's impressive performance on original test items does not provide adequate evidence of ToM. This could certainly be due to some form of data contamination -- after all, there is hardly any doubt that false belief tasks feature in GPT-3's training data. But \cite{ullmanLargeLanguageModels2023} also suggests that even if LLMs matched human performance on challenging variations of classic false belief tasks, we should be careful not to jump to conclusions about their putative ToM aptitude. Indeed, tests designed with human cognitive processes in mind warrant inferences about the target construct given plausible assumptions about the mechanisms that drive human performance on these tests. These assumptions may not translate to artificial systems like LLMs without independent motivation. For example, the developmental trajectory and cognitive architecture of human ToM may lend itself to simulation-based or causal models of other minds that generalize flexibly. In contrast, an LLM could potentially achieve similar task performance by memorizing and interpolating patterns in training data, without engaging in the same underlying reasoning. Thus, the construct validity of ToM tests depends not just on input-output mappings, but on the nature of the algorithms and representations that generate those outputs. We have independent reasons to posit Theory of Mind as a core socio-cognitive capacity in humans, based on its early emergence, specificity to agents, and dissociability from general intelligence. Analogous supplementary evidence might be needed to corroborate claims about emergent ToM in LLMs.

Most of the limitations of LLM benchmarking are inherent to behavioral evaluation more generally. This point is often highlighted with reference to the distinction between \textit{performance} and \textit{competence} \citep{chomskyAspectsTheorySyntax1965,firestonePerformanceVsCompetence2020}. The distinction is most commonly invoked by skeptics about LLMs to suggest that performance success, such as high scores of benchmarks, need not reflect the competence associated with the underlying construct \citep{kielaDynabenchRethinkingBenchmarking2021}. Humans may achieve good performance because they exercise some cognitive capacity $\phi$ (e.g., ToM), while LLMs may achieve similar performance through completely different means (e.g., memorization of linguistic patterns that correlate with common ToM evaluations). The core concern here is that input-output mappings provide insufficient evidence about the \textit{mechanisms} that lead from input to output in a given system. Consequently, serious research on the capacities of LLMs should go beyond behavioral evaluation, and seek to understand how they process information mechanistically. 

It should be noted that the performance-competence distinction cuts both ways. In humans, it is common to disregard performance failures as evidence of a lack of competence, because such failures may be explained away by contingent constraints on cognitive function (e.g., limitations of working memory or attention). Most famously, Chomskyan linguists argue that performance errors are mostly irrelevant to linguistic competence, because they are the result of processing limitations or external factors rather than a reflection of the underlying linguistic system.\footnote{There is much debate about how to delineate the performance/competence distinction, as well as some confusion which is partly due to somewhat inconsistent characterizations in Chomsky's work itself. For an enlightening exposition of the distinction in linguistics, see \cite{dupreRealismObservationView2022}. For a discussion of its relevance to the NLP and LLMs, see \cite{dupreWhatCanDeep2021} and \cite{milliereLanguageModelsModelsforthcoming}.} Many cognitive scientists have called into question the existence of such an absolute gap between performance and competence \citep{tomaselloConstructingLanguage2009,christiansenCreatingLanguageIntegrating2016}. Another concern is that hypotheses about performance limitations need independent empirical support (lest they be invoked ad hoc to preserve a classical model in the face of empirical disconfirmation), but this independent evidence is rarely provided even in the case of humans, and direct investigations of performance factors can clash with competence theories \citep{franksExplanationCognitiveSciences1995,tomaselloConstructingLanguage2009,theakstonRolePerformanceLimitations2001}. Nonetheless, it is intriguing to consider how the dissociation between performance and competence might also apply to LLMs in both directions. That is, just as LLMs' impressive performance on benchmarks may always not accurately reflect their true competence, one might also consider that the existence of some performance errors do not always conclusively establish the absence of such competence. 

\subsection{Mechanistic explanation}

In science, mechanistic explanations aim to reveal the causal structure underlying a phenomenon of interest by describing the organized entities and activities that are responsible for producing or maintaining that phenomenon \citep{machamerThinkingMechanisms2000}. More precisely, a mechanistic explanation identifies the component parts of a mechanism, characterized by their properties and capacities, the causal interactions between these component parts, and the organization of the parts and activities such that they give rise to the phenomenon. Such explanations stand in contrast to purely descriptive or phenomenological models that simply \textit{re-describe} the phenomenon itself, as well as covering-law explanations that explain by subsuming the phenomenon under empirically discovered regularities or governing laws. Mechanisms explain by revealing \textit{how} the phenomenon arises from the causal structure of the system, not merely describing empirical regularities in which it partakes. As such, mechanistic explanations reveal opportunities for manipulation and control over the phenomenon in a way that descriptive or law-based explanations do not.

Understanding the behavior of a simple system like a mechanical clock is easy enough -- one can simply open it up and observe the mechanism at work. This is not so straightforward with more complex systems, like the weather, the brain, or artificial neural networks. Neural networks are often described as `black boxes' precisely because the causal mechanisms that explain their behavior seem opaque to scrutiny. As we emphasized in Part I \citep{millierePhilosophicalIntroductionLanguage2024}, simply staring at the learning objective, architecture, or parameters of LLMs will reveal neither how they exhibit their remarkable performance on challenging tasks, nor what functional capacities can be meaningfully ascribed to them. In principle, one could even provide a complete mathematical description of an LLM as a giant composite function, consisting in an absurdly complex sequence of linear and nonlinear transformations across many layers; but such a description, on its own, would be useless to provide a genuine explanation of the network's behavior in specific contexts. The `black box' metaphor underscores this chasm: it highlights the difficulty to trace precise causal pathways in the network through which specific inputs are transformed into specific outputs. This is why merely \textit{re-describing} what an LLM does in terms of next-token prediction or matrix multiplication -- what we called the `Re-description Fallacy' in Part I -- cannot possibly settle philosophical debates that are fundamentally about causal mechanisms.

The search for causal mechanisms has become central across the life sciences and cognitive science. Like their artificial counterparts, biological neural networks are `black boxes'; yet neuroscientists are engaged in the project of uncovering multilevel mechanisms underlying psychological capacities and nervous system functions \citep{craverExplainingBrainMechanisms2007}. At the molecular and cellular levels, they describe mechanisms of protein synthesis, gene expression, and synaptic transmission; at an intermediate level are mechanisms of neuron spiking and oscillation as well as mechanisms of development and synaptic plasticity; at higher levels are mechanisms underlying learning, memory, reasoning, and other psychological capacities.

Mechanisms explain by opening up the causal `black box' linking cause and effect -- revealing the internal entities, activities and organization that transmit causal influence through the system. The notion of intervention is central to this explanatory project. In the philosophy of science, interventionism holds that causal relationships are best understood in terms of what would change under interventions or manipulations to parts of the system \citep{woodwardMakingThingsHappen2005}. More specifically, $X$ is considered a direct cause of $Y$ if and only if there exists a possible intervention on $X$ that will change $Y$ (or the probability distribution of $Y$) when all other variables in the system are held fixed. This relationship is asymmetric -- intervening on $Y$ does not change $X$ if the causal arrow truly points from $X$ to $Y$. Interventionism eschews regularity or correlational notions of causation, recognizing that a system’s behavior depends on more than merely observing regular successions of events. In the context of mechanistic explanation, interventions involve altering specific parts of a posited mechanism piecemeal in order to learn about their causal contribution to the target phenomenon. For example, pharmacology targets specific receptor mechanisms or cell signaling pathways; brain stimulation techniques activate or inhibit activity in restricted brain areas; optogenetics uses light to control neurons genetically modified to express light-sensitive channels; and knockout models remove genes hypothesized to be critical for mechanisms.  

A similar interventionist approach can be applied to unravel causal mechanisms in artificial neural networks like LLMs. Of course, there are major disanalogies between biological nervous systems evolved over millions of years and artificial neural networks designed by human engineers. Nevertheless, the motivation of interventionist research is similar: to achieve explanatory understanding by revealing multilevel mechanisms, not merely observing input-output patterns. Like neuroscientists, computer scientists can aim for explanatory understanding linking particular components of neural networks and patterns of internal activations to specific capacities, such as translating between languages or answering arithmetic questions. This explanatory project is often described as \textit{mechanistic interpretability}. In a broad sense, mechanistic interpretability could be characterized as the search for mechanistic explanations of the behavior of deep neural networks, including LLMs. As we shall see, however, the phrase is often used in a more restrictive sense, to denote a particular set of theoretical assumptions and intervention methods used to achieve mechanistic explanations in machine learning.\footnote{The lack of a common lexicon in interventionist research on neural networks is rather unfortunate -- many different labels exist for similar ideas and methods, with `interpretability' being a particularly confusing term \citep{liptonMythosModelInterpretability2018}. In what follows, we will attempt to give a cohesive overview of this fragmented literature.}

\subsection{Opening up the black box}

There are three main methodological approaches to investigate the inner structure of neural networks: \gls{probing}, \gls{attribution}, and \textit{bona fide} \gls{causal-intervention}. Probing involves training a separate supervised classifier, known as a diagnostic probe, to predict certain properties (e.g., part-of-speech tags, dependency relations) from the model's internal activations \citep{alainUnderstandingIntermediateLayers2018,hupkesVisualisationDiagnosticClassifiers2018}. High accuracy in decoding a particular linguistic feature $F$ from a probe tuned to a pattern of activation $A$ in a subset of the network can be though to provide evidence that $A$ is sensitive to $F$, and provide information about its presence or absence for downstream processing. However, a probe's successful prediction of a certain linguistic feature from a model's activations does not necessarily mean that the feature plays a causal role in the model's behavior \citep{belinkovProbingClassifiersPromises2022}. In addition, probes can pick up on spurious correlations, thereby failing to distinguish between genuine representation and incidental associations \citep{hewittDesigningInterpretingProbes2019}. 

Probing can tell us that some information is likely present in the activations of the system, but not that it is in fact used for a particular purpose or function in generating the system's outputs. Indeed, the mere presence of usable information does not demonstrate that it is actually exploited. By way of analogy, suppose you throw block letters spelling a word into a pond. With the right apparatus, you might be able to decode the word's identity from the ripples formed by letters over the pond's surface; but this does not provide evidence that the pond represents (let alone `understands') the word in any meaningful sense. It merely shows that the block letters created surface patterns whose origin could be recovered by a sufficiently powerful decoding method. 

Methods that do not involve training a separate classifier -- also known as \gls{nonparametric} methods -- have been explored as an alternative to probing. For example, one can directly investigate the weights (or attention scores) that attention heads place on different parts of the input sequence. A high attention score on a particular word might be thought to provide evidence that the head is playing a role in processing the semantic or syntactic role of that word in the rest of the sentence. This approach to interpretability falls within the broader and somewhat loose umbrella of \gls{attribution} methods in deep learning. Attribution methods assign importance scores to input features to explain individual model predictions; in other words, they are meant to identify parts of the input that are most influential for the output. While attention patterns in Transformers intuitively provide clues about the relevance of each input token to model predictions, simply analyzing these patterns in a given layer of the network only offers limited explanatory power \citep{cheferTransformerInterpretabilityAttention2021}. In particular, visualizing attention patterns is thoroughly insufficient to draw strong conclusions the flow of information through the network. If we want to understand what information the model represents in the input, and how that information is processed through step-by-step computations across layers to drive output predictions, we need to intervene directly on the network to reveal its causal structure.

\subsection{Interventions on neural networks}

To assert that a certain activation pattern in a model genuinely represents a given feature, mainstream philosophical theories of representation require that three criteria be satisfied: (a) the activation pattern should carry information about the feature; (b) the activation pattern should influence the model's behavior in a task-relevant way, and (c) the model should be capable of misrepresenting the feature \citep{hardingOperationalisingRepresentationNatural2023}. Standard probing methods only provide evidence for the first criterion. Showing that some information about a feature is \textit{actually} used by the LLM to generate its outputs requires suitable \textit{interventions} on patterns of activation encoding that information. Changes in model behavior caused by such interventions should be consistent with the hypothesis that the model represents the target feature, but should not be explainable by mere perturbations from the training set. Accordingly, an increasingly large body of work uses targeted intervention methods to establish causal relationships between language models' internal representations and their behavior.

The simplest kind of intervention on neural networks – both artificial and biological – is an \gls{ablation}. In the context of artificial neural networks, ablation involves disabling or eliminating individual neurons or groups of neurons within a trained model to observe the resulting changes in behavior. By ablating neurons and observing the resulting change in performance metrics such as classification accuracy or reconstruction error, researchers can determine how much each neuron or group of neurons contributes to the network's overall function. Neurons that are critical to network performance will result in a substantial decline in metrics when ablated.

Indiscriminate ablations, common in the early days of connectionist research, involved disabling nodes at random. This approach was aimed at demonstrating general properties of neural networks. One key finding from such studies was the concept of \gls{gracefuldegradation}: like biological brains, the performance of neural networks tends to decline in a gradual, rather than abrupt, manner when parts of the network are damaged or disabled \citep{sejnowskiParallelNetworksThat1987,smolenskyProperTreatmentConnectionism1988}.

Targeted ablations, on the other hand, involve disabling specific nodes or modules believed to serve distinct representational or functional roles within the network. By analyzing the network’s altered behavior, researchers could infer the impact of the disabled nodes, thereby gaining insights into their hypothesized functions \citep{meyesAblationStudiesArtificial2019}. This method mirrors a common approach in neuroscience where the study of natural or induced brain lesions helps to unravel the functions of specific brain areas.

However, both indiscriminate and targeted ablation studies are somewhat limited in their ability to fully uncover representational or functional roles in neural networks. This limitation is partly due to the nature of how information is represented within these networks. Traditional views centered around \textit{localized} representations, positing that specific neurons or small groups of neurons could be responsible for representing distinct, complex stimuli or concepts. The classic example in neuroscience is the hypothetical `grandmother cell' -- a neuron that would activate exclusively in response to the mental image or concept of one's grandmother. While intuitively appealing, this model of localized representation has largely been abandoned in favor of a distributed model \citep{plautLocatingObjectKnowledge2010,barwichValueFailureScience2019}.

In practice, individual neurons or small clusters of neurons in neural networks often encode information about multiple concepts \citep{smolenskyNeuralConceptualInterpretation1986,rumelhartParallelDistributedProcessing1987,henighan2023superposition}. This is consistent with distributed representations: concepts are represented by patterns of activity that are spread across a larger number of neurons within the network. When components of distributed representations overlap, the same neurons might simultaneously represent multiple features – they are `polysemantic'. This property is studied under the name of `superposition'; for example, \cite{elhage2022superposition} demonstrate that superposition can allow networks to linearly represent many more features than they have neurons or dimensions, at the cost of some interference between features. The distributed model of representations in neural networks reflects a more integrated and holistic approach to neural processing, where information is not stored in isolation but as part of a dynamic, interconnected system. 

The shift from localized to distributed representation models has significant implications for interpreting the results of ablation studies. In a distributed system, disabling a node or a set of nodes affects more than just the immediate functionalities those nodes are associated with. It impacts the network-wide interplay of neural activities. Therefore, determining the specific effects of localized damage on the overall behavior of the network becomes increasingly complex \citep{jonasCouldNeuroscientistUnderstand2017}.

Fortunately, the level of control we have over artificial neural networks allows for more sophisticated causal interventions that improve on localized ablations by allowing us to `edit' distributed representations in ways consistent with representational or functional hypotheses, and verifying corresponding changes in behavior. For example, \cite{giulianelliHoodUsingDiagnostic2018} used a probe to identify activations associated with subject-verb number agreement, then modified these activations to improve the model's performance on a subject-verb agreement task. Such interventions provide more robust evidence about the causal role of specific internal features in a model's behavior.

A more sophisticated approach called `iterative nullspace projection' was developed by \cite{ravfogelNullItOut2020}. Iterative nullspace projection can determine whether some particular information is causally involved in a neural network's predictions by identifying and removing that information from distributed neural representations, and then assessing the consequence on model behavior. The approach involves iteratively projecting neural representations onto the \glspl{nullspace} of a probe to remove detectable information about user-defined target concepts.

More specifically, the first step is to train linear probe on internal neural representations to predict values of the concept of interest. For example, probes could be trained to classify grammatical number from the hidden state of a language model. The \gls{nullspace} of a probe is the subspace consisting of all activations for which the probe makes the same constant prediction. Projecting representations onto these nullspaces removes information that linearly correlates with concept values while preserving unrelated information. After projecting onto the first probe's nullspace, a second probe is trained on the transformed representations to predict the target concept. This process is repeated iteratively, with projections onto each new probe's nullspace removing additional detectable information about the target concept. Finally, the network makes predictions using the fully projected representations where linear information about the target concept has putatively been eliminated. If performance degrades, the target concept can be inferred to be causally significant for model performance, suggesting that the network exploits it for its original computations.

\begin{figure}[h!] 
    \centering 
    \includegraphics[width=0.8\linewidth]{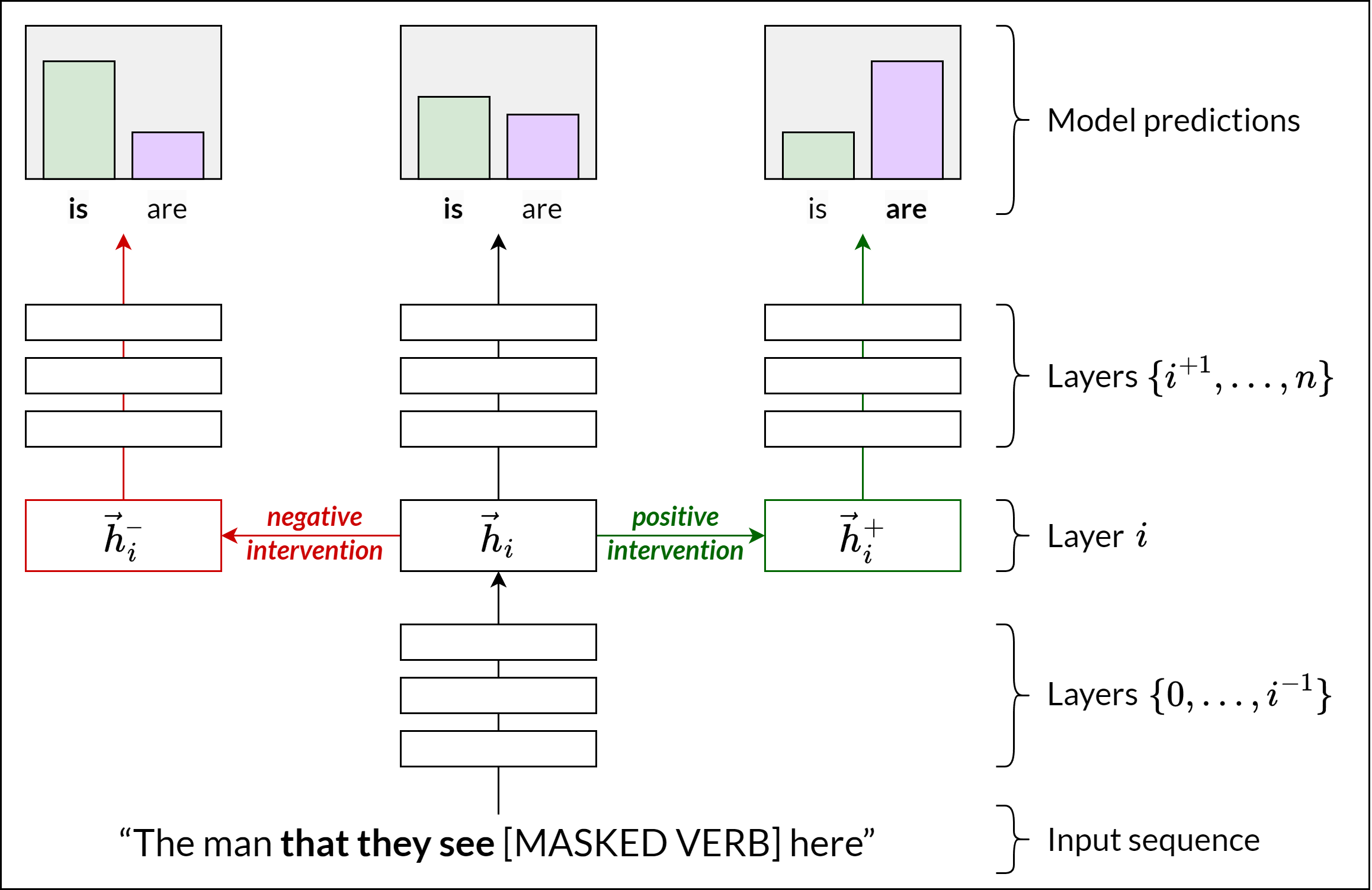} 
    \caption{\textbf{Iterative nullspace projection}. Given the representation $\vec{h_i}$ of a masked word in layer $i$ of a Transformer model, a probe is trained to predict relative clause boundaries. The probe's nullspace, which encodes information not relevant to relative clause boundaries, is identified. Two counterfactual representations, $\vec{h^{-}_{i}}$ and $\vec{h^{+}_{i}}$, are derived by projecting $\vec{h}_{i}$ onto the nullspace and then performing negative and positive interventions, respectively, along the probe's decision boundary. $\vec{h^{-}_{i}}$ encodes that the word is outside a relative clause, while $\vec{h^{+}_{i}}$ encodes that it is inside a relative clause, with other information preserved. The model's predictions and using these counterfactual representations are compared to its original prediction to assess the causal effect of the relative clause boundary information on the model's behavior in number agreement.}
    \label{fig:INSP} 
\end{figure}

For example, \cite{ravfogelCounterfactualInterventionsReveal2021} investigated whether information about relative clause boundaries is used by language models like BERT to predict subject-verb agreement across a relative clause. They trained a set of linear probes to predict whether a word is inside a relative clause, based on the model's contextual word embeddings. Each probe defines a direction in the representation space that separates words inside relative clauses from words outside relative clauses. Collectively, these directions span a `feature subspace' that contains relative clause information. The orthogonal complement of this subspace is the nullspace which does not contain information useful for predicting whether a word is inside a relative clause. Then, iterative nullspace projection is used to generate `counterfactual representations' for the masked verb token. The representation is projected into the relative clause feature subspace, and then flipped to the opposite side of the separating hyperplane, either towards the side containing relative words (`positive counterfactual') or away from that side (`negative counterfactual'). This process minimally modifies the representation to incorrectly encode that the word is inside or outside a relative clause. Finally, the effect of the counterfactual representations on the model's number agreement predictions is measured. If swapping in the positive counterfactual increases error rate and the negative counterfactual decreases error rate, this alignment with predictions from linguistic theory suggests the model uses relative clause boundary information appropriately for agreement, confirming the causal effect (fig. \ref{fig:INSP}). 

This method is directly inspired by counterfactual approaches to causal explanation in philosophy of science \citep{woodwardMakingThingsHappen2005}. Such approaches aim to isolate the causal contribution of some factor $X$ to an outcome $Y$ by minimally altering the value or presence of information about $X$ in the system, while holding all other factors fixed. This allows assessment of whether and how much the factor $X$ is exploited causally by the system to produce $Y$. The removal of detectable information about a target variable through iterative nullspace projection is analogous to a hypothetical intervention that breaks the links between that variable and the system while leaving other causal mechanisms intact. The comparison between original model outputs and outputs based on projected representations missing information about $X$ mirrors the evaluation of interventionist counterfactuals, which aims to answer questions such as ``What would happen to $Y$ if an intervention prevented information about $X$ from influencing the mechanism?''.

\subsection{Mechanistic interpretability}

Intervention methods such as iterative nullspace projection can help us determine whether some target concept, such as a particular syntactic feature, is represented by a language model and causally efficacious in its behavior. But this is insufficient to understand language models and other neural networks the way we understand classical computer programs. Indeed, the guiding ideal of interpretability research is to model the internal causal structure of neural networks, such that we can explain some behavior of interest in terms of a series of computational steps applied on the input. Mechanistic interpretability refers to this concerted effort to reverse engineer the internal computations performed by artificial neural networks. Rather than solely focusing on interpreting individual model predictions, it aims to provide a detailed and systematic understanding of how the model transforms inputs to outputs \citep{elhage2021mathematical}. The overarching goal of mechanistic interpretability is to open up the `black box' of neural networks by providing human-intelligible descriptions of the functional modules that drive the emergence of model behaviors.

Like systems neuroscience, mechanistic interpretability specifically targets the algorithmic level of analysis \citep{marrVisionComputationalApproach1982,lindsayTestingMethodsNeural2023}. An algorithmic explanation elucidates the flow of information through a network and the sequence of operations performed upon it. Crucially, it abstracts away from engineering (or biological) details, instead focusing on the computations performed by the system. Algorithmic explanations support counterfactual inferences about how changing inputs or network components would affect computations and outputs. This involves identifying structural elements like motifs and circuits that are reusable, such that manipulating them produces systematic effects across many inputs and conditions.

Mechanistic interpretability specifically seeks to reverse-engineer neural networks in terms of learned \textit{features} and reusable \textit{circuits} that operate on those features. Features refer to human-interpretable properties of the input data that the model represents internally. For computer vision models, such features might be edges, textures, or shapes. For language models, features may correspond to part-of-speech tags, named entities, or semantic relationships. By studying a network's internal activations, researchers aim to determine which features are encoded where.

Circuits, in turn, are chains of operations that detect certain input features and transform them into output features. The goal is to decompose an entire neural network into a hierarchy of understandable features and circuits that process information step-by-step. This modular view would explain how trained models transform inputs to outputs via learned features and program-like circuits, providing a form of algorithmic understanding.\footnote{In fact, it is even possible to \textit{program} a Transformer from scratch using a specialized programming language like RASP (Restricted Access Sequence Processing Language). RASP allows mapping the basic components of a Transformer, like attention and feed-forward layers, into simple primitives that can be composed into programs. These RASP programs can then be compiled into the weights of a Transformer network that implements the specified computation \citep{weissThinkingTransformers2021}. More recently, methods have been developed to train Transformers whose weights are constrained to implement human-interpretable RASP-like programs. These Transformer Programs can be learned end-to-end from data and decompiled back into discrete, readable code \citep{friedmanLearningTransformerPrograms2023}. Such approaches provide a more direct way to understand the computations of a Transformer in terms of modular algorithmic components.}

More formally, a given neural network can be modeled as a graph (sometimes called a computational graph) whose nodes correspond to components of the network at some level of granularity – such as attention heads or individual neurons. The directed edges between nodes represent the flow of information and computations in the neural network. The connectivity defined in the computational graph needs to faithfully represent the actual computation flow in the neural network. In this model, a circuit is a subgraph within the overall computational graph that implements some specific behavior or functionality of interest. The goal of representing the neural network as a computational graph is to systematically analyze it and localize particular circuits that are causally responsible for certain model behaviors.

Once candidate circuits are found, their functionality must be validated through causal interventions. This often involves surgically replacing components of suspected circuits and observing the effect on model outputs. Validated circuits can then be composed into a hierarchical understanding of the succession of transformations enacted by the model. The resulting mechanistic explanation should provide significant behavioral control, for instance allowing targeted editing of model computations.

\begin{figure}[h!] 
    \centering 
    \includegraphics[width=0.6\linewidth]{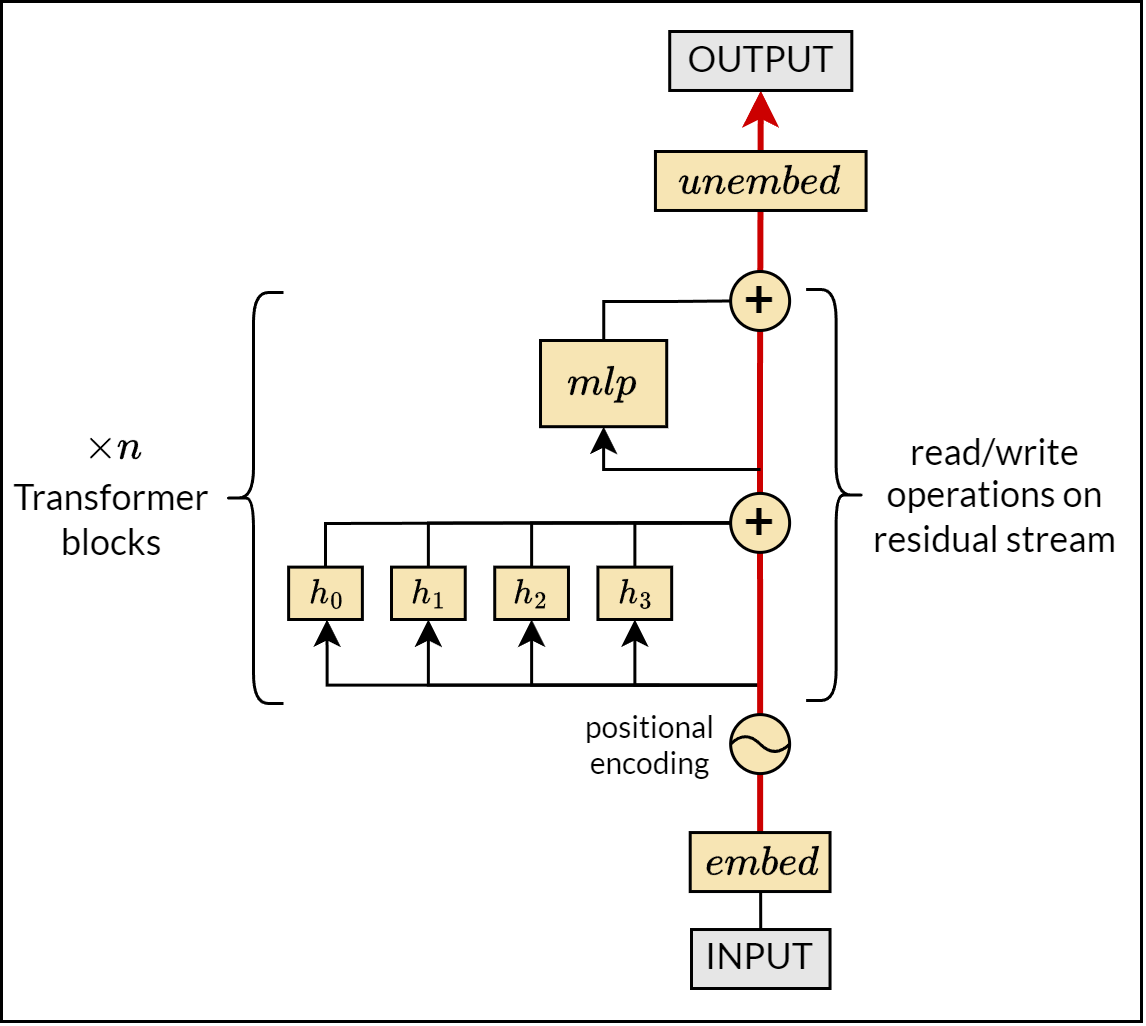} 
    \caption{\textbf{The \gls{residual-stream} view of the Transformer.} Each input token is first embedded into a dense vector representation and combined with a positional encoding injecting information about their position in the input sequence. This forms the initial state of the "residual stream" (depicted by the red arrow), which flows through the entire network. Each Transformer block, consisting of multi-head self-attention and a multi-layer perceptron (MLP), reads from the residual stream, transforms the representation, and writes the result back into the stream via residual connections. This process is repeated across multiple Transformer blocks. Finally, the output of the residual stream is `unembedded' to map the transformed representation back to the original token space. In this view, the Transformer's components are seen as operators that successively refine the residual representation.}
    \label{fig:residual-stream} 
\end{figure}

The picture that emerges from mechanistic interpretability research is that Transformer models can be viewed as containing parallel processing streams (also known as `\glspl{residual-stream}'), one at each input token position (fig. \ref{fig:residual-stream}). While each residual stream encodes information in a very high-dimensional vector space, attention heads at each layer operate over much smaller (low-dimensional) subspaces of the stream that may not overlap with one another. Making use of these disjoint subspaces allows Transformers to route information about tokens and their dependencies dynamically across layers and positions. Specifically, each stream is functionally analogous to a \gls{addressable-memory}, in which attention heads can write to and read from subspaces of the main embedding space. Such information may include, for example, syntactic dependencies between tokens in the input sequence.

\begin{figure}[h!] 
    \centering 
    \includegraphics[width=0.7\linewidth]{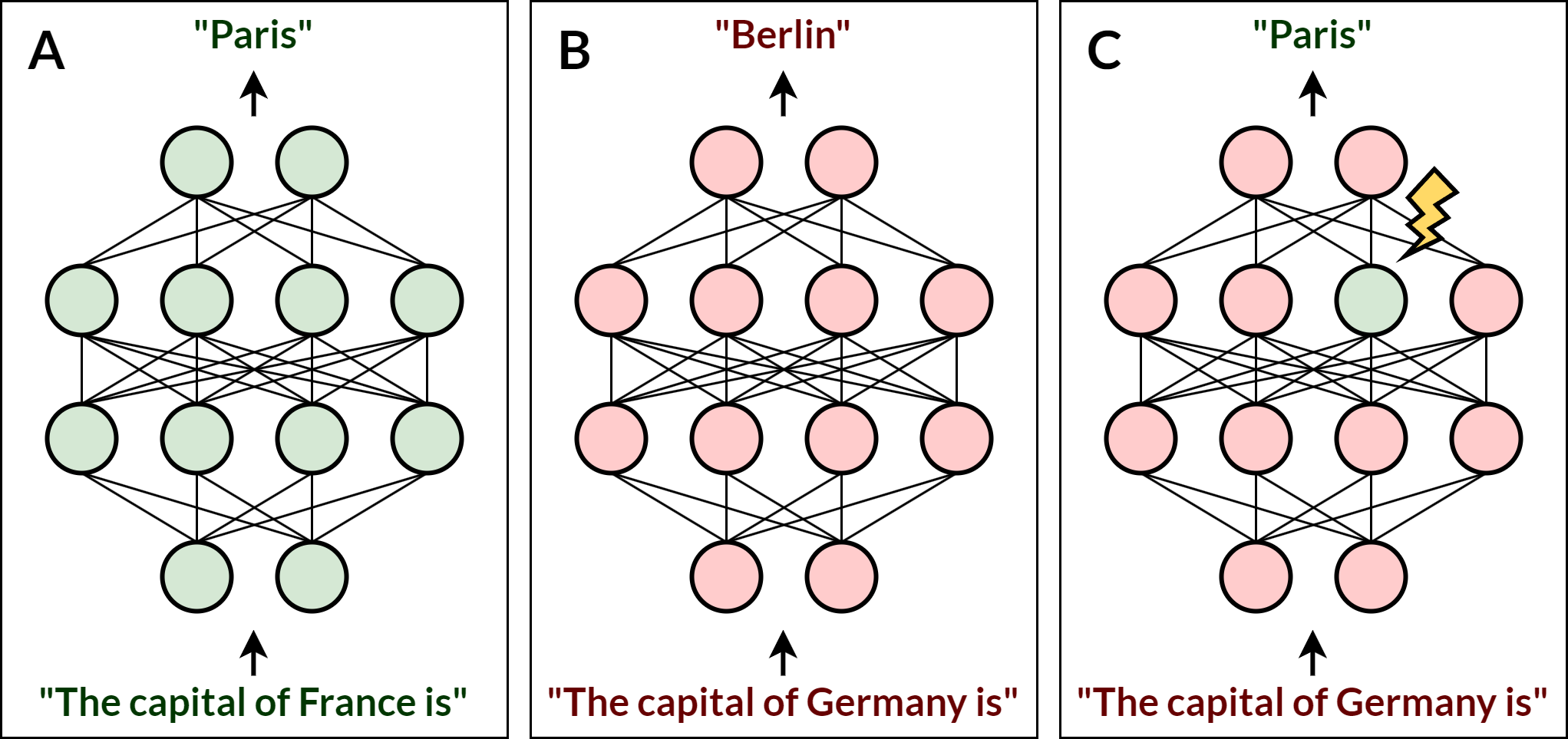} 
    \caption{\textbf{Activation patching.} \textbf{A.} In the original forward pass, the model takes as input the prompt ``The capital of France is'' and outputs the correct answer ``Paris''. The model activations from this forward pass are cached. \textbf{B.} In the alternative forward pass, the prompt is changed to ``The capital of Germany is''. The model now outputs ``Berlin'' as the answer. \textbf{C.} Activation patching is applied in a third forward pass. The model is given the alternative prompt ``The capital of Germany is'' once again, but a specific component of the model has its activations replaced (\textit{patched}) with those from the original forward pass on the France prompt. This causes the model to output ``Paris'' instead of ``Berlin'', despite being given the Germany prompt. The restoration of the original output through patching the activations of a particular model component provides evidence that this component encodes information that is causally implicated in the target behavior.}
    \label{fig:activation-patching} 
\end{figure}

A common intervention method used in mechanistic interpretability research is activation patching \citep{zhangBestPracticesActivation2023}, also known as causal tracing \citep{mengLocatingEditingFactual2023} and interchange intervention \citep{geigerCausalAbstractionsNeural2021} (fig. \ref{fig:activation-patching}). The basic method involves three steps. First, the neural network must be run on an original input that relates to some behavior of interest (e.g., answering factual questions). For example, the original input might be ``The capital of France is...'', with the expected output being ``Paris''. Importantly, the activations of the network during the forward pass on this original input are cached for later use. The second step is to run the model again with an alternative input that introduces a key variation on the original input that changes the behavior (output). For example, an alternative input might be ``The capital of Germany is...'', with the expected output being ``Berlin''. Finally, the alternative input is run again but a specific component of the network's activation is swapped for its cached value from the original forward pass – an intervention known as `patching', because it patches the alternative forward pass with a component of the original forward pass. The effect of patching a component's activation is then evaluated by comparing the model's performance in the regular forward pass on the alternative input versus the patched forward pass on the same input. For example, the metrics used could be the probability the model assigns to the original, correct token (``Paris''), or the difference in logits between the correct and incorrect tokens (``Paris'' vs ``Berlin''). Intuitively, changing the input hurts model performance on the expected behavior, while patching activations from the original run helps restore it. So if patching a particular component leads to a significant restoration of performance (e.g., an increase in the probability of ``Paris'' being the output), it suggests that component is important for the model's behavior on the task (i.e. the component not only contains information on the target feature, but that information is causally implicated in producing the desired result). By iterating this procedure over many components of the computational graph, such as attention heads in a Transformer model, activation patching aims to identify the key circuits that enable behaviors like factual recall or logical reasoning. 

The mechanistic interpretability framework has been applied to many different problems, and has started shedding light on some abilities of Transformers. It is worth noting that much of this research is painstaking work conducted with toy models that are easier to interpret. While these toy models are based on the Transformer architecture, they may differ from LLMs in terms of architectural details, learning objective, dataset, and of course size (parameter count). Nonetheless, efforts are underway to automate and scale mechanistic interpretability techniques to bona fide LLMs, with promising initial results \citep{wuInterpretabilityScaleIdentifying2023,conmyAutomatedCircuitDiscovery2023,syedAttributionPatchingOutperforms2023}. In what follows, we will briefly illustrate the fruitfulness of the mechanistic interpretability program through three case studies. 

\subsubsection{Case study 1: Induction heads}

A canonical example of circuit discovery in language models through the tools of mechanistic interpretability is that of so-called `induction heads' \cite{olsson2022context}. These are specialized attention heads that emerge through training even in very small models, and implement a form of pattern completion, allowing models to repeat or generalize sequences based on similar context patterns.

Specifically, induction heads use a `prefix matching' attention pattern to look back over previous tokens and detect if any match the current token. Rather than relying merely on memorized statistics about which tokens tend to follow others, they flexibly attend to whichever prior token is most similar to the current one based on learned representations. If a previous token is sufficiently similar, the induction head will attend to the next token after it. The head then increases the probability of that attended next token, effectively predicting that the current sequence will continue like the previous matched sequence. For example, if the input contains the sequence ``...the cat sat on the mat. The cat...,'' the induction head matches the second ``cat'' token to the first one, attends to ''sat'' from the first sequence, and increases the likelihood of outputting ``sat'' again. This allows Transformer models to repeat sequences and generalize patterns.

\begin{figure}[h!] 
    \centering 
    \includegraphics[width=1\linewidth]{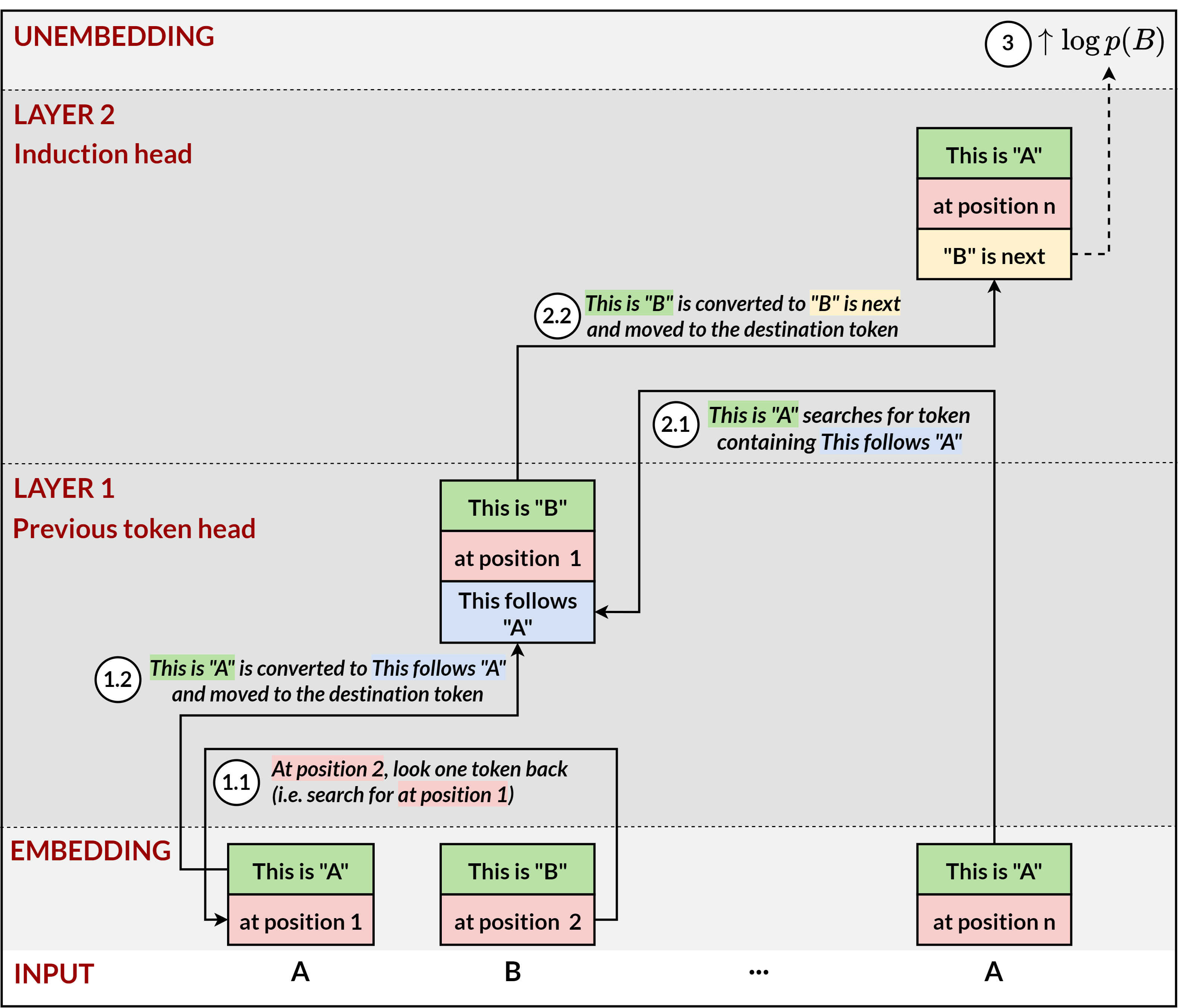} 
    \caption{\textbf{A schematic illustration of the induction head circuit in a two-layer Transformer model.} At the embedding stage, each token from the input sequence is encoded as a vector, together with information about its position in the sequence. The first layer contains an attention head – known as the \textit{previous token head} – that acquired a specialized function during training. When processing token \texttt{[B]} at position $2$, the previous token head does the following: (1.1) it attends to the previous token at position $1$; (1.2) It writes the identity of this preceding token to a dedicated subspace of the residual stream at the current position (position $2$), effectively storing the information ``the token before me is \texttt{[A]}''. Layer 2 contains another specialized attention head known as the \textit{induction head}. When processing the second instance of \texttt{[A]} at position $n$, the induction head does the following: (2.1) it queries the \gls{residual-stream} for information in the `previous token' subspace matching the current token's identity; (2.2) having located this previous token information in the residual stream at position $2$, it retrieves the identity of the token at that position (\texttt{[B]}), then writes this identity to a dedicated subspace of the residual stream at the current position (position $n$), effectively storing the information ``predict that the next token will be \texttt{[B]}''. (3) The unembedding layer maps information in the `next token' subspace at position $n$ to an increased logit for \texttt{[B]} at position $n+1$, which translates to an increased log likelihood of \texttt{[B]} being predicted as the next token.}
    \label{fig:induction-heads} 
\end{figure}

Importantly, the computations performed by induction head circuits do not rely on \gls{bigram} statistics memorized from the training data; rather, they operate over abstract patterns in the input sequence (prompt), even if the latter does not contain familiar strings. As such, induction head circuitry can be seen as an instance of what \cite{sheaMovingContentspecificComputation2023} called non-content-specific computations in neural networks: computations that apply the same procedure or algorithm irrespective of the specific content represented at input and output. Specifically, when the network is processing a sequence \texttt{[A][B]...[A]}, the induction head circuit can be characterized algorithmically as storing the value of the first \texttt{[A]} token in a specific subspace of the \gls{residual-stream} -- which we may call the \texttt{previous\_token} subspace -- at the position of the \texttt{[B]} token. Importantly, this operation occurs whatever that specific value (content) of \texttt{[A]} might be. Functionally, the \texttt{previous\_token} subspace in which this information is stored operates rather like a variable in a classical symbolic program. Its value is accessed at the next layer by the second part of the induction head circuitry, which reads the content of the \texttt{previous\_token} subspace in the residual stream at the position of the \texttt{[B]} (fig. \ref{fig:induction-heads}). This is somewhat analogous to the indirect addressing mechanism that enables variable binding in classical systems: an attention head stores information about the token that precedes another token in a dedicated subspace of the \gls{residual-stream} (what we called the \texttt{previous\_token} subspace), such that a distinct attention head in a later layer may retrieve it for downstream processing. This kind of circuit satisfies the distinction between storage and use that is crucial to computation over bound variables \citep{smolenskyProperTreatmentConnectionism1988,gallistelMemoryComputationalBrain2011}.\footnote{For a discussion of induction heads circuitry as implementing a form of variable binding, and the implications of this view for the debate about compositionality in connectionist models, see \cite{millierePhilosophyCognitiveScienceforthcoming}.}

Circuits using induction heads may be an important building block for the advanced in-context learning abilities exhibited by LLMs. Even in toy models, for instance, if the context contains ``...the cat sat on the mat. The dog...'', an induction head can generalize that ``dog'' signals repeating the sequence, despite never seeing ``dog'' next to ``sat'' during training. Research shows that Transformer models excel at discovering complex abstract patterns in context and extrapolating from them \cite{mirchandaniLargeLanguageModels2023}.

\subsubsection{Case study 2: Modular addition}

Mechanistic interpretability is not just helpful to investigate how trained neural networks process information through algorithmic circuits, but also how they \textit{learn} such algorithms during the course of training. Studying the training dynamics of neural networks is important, because it can provide insights into learning transition phases that are highly relevant to ongoing debates about the capacities of LLMs. Thus, \cite{nandaProgressMeasuresGrokking2022} investigated the puzzling phenomenon of \gls{grokking}, where neural networks trained on algorithmic tasks with regularization initially overfit to the training data but later suddenly generalize after many training steps. As a case study, the authors trained small Transformer models on modular addition tasks. They find these networks exhibit grokking, initially overfitting but later learning to generalize.

\begin{figure}[h!] 
    \centering 
    \includegraphics[width=1\linewidth]{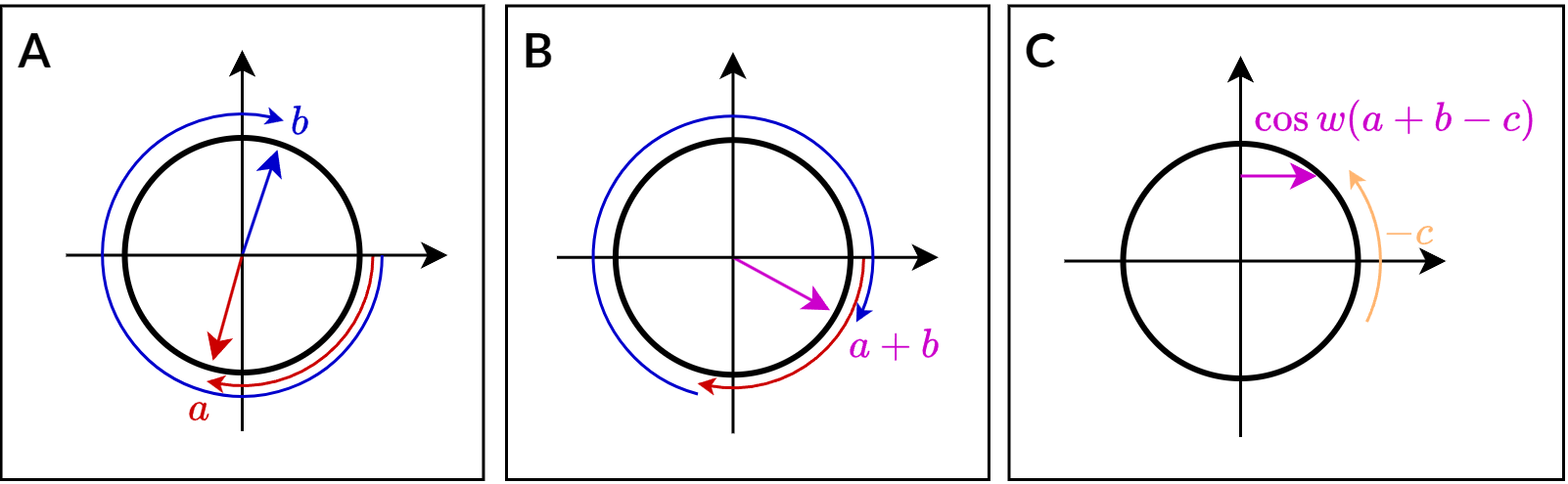} 
    \caption{\textbf{A learned algorithm for modular addition} (figure adapted from \citet{nandaProgressMeasuresGrokking2022}). \textbf{A. Embedding projection.} Given two input numbers $a$ and $b$ in the modular addition $a + b \equiv c \mod P$, the model uses its embedding matrix to project each number onto a corresponding rotation around the unit circle. The embedding matrix essentially memorizes a mapping between each possible input number and a specific rotation amount, converting the numbers into geometric representations. \textbf{B. Rotation composition.} The model composes the two rotations generated for $a$ and $b$. This step effectively adding the two rotation amounts together, resulting in a new, single rotation that represents the sum $a+b$ in modular arithmetic. In modular arithmetic, numbers `wrap around' after exceeding the modulus $P$, so if $a+b$ is greater than $P$, the resulting rotation will correspond to $a+b \mod P$, which is the remainder when $a+b$ is divided by $P$. \textbf{C. Output decoding.} To produce the output logits (raw scores used for next-token prediction), the model considers each possible result $c$ (ranging from $0$ to $P-1$) and performs a reverse rotation by $-c$. This step essentially checks, for each $c$, whether undoing the rotation by $c$ results in a rotation that matches the one representing $a+b \mod P$. The output $c$ that produces the rotation most closely matching the $a+b \mod P$ rotation is assigned the highest logit. This works because the trained model ensures that the correct $c$ satisfying $a + b \equiv c \mod P$ will undo the rotation by exactly the right amount to point back to the $a+b \mod P$ rotation. The trigonometric functions cosine and sine are used to implement these rotations and achieve the desired result mathematically using angle addition identities, but conceptually, the algorithm is based on representing numbers as rotations and composing these rotations together.}
    \label{fig:modular-addition} 
\end{figure}

To understand this phenomenon, they reverse engineered the mechanisms learned by these networks using techniques from mechanistic interpretability. That found that the network learns to perform modular addition tasks by mapping inputs onto rotations in the plane and composing those rotations using trigonometric identities (fig. \ref{fig:modular-addition}). This clever algorithm, dubbed `Fourier multiplication,' allows the network to perform addition modulo the prime.

Using this understanding, Nanda et al. defined new progress metrics allowing them to study the training dynamics of the models as they learn to perform modular addition algorithmically. They identified three distinct learning phases, each marked by continuous progress on certain metrics:

\begin{enumerate}
    \item \textit{Memorization phase}: In the early part of training, the models fit to the training data by simply memorizing input-output pairs. Performance on the test set remains low while performance on the training set increases rapidly, indicating the models are overfitting.
    \item \textit{Circuit formation phase}: After memorization, there is a transition period where the models internalize the algorithm for modular addition using trigonometric identities and rotations (the `Fourier multiplication' circuit). However, performance on the test set remains low, implying memorization components still persist.
    \item \textit{Cleanup phase} Finally, weight decay drives the removal of the initial memorization components. Performance on the test set abruptly improves to match performance on the training set at the end of this phase, corresponding to the `grokking' transition in generalization capability.
\end{enumerate}

The entire phenomenon is thus not a sudden onset of generalization ability, but rather a gradual amplification of structured mechanisms encoded in the weights, followed by pruning of unnecessary components. Importantly, these findings speak against the skeptical view of LLMs discussed in Part I \citep{millierePhilosophicalIntroductionLanguage2024}, according to which they are analogous to giant lookup tables with memorized input-output pairs. While the initial training phase does support that view, subsequent phase transitions show that Transformer models are perfectly capable to learn general, rule-like algorithms to solve tasks, including tasks as rigid and well-defined as arithmetic problems.

\subsubsection{Case study 3: World models}

As we discussed in the complement paper \citep{millierePhilosophicalIntroductionLanguage2024}, there is an ongoing debate about whether LLMs possess world models. While behavioral evaluations are generally not helpful to settle this debate, mechanistic interpretability shows promise to uncover what the LLMs and related Transformer models represent internally.

To investigate whether Transformer models trained to predict sequences can learn interpretable world representations rather than mere surface statistics, \cite{liEmergentWorldRepresentations2023} focus on the board game Othello as a simplified yet non-trivial domain. They trained a GPT variant (Othello-GPT) to sequentially predict tokens representing Othello board positions. The only inputs to the model are move sequences derived from game transcript; no explicit game rules or board structure are provided. After training on championship games and synthetic games, Othello-GPT recommends legal moves with high accuracy, suggesting it has learned more than surface statistics. Furthermore, nonlinear probes reliably predict board states from model activations, implying that a nonlinear representation of the board state had emerged during training. To validate the probes' accuracy, the authors performed intervention experiments that modify Othello-GPT's internal activations to reflect altered board states. The model's subsequent move predictions change accordingly, confirming the causal role of these latent board state representations. 

\begin{figure}[h!] 
    \centering 
    \includegraphics[width=0.8\linewidth]{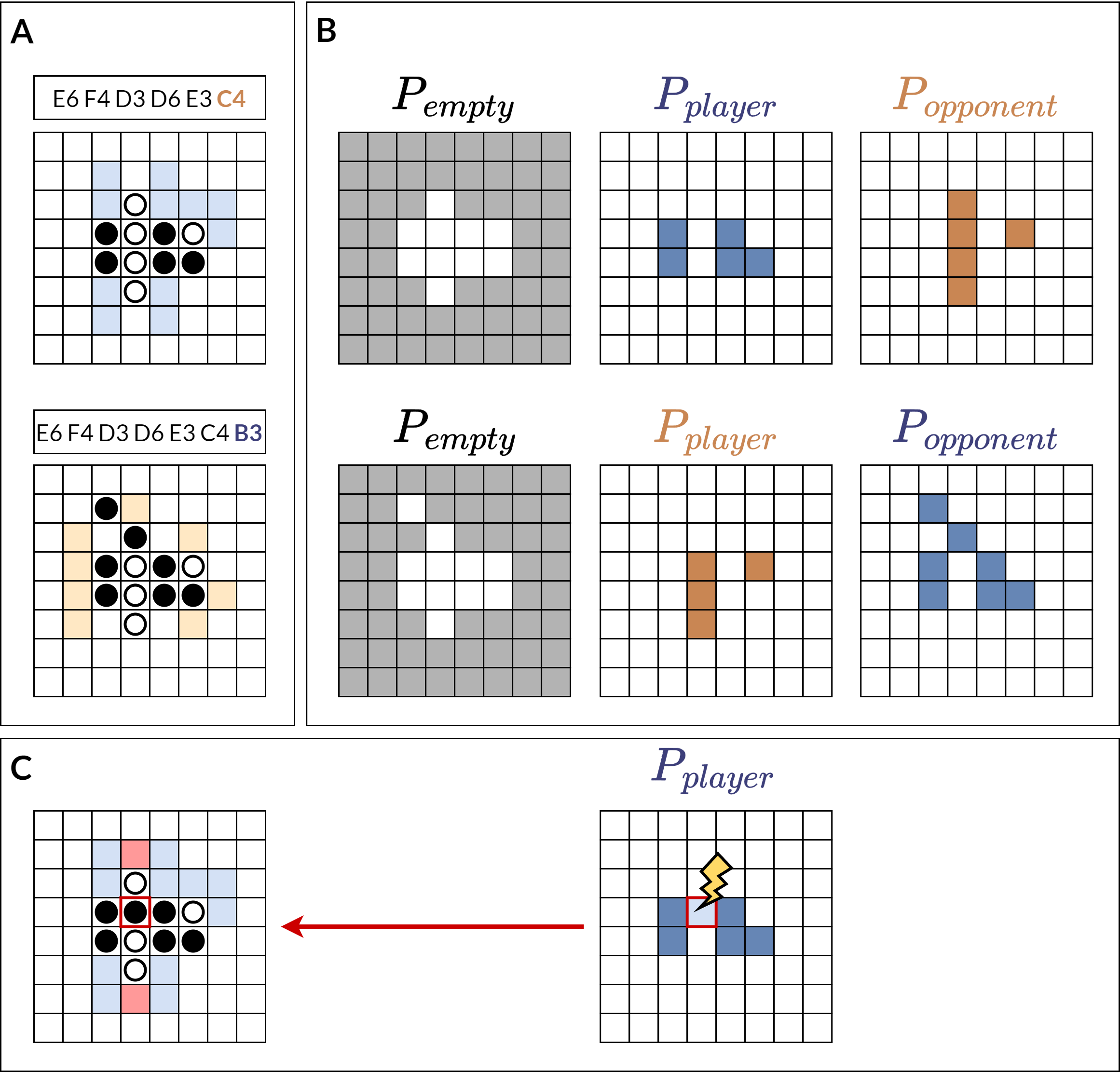} 
    \caption{\textbf{Emergent representations of the board state in Othello-GPT} (adapted from \cite{nandaEmergentLinearRepresentations2023}). The model represents board states relative to the current player. \textbf{A.} The ground truth board states at two consecutive time steps. Colored tiles show legal moves for the current player (light blue for \textsc{black} and light orange for \textsc{white}). \textbf{B.} The board states at the same two time steps decoded by linear probes. The probes are trained to classify the board relative to the current player: \texttt{empty} for empty tiles, \texttt{player} for tiles occupied by the current player, and \texttt{opponent} for tiles occupied by the current player's opponent. Note that the \texttt{player} and \texttt{opponent} colors are flipped between the two time steps, as the current player changes. \textbf{C.} We can intervene on the model's internal board state representation by pushing an empty tile's representation in the direction of the \texttt{player} vector, to make the model represent that tile as occupied by the current player. This simple linear intervention is sufficient to alter Othello-GPT's move predictions (red tiles on the left), demonstrating that the linearly represented board state causally determines the model's outputs.}
    \label{fig:othello-gpt} 
\end{figure}

In a follow-up study, \cite{nandaEmergentLinearRepresentations2023} discovered that rather than representing the board state (e.g., representing each board tile as \texttt{black}, \texttt{white}, or \texttt{empty}), Othello-GPT actually encodes tiles relative to the current player (as \texttt{player}, \texttt{opponent}, or \texttt{empty}). By re-orienting probes to classify this player-centric representation, Nanda et al. demonstrated that the board state is in fact \textit{linearly} encoded with high accuracy in the network, contrary to \cite{liEmergentWorldRepresentations2023}'s claim that the board state is only encoded non-linearly (fig. \ref{fig:othello-gpt}). They further demonstrated behavioral control by conducting simple vector arithmetic interventions to alter the model’s encoding of board states and change predictions accordingly. \cite{hazinehLinearLatentWorld2023} found similar evidence that information about board state is encoded in a simple, linear way in the deeper layers of Othello-GPT models. Like \cite{nandaEmergentLinearRepresentations2023}, they decoded a representation corresponding to tiles marked \texttt{player}, \texttt{opponent}, or \texttt{empty}, which aligns well with the role of the model in alternating between playing as white or black. To test whether these internal representations play a causal role in the model's predictions, they also intervened by manipulating activations to trick the model about the state of the board. Through visualizing effects on predicted logits and comparing distributional similarity of logit outputs, they demonstrated layers in which this internal representation steers next-move predictions. The internal representation appears fully developed and utilized in middle layers of deeper models, while shallow models fail to use the representation causally. 

These findings align with the \textit{linear representation hypothesis}, according to which high-level concepts or features are represented linearly as directions in a neural network's activation space. This hypothesis is often taken to be central to the agenda of mechanistic interpretability \citep{elhage2022superposition}. Indeed, if features correspond to directions, we can in principle extract and understand them by projecting activations onto those directions. This should allow us to decompose the high-dimensional activation space of the network into interpretable components. We can also check whether the weights of downstream circuits align with a feature vector and thus conclude that the model is using this feature. Linear directions provide a global foothold across the whole model to reason about a specific feature. Furthermore, manipulating model behavior via causal intervention becomes significantly simpler when features are linear directions. We can predict model behavior under counterfactual input features by pushing model activations along feature directions. In the case of Othello-GPT, we can linearly decode information about the state of the board from activations – namely, which tiles are occupied by the current player's pieces as opposed to the opponent's pieces, and which tiles are empty. The fact that this information is linearly decodable in a given layer makes it easy to manipulate with simple vector arithmetic by pushing activations along feature directions. It should be noted that the linear representation hypothesis can be spelled out in slightly different ways, depending on how one technically defines what it means for a concept or feature to be linearly represented in the network \citep{parkLinearRepresentationHypothesis2023}. In addition, there is room for disagreement about the theoretical grounding of this hypothesis. While linear decodability does make interpretability easier, this does not entail that complex neural networks do not also encode information non-linearly and cannot make sure of such information in downstream processing.

In Part I, we defined world models in the context of LLMs as internal representations of the world that allows them to parse and generate language that is consistent with real-world knowledge and dynamics \citep{millierePhilosophicalIntroductionLanguage2024}. Studies on Othello-GPT provide tentative evidence that Transformer models can acquire world models in this sense, at least in toy domains.\footnote{For example, similar results have been found by studying Chess-GPT, a Transformer model trained on chess moves \citep{karvonenChessGPTInternalWorld2024}.} While Othello-GPT is not a language model properly speaking, it generates legal game moves in written form, and interventionist experiments reveal that it generates such moves on the basis of linearly decodable representations of the board state. To what extent can we generalize these findings to actual LLMs? This is a challenging question. A reasonable assumption is that Othello-GPT acquires a world model – an internal representation of the game world – because this is useful to improve its predictions of legal moves past a certain threshold. However, neural networks are notorious in their ability to achieve high accuracy scores on problems by mastering surface statistics; they tend to learn shortcuts (e.g., shallow heuristics) to achieve good performance on their learning objective until they hit a bottleneck. Furthermore, acquiring world models about the real world, rather than an extremely simple game world, is presumably costly in terms of training dynamics; it would require a substantial reorganization of the model's internal structure, akin to the phase transition undergone by \cite{nandaProgressMeasuresGrokking2022}'s model to `grok' modular addition, albeit on a much broader scale. 
 
It may well be that LLMs do undergo such phase transitions during training, and acquire representations akin to world models at least in some limited domains. This might be required to unlock certain abilities such as commonsense reasoning about intuitive physics, which in turn would further reduce the loss (i.e., improve next-token prediction performance) in certain contexts. For this to happen, two conditions must presumably be satisfied: (a) the text-based training data should provide enough information to induce the relevant components of world models; (b) the learning pressure to reduce loss on the next-token prediction objective should be sufficient to push the model to acquire such representations given long enough training trajectories. Neither condition is antecedently guaranteed, and there is currently limited evidence that they apply to most existing LLMs. In particular, interventionist evidence regarding the most capable LLMs such as GPT-4, whose behavior on certain tasks is most consistent with the acquisition of world models, is still sorely lacking.\footnote{This is partly due to the fact that the weights of state-of-the-art LLMs are not publicly released, which precludes interpretability research from anyone other than researchers in the team that trains them. We will come back to this issue in section \ref{sec:secrecy}.} As far as behavioral evidence is concerned, lingering failure modes on out-of-distribution tasks cast doubt on the hypothesis that current LLMs acquire sophisticated and robust world models \citep{mccoyEmbersAutoregressionUnderstanding2023,yildirimTaskStructuresWorld2023}.

There are nonetheless ongoing efforts to assess the existence of world models in LLMs using the tools of mechanistic interpretability. For example, \cite{liImplicitRepresentationsMeaning2021} used probing and intervention methods on models fine-tuned on the Alchemy and TextWorld datasets. Alchemy contains sequences of instructions for manipulating colored liquids in beakers (i.e., instructions for performing fictional chemical experiments and their outcomes), while TextWorld consists of textual transcripts of navigation in simulated worlds. Li et al. designed probes to test if the models' contextual token embeddings encode and track the state of entities mentioned in the discourse. For example, in Alchemy the probe tries to determine if a representation encodes that a beaker is empty after its contents are drained. Li et al. used these probes to test whether intervening on the decoded entity representations would change model behavior. Specifically, they constructed two discourses $x_1$ and $x_2$ that describe draining liquid from two different beakers, $b_1$ and $b_2$, resulting in one empty beaker per discourse. After encoding each discourse, they created a synthetic representation $C_{mix}$ by taking the encoding $C_1$ of discourse $x_1$ and replacing the vector representations corresponding to the initial description of beaker $b_2$ with those from $C_2$. Although $C_{mix}$ does not correspond to any real textual input, it implicitly represents a situation in which both $b_1$ and $b_2$ are empty. When generating text conditioned on $C_{mix}$, they found that the model generates instructions that are more often consistent with both beakers being empty compared to generating from $C_1$ or $C_2$ alone. However, the generated instructions from $C_{mix}$ are still not always fully consistent with the implicit state, suggesting the induced representation is approximate rather than perfect. By editing the entity representations to model a new state not seen in the actual training data or prompt, and observing changes in the model's outputs that are often consistent with this new state, Li et al. provided tentative evidence that the model can induce an approximate implicit representation of the state of discourse entities purely from text. Note that producing text that is consistent with the hypothesis-driven manipulation of the LLMs activation space is not something that the system was trained to do. Furthermore, the method was found to be robust against `sanity checks' to confirm that the intervention itself is not wholly responsible for injecting appropriate information into the system, such as finding negative results when attempting the intervention method on similarly-complex Transformer architecture with randomized weights. However, the generality of these findings remains uncertain, as the experiments focused on narrow domains with simple objects and dynamics. More systematic testing would be needed to determine how well they generalize to more complex and open-ended settings. 

\subsection{Interpretability and causal abstraction}

The project of mechanistic interpretability can be seen through the lens of causal abstraction \citep{geigerCausalAbstractionsNeural2021}. Causal abstraction is a theoretical framework that aims to provide an interpretable high-level causal explanation of the behavior of a complex system, such as a neural network, that is consistent with the low-level causal structure of that system.

The core intuition is that the variables of a low-level causal model can be clustered into sets and aligned with the variables of a high-level causal model. The high-level model provides an abstract characterization of the low-level model if aligned high-level and low-level variables have equivalent causal roles and information content. This equivalence is experimentally verified with interventions on both levels that produce the same counterfactual behavior. More formally, an alignment between a low-level model $\mathcal{L}$ and high-level model $\mathcal{H}$ is a partition of the low-level variables into sets, with each set aligned to a high-level variable. A translation function maps low-level variable values to high-level values \citep{geigerCausalAbstractionFaithful2023}. The alignment is causally consistent if, for any low-level intervention that has a corresponding high-level intervention under the translation function, intervening on both models produces equivalent results after translation. If an alignment is causally consistent, then $\mathcal{H}$ is said to be a constructive causal abstraction of $\mathcal{L}$.

Any program or algorithm can be represented as a causal model \citep{icardProgramsCausalModels2017}. In the context of interpretability research in deep learning, the low-level causal model is a neural network, containing many interconnected nodes and weights that determine its function. The high-level causal model proposed by mechanistic interpretability researchers offer hypotheses about the abstract computations and algorithms implemented by the network. Aligning groups of neurons and weights to single variables in the high-level model allows interpreting their collective causal role. The causal consistency condition ensures the high-level causal model faithfully captures the causal mechanisms embodied by the low-level neural network model. Causal abstraction thus enables the development of human-interpretable high-level causal models that accurately explain the reasoning and computations inside opaque neural networks.

A key method for assessing causal abstraction is interchange interventions, where neural representations created for one input are swapped into the model when it is processing another input. If the low-level neural model and high-level algorithm have the same counterfactual behavior under aligned interchange interventions, that provides evidence that the alignment witnesses a causal abstraction. Many of the methods described above, such as activation patching, can be used as interchange interventions. In fact, iterative \gls{nullspace} projection can also be formalized within a causal abstraction framework \citep{geigerCausalAbstractionFaithful2023}. Thus, causal abstraction is a useful framework to unify interventionist approaches to interpretability.

In practice, the project of mechanistic interpretability with large neural networks like LLMs can be seen as aiming for \textit{approximate} causal abstraction, which relaxes the criteria for alignment \citep{beckersApproximateCausalAbstractions2020}. The degree of abstraction can be quantified as the proportion of aligned interchange interventions that have equivalent effects in the high-level causal model and low-level causal model or target neural network \citep{geigerCausalAbstractionFaithful2023}. When interchange intervention accuracy is 100\%, the high-level model exactly abstracts the neural network. Otherwise, the high-level model approximately abstracts the neural network approximately, to the degree quantified by interchange intervention accuracy. Approximate abstraction can provide an interpretable high-level explanation that still reflects the causal structure of the target neural network with a degree of faithfulness suitable for genuine explanation.

\subsection{Biological plausibility of decoded computations}

Even if the preceding evidence regarding the complex and relevant functional roles attributed to LLM activation vectors are accepted, readers may be frustrated at the Transformer architecture's apparent lack of biological inspiration or plausibility. Whereas other deep learning architectures like DCNNs for computer vision were substantially inspired by biological investigations into cortical tissues and have since even been further aligned with data from functional neuroscience (Buckner 2019), Transformers arose almost independently of the neuroscience of language processing. Moreover, the \gls{keyqueryvalue} division on which self-attention depends was inspired by an analogy to database theory in computer science rather than theory in psychology or cognitive science. What, we might wonder, could possibly correspond to the mathematical operations of self-attention in the brain, even at a gross functional grain of analysis?

Nevertheless, numerous imaging and alignment studies have suggested that some families of Transformers are also excellent predictors of activation patterns in human language processing areas, such as \cite{schrimpfNeuralArchitectureLanguage2021,caucheteuxGPT2ActivationsPredict2021}. Such alignment is defended even by researchers who concede that state-of-the-art Transformers like GPT-4 nevertheless lack substantial amounts of grounded world knowledge and other kinds of understanding often attributed to language comprehension, such as of complex semantic intentions or discourse representations. This has led some researchers to argue that such alignment demonstrates that left temporal areas of the brain associated with linguistic processing may play a limited word prediction role, arguing for a narrower construal of the language faculty than has traditionally attributed to such areas in cognitive neuroscience \citep{mahowaldDissociatingLanguageThought2023}. Others have been led to draw even more ambitious comparisons between the architecture of self-attention and the brain; for example, \cite{whittingtonRelatingTransformersModels2022} have argued that operations hypothesized to be implemented by medial temporal lobe tissues in computational neuroscience -- specifically, place and grid cell coding between the entorhinal cortex and hippocampus – can be seen as mathematically equivalent to self-attention operations in Transformers. Whether this abstract mathematical equivalence supports closer mechanistic alignment between Transformers and human neural tissues remains to be seen, but may yet be seen to soften some of the initial skepticism that such alignment is outright implausible.

\section{Newer philosophical questions} \label{sec:newer-philosophical-questions}

The previous section summarized some of the strongest mechanistic evidence to date that current-generation LLMs can, at least in principle, acquire `world models' – structure-preserving representations of the state of the world described in text inputs that are causally efficacious on the model's ability to generate text outputs consistent with that world state. Overall, this evidence weighs against what we characterized in Part I as our null hypothesis, the claim that LLMs can be adequately described as giant lookup tables that merely retrieve sequences memorized during training. Nevertheless, traditional LLMs limited to next-token prediction over linguistic inputs have significant limitations that preclude naive comparisons to human or animal cognition. Research on generative modeling is fast-paced, however. Newer approaches have already moved past traditional LLMs, enhancing their architecture with multimodal capabilities and/or modular designs in which the text-generating aspect is simply one component of a more complex system. In this section, we address newer philosophical questions that may arise from these recent developments in LLM research. We also explore more difficult questions raised by LLM's unprecedented abilities (such as whether they meet tentative criteria for consciousness), as well as new worries about their scientific legitimacy precipitated by their unprecedented scale and the secrecy under which they are developed.

\subsection{LLMs and modular architectures}

LLMs exhibit impressive performance on traditional natural language processing tasks, where they have increasingly superseded task-specific models. However, there is only so much one can expect from a system exclusively trained to predict sequences of linguistic tokens. One promising idea to expand the capacities of LLMs is to augment or integrate them with components subserving other basic faculties -- such as perception, imagination, and planning. There are two ways to execute this vision. The first involves modifying the Transformer-based architecture of traditional LLMs to allow them to process inputs in multiple modalities (e.g., text and images). The second involves incorporating LLMs as modules within a composite system, which may include multiple neural networks as well as classical symbolic algorithms. Each of these strategies has their own benefits and tradeoffs, but they are not mutually exclusive; a vision-language model designed to process images and text can itself be integrated with other module. 

Language modeling plays a key role in these experiments. When Transformer architectures are augmented to receive visual input in addition to linguistic input, for example, their abilities to parse images is critically informed by linguistic concepts and dependencies induced from next-token prediction. By contrast, when traditional text-only LLMs are integrated within an ensemble of models, language acts as a common ground or `universal API' for these models to communicate with each other without sharing differentiable parameters \citep{zengSocraticModelsComposing2022}.\footnote{An API (Application Programming Interface) is a set of rules and protocols that allows different software components to communicate and interact with each other. By analogy, language can serve as a `universal API' for generative models by providing a common interface to exchange information without being part of the same neural network.} In either case, language modeling forms the backbone of more capable systems.

Efforts to move beyond text-only LLMs have converged in around two trends in deep learning research: multimodality and so-called `agent' systems. In the rest of this section, we will discuss each of these trends and the philosophical questions they raise.

\subsubsection{Multimodality}

While the Transformer architecture was initially applied to language modeling tasks, it has been adapted and expanded to tackle other modalities. For example, Vision Transformers can process visual inputs through a clever trick that involves splitting images into `patches' that can be fed to the model as sequential tokens, as one would with linguistic tokens \citep{dosovitskiyImageWorth16x162021}. Subsequent research efforts have explored different ways to hybridize the Vision Transformer architecture with other architectures to leverage their respective strengths. 

In the Taming Transformers architecture for image generation \citep{esserTamingTransformersHighResolution2021}, the authors created a modular architecture in which raw images were first transformed into abstract feature maps by a deep convolutional neural network; those abstracted feature maps were then passed to a Transformer for encoding, and finally passed to a generative deep neural network for image reconstruction. The authors of this paper argue that their system can combine the ability of deep convolutional neural networks to extract abstract local features and the ability of Transformers to recognize long-distance compositional dependencies in inputs. As they put it,

\begin{quote}
Our key insight to obtain an effective and expressive model is that, taken together, convolutional and Transformer architectures can model the compositional nature of our visual world: We use a convolutional approach to efficiently learn a codebook of context-rich visual parts and, subsequently, learn a model of their global compositions. The long-range interactions within these compositions require an expressive Transformer architecture to model distributions over their constituent visual parts. \cite[][p. 12874]{esserTamingTransformersHighResolution2021}
\end{quote}

This combination of strengths allowed them to achieve significant performance gains at the time over other generative architectures like \glspl{GAN}, especially regarding holistic and relational aspects of an image's composition. Notably, the transformer's strengths in these regards can help neural network models address some classical criticisms of the previous generation of artificial neural networks, as noted in Part I \citep{millierePhilosophicalIntroductionLanguage2024}; such modular networks demonstrate that these benefits can be provided to other domains like image labeling and generation as well.

An entire generation of multimodal models has been based on the insight that representing linguistic and visual information in a joint latent space is helpful to parse the structure of images. For example, OpenAI's CLIP – which stands for Contrastive Language–Image Pre-training – learns from a massive dataset comprising millions of image-text pairs, enabling it to predict the most relevant text description for a given image without being explicitly optimized for this task \citep{radfordLearningTransferableVisual2021}. CLIP has two main components: an image encoder and a text encoder. These encoders transform images and text into a shared high-dimensional space where linguistic and visual representations can be compared directly. The image encoder is typically a modified version of a Vision Transformer that encodes images into vectors, and the text encoder is also a Transformer model similar to LLMs that encodes text snippets into vectors in the very same vector space. CLIP learns these vector representations through \textit{contrastive learning}, which involves contrasting similar and dissimilar pairs of images and captions. Each training step involves a batch of image-caption pairs that are encoded into vectors by their respective encoders; CLIP's contrastive learning objective consists in maximizing the similarity of image vectors with the text vectors corresponding to their matching captions, while minimizing the similarity of image vectors with the text vectors corresponding to non-matching captions. After training, CLIP can parse and categorize images it has never seen before. For example, it can accurately identify objects, actions, or scenes in new images based on the learned associations between images and text during training. Beyond simple classification, CLIP can be used to generate new captions for images or find images that match a given caption.

CLIP was a stepping stone towards multimodal models that descend from LLMs, known as vision-language models (VLMs), which can receive both images and text as input and generate text like as LLMs \citep{alayracFlamingoVisualLanguage2022}. VLMs can answer question about images, and the best of them demonstrate a sophisticated ability to parse visual content. For example, GPT-4V, the VLM version of GPT-4, achieves strong performance on image captioning and visual question answering benchmarks, and shows some ability for commonsense reasoning about visual inputs \citep{openaiGPT4VIsionSystem2023,yangDawnLMMsPreliminary2023,wuEarlyEvaluationGPT4V2023}.

Systems that can generate images from text description, like Stable Diffusion \citep{rombachHighResolutionImageSynthesis2022} or DALL-E \citep{rameshHierarchicalTextConditionalImage2022}, also build on the same technical foundation. CLIP's text encoder plays a critical role in these systems, by converting text descriptions into vectors that can be used to condition the image generation process. Images are generated by iteratively denoising latent representations – a process known as \textit{latent denoising diffusion}. At each step of the denoising process, the model considers both the current state of the image being generated, and the semantic direction provided by the vector from CLIP's text encoder. This ensures that the generated images are not only visually coherent but also semantically aligned with the text prompt. In other words, cross-modal alignment between the linguistic and visual domains achieved through the encoding of text descriptions in a shared latent space is fundamental to image generation models' ability to generate outputs whose content and composition adhere to the semantic and syntactic structure of the input. In fact, text-space encodings can serve as the common medium for breakthrough results in a variety of other modalities, include text-to-audio, text-to-code, and text-to-action \citep[for a review, see][]{gozalo-brizuelaSurveyGenerativeAI2023}.

There is a direct lineage between the Transformer architecture of LLMs, the Transformer architecture of CLIP's text encoder, and both advanced VLMs like GPT-4V and advanced text-to-image generation models like DALL-E. Nevertheless, seemingly subtle differences in architecture and training objective matter greatly to the performance of these systems. For example, VLMs and image generation models based on CLIP's contrastive text encoder have been found to suffer from significant limitations in their ability to process the compositional structure of text prompts and images appropriately. In particular, these models often have difficulty correctly associating attributes with their corresponding objects when multiple objects are present in an image. They also struggle to accurately interpret spatial prepositions describing relative positions of objects, properly account for specified numbers of objects, and handle negation terms that exclude certain attributes or objects from a scene \citep{zhangEvaluatingCLIPUnderstanding2023,hsiehSugarCrepeFixingHackable2023,tongMassProducingFailuresMultimodal2023,kamathWhatVisionlanguageModels2023,lewisDoesCLIPBind2023,murphyComparativeInvestigationCompositional2024}. 

However, these failures should be interpreted carefully. They do not seem symptomatic of a general limitation of the Transformer architecture; as discussed in Part I, text-only Transformer-based LLMs are much better at compositional generalization \cite{millierePhilosophicalIntroductionLanguage2024}. Rather, the compositional failures observed in many multimodal models can be attributed in large part to limitations of the contrastive learning objective used to train their text encoders \citep{yuksekgonulWhenWhyVisionLanguage2022,kamathTextEncodersBottleneck2023}. Contrastive learning trains the model to match images to captions in a way that does not require preserving detailed compositional and syntactic information from the text. In essence, text encoders trained in this way treat linguistic inputs like `bags of words', discarding critical aspects of sentence structure and word order. In fact, recent work has shown that replacing the contrastively-trained text component of a multimodal model with one that has instead been pre-trained on image captioning can significantly improve the model's compositional abilities \citep{tschannenImageCaptionersAre2023}. Image captioning models can excel at understanding attributional and relational information in text compared to CLIP-style contrastive models -- outperforming the latter by large margins on benchmarks that test sensitivity to word order, object attributes, and spatial relations. Importantly, these models all use a Transformer architecture. The key difference is in how the text component is trained, either with a contrastive objective that ignores detailed sentence structure, or an image captioning objective that preserves it. The fact that simply changing the training objective while holding the architecture type fixed leads to such dramatic differences in compositional performance provides compelling evidence that the Transformer itself is not the root issue. This is a useful cautionary tale about the interpretation of failure modes in deep learning.

\subsubsection{`Agent' systems}

Influential theorists like \cite{goyalInductiveBiasesDeep2022} have speculated that integrating Transformer-based language models in composite systems can provide other benefits to other modules. For example, they speculate that common causal variables tend to correspond to words in natural language, and so a language model that biased an image classifying module or a reinforcement-trained agent module would be more likely to focus on causally-relevant features of its training set (which they summarize with the hypothesis ``that semantic variables are also causal variables'', \citet[][p. 14]{goyalInductiveBiasesDeep2022}). These variables include words referring to agents or subjects of sentences, actions (often captured by verbs), objects of agents' actions (often captured by noun phrases in the direct object position), and modalities or properties of objects (often captured by adjectives and adverbs). From this perspective, natural languages are a kind of cultural technology that help us focus on the most important aspects of our environment \citep[cf.][]{clarkMagicWordsHow1998}, and systems that draw on language models to label incoming sensor data and formulate plans of action would thus be more likely to focus on causally relevant and robust properties in the environment and less likely to focus on artifacts and chimeras.

Many such architectures are already being implemented by industry and academic research groups, often inspired by the old Vygotskyan hypothesis in cognitive science that simulated inner speech is a powerful medium of thought that reliably develops in humans between the ages of 3-6, when children begin to demonstrate distinctively human success on problems of causal, logical, and social inference \citep{vygotskyThinkingSpeech1987,carruthersCognitiveFunctionsLanguage2002,lupyanChapterSevenWhat2012,colasLanguageCognitiveTool2021}. To give an early example of such a system, Google Robotics has implemented an `inner monologue' agent \citep{huangInnerMonologueEmbodied2022} that integrates a scene descriptor, success detector, and planning module together with a human interlocutor in a closed feedback loop. This agent can use a language model to interpret instructions given to it by a human, label visual input from sensors, formulate internal plans in terms of a constrained action language, and verify the success or failure of plan components by deploying a success detector on sensor input. Even more ambitious agent architectures deploying internal monologues are possible, such as using text encodings to drive internal simulations of expected perceptual output, which could then be used for long-term planning and counterfactual `imagination' of possible scenarios -- such as text-driven planning rollouts of the sort seen in DeepMind's Imagination-Augmented Agents (I2A) architecture \citep{racaniereImaginationAugmentedAgentsDeep2017,bucknerDeepLearningRational2023}.

These developments can be seen as specific steps on a broader journey from narrowly language-focused Transformer models to multi-modal Transformer-based `agents', which many theorists see as the natural next step in deep-learning-based artificial intelligence. According to this frame, LLMs were an important breakthrough in making DNNs more compositional and in particular allowing them to tackle language-scaffolded effects in thought and action, but they are only one component in a full agent architecture. Moreover, the standard context of deployment for LLMs – continual next-word completion – lacks the stability and impetus of rational agency. This paradigm requires massive, ecologically-invalid datasets or at least deliberately-harvested instruction-tuning or reinforcement learning, and standard LLMs cannot gather their own ongoing training data by interacting with the world directly. Those trying to build Transformer-based agents seek to address these shortcomings by embedding LLMs in broader architectures and decision loops that allow them to generate their own goals (so-called `autotelic' agents, \citealt{colasAutotelicAgentsIntrinsically2022}), formulate plans to pursue them, verify their success using a variety of modalities, and learn from continual self-driven exploration of a world.

\begin{figure}[h!] 
    \centering 
    \includegraphics[width=1\linewidth]{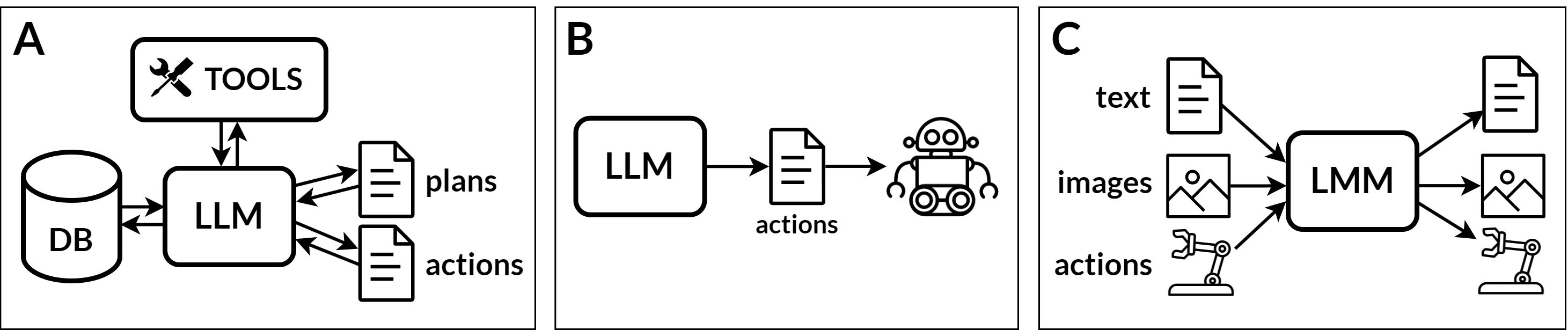} 
    \caption{\textbf{Three kinds of agent systems based on LLMs or multimodal models.} \textbf{A.} A modular language agent system compose of a LLM that can interacts with (i) an external database (DB) to store and retrieve long-term knowledge or `memories,' (ii) text files to store and retrieve plans and actions, and (iii) external tools that can be used through function calls. \textbf{B.} A robotic system that makes use of a LLM to translate high-level natural language instructions into low-level policies to plan and act in the real world. \textbf{C.} A large multimodal model (LMM) that can not only receive and produce text, but also images and even action tokens to control a robotic body.}
    \label{fig:agents} 
\end{figure}

There are three different kinds of `agent' systems that build on the success of standard LLMs: (a) language agents that make use of external databases or tools; (b) embodied models that control a robotic body from natural language instructions with an LLM; and (c) multimodal models trained end-to-end to process actions in addition to text (Fig. \ref{fig:agents}).

Language agents are systems with specialized modules orchestrated by calls a LLM (see \citealt{wangSurveyLargeLanguage2023} for a review). These modules may include a memory component, a planning component, and an action component. The memory component extends the input context window of the LLM, which acts as a short-term memory buffer, with an external long-term storage (like a text file or a structured database). For example, Reflexion combines a sliding context window with persistent memory banks \citep{shinnReflexionLanguageAgents2023}. The LLM can write to and read from the external memory, as well as summarize multiple memories into high-level insights to inform planning and action. A planning component can structure the reasoning process to accomplish goals. For example, ReAct utilizes action-observation cycles to refine plans based on environmental outcomes \citep{yaoReActSynergizingReasoning2023}. Finally, an action component can interpret the agent's decisions into concrete outputs and behaviors.

A striking example of this approach is the `generative agents' of \cite{parkGenerativeAgentsInteractive2023}. The architecture has three main components: the memory stream stores a comprehensive natural language record of the agent's experiences, with a retrieval model combining relevance, recency and importance to surface the most pertinent records; `reflection' synthesizes memories into higher-level inferences that guide behavior over time; and planning translates conclusions from reflection and the current environment into high-level action plans that are recursively decomposed into detailed behaviors. Park et al. demonstrate the effectiveness of this approach in a Sims-like sandbox game environment called Smallville, inhabited by 25 unique language agents. These agents exhibit emergent social behaviors like information diffusion, relationship formation, and coordination (e.g. for spontaneously organizing a Valentine's Day party), based solely on initial seed prompts. Another impressive example is Voyager, a language agent powered by GPT-4 that is able to continuously explore, develop skills, and make discoveries in the open-ended 3D world of Minecraft without any human intervention \citep{wangVoyagerOpenEndedEmbodied2023}. The key components of Voyager consist of an automatic curriculum module that proposes a stream of progressively more complex tasks optimized for exploration; a skill library for developing and composing complex, reusable behaviors; and an iterative prompting mechanism that leverages execution feedback and self-verification to iteratively refine GPT-4's code generation until a task is successfully completed.

A concurrent trend consists in augmenting LLMs with external tools that can be manipulated through generated code or calls to APIs. One strategy consists in fine-tuning LLMs to enable them to learn in context when API calls are helpful in an entirely self-supervised manner, requiring only a handful of demonstrations per tool \citep{schickToolformerLanguageModels2023}. Augmenting LLMs with external storage and tools effectively creates hybrid neurosymbolic systems that use language and/or code as a universal interface between connectionist and symbolic components. In turn, this goes a long way towards addressing some remaining limitations of pure LLMs; for example, using tools such as a code interpreter or Wolfram Alpha significantly improves the performance of LLMs on challenging math problems \citep{zhouSolvingChallengingMath2023,davisTestingGPT4Wolfram2023}.

The second category of agent systems encompasses regular LLMs embedded in a system that include a physical robotic body roaming the real world. This approach combines the linguistic competence of LLMs with pre-trained robotic skills without requiring special modifications to the LLM component itself. For example, SayCan takes as input a high-level natural language instruction that describes a task for the robot to perform \citep{ahnCanNotSay2022}. It then scores candidate low-level skills based on the probability that the skill's description is relevant for the high-level goal -- as determined by the LLM component -- and the probability from a learned value function that the skill can successfully execute in the current state. By multiplying these probabilities, the method selects skills that are both useful for the goal and possible to achieve. The process repeats by appending selected skills until termination. SayCan can be implemented in a robot with vision-based manipulation skills trained via behavioral cloning and reinforcement learning, to perform real-world long-horizon tasks in a complex environment such as a kitchen. 

The third category of agent system consists in Transformer-based models trained to receive multimodal inputs (including text) and able to produce actions as outputs. Unlike text-only LLMs embedded in robotic systems, these models are trained end-to-end to generate action commands. DeepMind's Gato \citep{reedGeneralistAgent2022} is designed to be a `generalist agent' that operates in a multi-modal, multi-task architecture on a generalist learning policy -- which allows the same system to play Atari games, caption images, chat with humans, manipulate a real robot arm to stack blocks, and much more. In a similar spirit, \cite{zitkovichRT2VisionLanguageActionModels2023} propose a method for incorporating Transformer-based vision-language models directly into robotic manipulation policies. The key idea is to represent robotic actions as special text tokens that can be generated as part of the model's text output. The resulting model, RT-2, is pre-trained on image, text and action captioning datasets scraped from the internet, then fine-tuned on robotics datasets to output tokenized actions. The unified text-based output space of RT-2 is a key advantage over prior works, enabling it to successfully generalize to novel instructions after training.

The evolution of LLMs towards `agent' systems capable of executing instructions, forming plans, and performing actions in the real world raises interesting philosophical questions. One pertains to the grounding problem: are embodied, LLM-based agents more likely to meet the requirements for referential grounding? \cite{molloVectorGroundingProblem2023} provide reasons to doubt that this is the case for systems, like SayCan, that simply make calls to a distinct text-based LLM component -- where the latter is not pre-trained or fine-tuned on world-involving tasks. However, systems trained end-to-end to generate actions like Gato or RT-2 might deserve a different treatment. As Transformer-based models expand beyond text-only training to process any information that can be serialized -- including images, videos, joint torques, button presses, etc. -- it becomes less obvious that they should be deemed intrinsically incapable of inducing normative world-involving functions, even if they are not explicitly fine-tuned from human feedback. These systems also raise obvious question about agency itself: is the term `agent' (let alone `autonomous agent') a misnomer when it comes to systems that do not fundamentally have intrinsic goals or motivations? The answer largely depends on how thick one's notion of agency is. LLM-based language agents are often described as generating, storing and retrieving goals, as well as breaking them down into sub-goals and individual actions. However, their behavior remains ultimately grounded in a human-written prompt. As such, they fail to meet the requirements for pure agential autonomy.

\subsection{Consciousness} \label{sec:consciousness}

LLMs can eloquently converse about any topic sufficiently prevalent in their training data -- including consciousness. Indeed, they are perfectly capable of generating compelling first-person reports that look like self-ascriptions of subjective experiences. In humans, such reports are generally taken at face value; while introspection can be unreliable \citep{irvineDevelopingDarkPessimism2021}, verbal reports typically provide (defeasible) evidence about someone's experience. This intuition breaks down when it comes to LLMs. No matter how convincing their fluent mimicry of experience reports might be, it cannot be taken as prima facie evidence that they are conscious. Imitating patterns of human language use is precisely what LLMs are trained to do. Nonetheless, if LLMs can acquire emergent capabilities in the process of learning from a next-token prediction objective, then we ought to wonder whether current or future systems with a similar architecture could, in principle, acquire the capacity for conscious experience. Like other debates about the putative cognitive capacities of LLMs, this issue cannot be settled merely through armchair speculation, but calls for careful empirical research informed by mature theories.

While consciousness may seem like a particularly high bar to clear for machines that still clearly lack some of the hallmarks of human intelligence, it is worth nothing that it need not correlate with sophisticated cognitive abilities. Indeed, there is a rather broad consensus that many non-human animal species that presumably lack such abilities have the capacity for consciousness -- not only mammals \citep{sethCriteriaConsciousnessHumans2005}, but more largely vertebrates including reptiles, birds, and fish \citep{cabanacEmergenceConsciousnessPhylogeny2009,merkerLiabilitiesMobilitySelection2005}, and perhaps even invertebrates including mollusks and arthropods \citep{barronWhatInsectsCan2016,godfrey-smithMetazoaAnimalLife2021}. Unlike animals, however, AI systems do not share a common evolutionary history or physical substrate with humans. As such, considerations about phylogeny or similar neurophysiological features that are central to the discussion of animal consciousness do not straightforwardly apply to LLMs.

The hypothesis that artificial neural networks running on silicon chips could be conscious is premised upon computational functionalism -- the view that what makes a physical system conscious is not its particular physical makeup, but whether it implements an appropriate set of computations \citep{chalmersAbsentQualiaFading1995}. If computational functionalism is true, biological brains and computers could in principle implement the same consciousness-enabling computations using different physical machinery. This assumption is not uncontroversial. Biological nervous systems exhibit complex dynamics involving global electrical patterns that modulate fine-grained neural signaling. Biochemical mechanisms and metabolic processes in the brain may impose specific constraints on the realization of conscious mental states \citep{caoMultipleRealizabilitySpirit2022,godfrey-smithMindMatterMetabolism2016}. Artificial neural networks implemented on digital computing hardware bear little resemblance to biological nervous systems at this level of granularity, and it may be that the kind of silicon substrates used by modern computers cannot support consciousness even in principle.

If computational functionalism holds, however, then it should be possible in principle to identify the computational correlates of consciousness in biological brains, and assess whether similar computations are implemented in systems like LLMs. Given the lack of a complete and widely accepted scientific theory of consciousness, there is currently no definitive test or criteria to determine if an AI system is conscious. It is nonetheless possible to make qualified claims about the plausibility of consciousness in AI systems based on computational markers of biological consciousness derived from cognitive neuroscience research \citep{aruFeasibilityArtificialConsciousness2023,butlinConsciousnessArtificialIntelligence2023,ledouxConsciousnessHumanCase2023}.

\cite{butlinConsciousnessArtificialIntelligence2023} survey prominent neuroscientific theories of consciousness that are compatible with computational functionalism. From these theories, they derive a list of `indicator properties' -- features or mechanisms that the theories associate with consciousness. The more indicator properties an AI system exhibits, the more likely it is to be conscious. Prominent theories of consciousness include, among others: recurrent processing theory, which proposes that recurrence within neural networks distinguishes conscious from unconscious processing \citep{lammeTrueNeuralStance2006}; global workspace theory, which claims that consciousness arises when information is broadcast to widespread networks from a limited-capacity workspace \citep{baarsCognitiveTheoryConsciousness1993,dehaeneCognitiveNeuroscienceConsciousness2001}; and higher-order theories, which hold that consciousness depends on higher-order representation of lower-level activity \citep{carruthersHigherOrderTheoriesConsciousness2023,brownUnderstandingHigherOrderApproach2019}. 

The authors cite evidence that current DNN architectures, including Transformer-based LLMs, fail to satisfy key indicator properties of consciousness associated with these theories. For example, the global workspace theory of consciousness requires the existence of modules operating in parallel, feeding into and taking inputs from a global workspace with limited capacity to create an information bottleneck. It is tempting to view the \gls{residual-stream} of Transformer-based LLMs as loosely analogous to a global workspace that `broadcasts' information to dedicated attention modules. However, the bottleneck requirement is arguably not satisfied: the residual stream has the same dimensionality from input to output, and attention heads work with much lower-dimensional spaces. In addition, there is no genuine recurrent processing in a single forward pass through the model: an attention head in one layer can only store information for downstream layers to process, as opposed to making it globally available to all other attention heads.

This example highlights the challenge of reaching strong conclusions about the computations that current model architectures may or may not implement to vindicate claims about whether they satisfy `indicator properties' of consciousness. Attention heads in Transformers do compress information in subspaces of the residual stream that are globally available for downstream layers to read from, albeit only within the forward pass. However, some variations of autoregressive Transformers enable recurrent processing that might, in principle, satisfy at least some of the core computational requirements identified by leading theories of consciousness \citep{giannouLoopedTransformersProgrammable2023,hutchinsBlockRecurrentTransformers2022,bulatovRecurrentMemoryTransformer2022}. Ultimately, as \cite{butlinConsciousnessArtificialIntelligence2023} note, most of the indicator properties extracted form these theories could likely be implemented in AI systems using current methods and architectural tweaks, although no existing system seems a strong candidate for consciousness in this respect. As with the assessment of other psychological capacities, more work is required to bridge low-level descriptions of algorithmic circuits in LLMs and high-level descriptions of their architectural features and behavior with intermediate-level descriptions of functional building blocks.

\subsection{Secrecy and the reproducibility crisis} \label{sec:secrecy}

Much like the brains of humans and animals, LLMs today are large, opaque, and produce strikingly sophisticated behavior; this has led some researchers to suggest that we need a new `science of machine behavior' to develop appropriate instruments and methods of analysis to study them \citep{rahwanMachineBehaviour2019}. In developing these methods, we should be careful to import general lessons that were hard won in other sciences of behavior, such as human and comparative psychology. As mentioned in the companion paper, these sciences have from the beginning been fraught with biases like anthropomorphism and anthropofabulation, and as a result have hit numerous speed bumps in their history. For example, early comparative psychology relied too much on the method of anecdotes, which led to numerous dead ends and overinterpretation of behaviors which might have simply been statistical flukes \citep{thomasLloydMorganCanon1998}. This led modern comparative psychology to emphasize the need for reproducible experimental conditions on model organisms that could be raised in controlled conditions for all labs. Several areas of social, comparative, and developmental psychology are today also facing `replication crises', where well-known psychological effects and mechanisms that were thought to be well-confirmed and taught routinely in textbooks have been found to be based on statistical artifacts that were reified through questionable methodological practices, such as p-hacking or outright data fraud \citep{wigginsReplicationCrisisPsychology2019, frankCollaborativeApproachInfant2017}. This has similarly led to calls for pre-registration of all experimental protocols, training stimuli, datasets, and analysis methods to be published alongside with experimental results, so that other researchers can confirm findings on their own and scrutinize experimental materials directly to explain any discrepancies \citep{beranReplicationPreRegistrationComparative2018,frankCollaborativeApproachInfant2017}.

Unfortunately, research on LLMs seems set to repeat and even exacerbate these problematic practices from the history of psychology. Most of the highest-performing LLMs are the results of enormous investments in high-quality datasets for training, labor-intensive fine-tuning, and proprietary architecture tweaks. These datasets and training protocols are regarded as valuable trade secrets that are closely guarded by companies like Google and OpenAI who seek to control market share by outperforming their competitors. As a result, many of the most alluring behavioral results, such as those reported in \citet{bubeckSparksArtificialGeneral2023}, are essentially anecdotes about systems which are unavailable to outside researchers (especially by interventionist methods) and by design irreproducible by other research groups. For commercial reasons, the results are frequently hyped on social media and in press releases, often beyond the scientific merit of the underlying findings. Even before the advent of LLMs, methodologically-focused researchers worried about the lack of reproducibility in deep learning research \citep{hendersonDeepReinforcementLearning2018}, which has only been exacerbated in the age of LLMs. Most academic research has been conducted on models that are publicly available for academics such as BERT, which has led to the field being metonymically dubbed `BERTology' \citep{rogersPrimerBERTologyWhat2020}. Prominent researchers on LLMs have already called for disregarding analyses based on anecdotal evidence and urged for early adoption of best practices for reproducibility in architecture, data, training methods, and analyses \citep{frankLargeLanguageModels2023,sellamMultiBERTsBERTReproductions2022}. There have also been developments of large, collaborative, open-access benchmarking methods such as BIG-bench to evaluate LLMs in a more controlled and objective manner \citep{srivastavaImitationGameQuantifying2023}. At the same time, empirical results may suggest that certain discontinuous gains in performance in LLMs can only be achieved at extremes of scale and computation power \citep{kaplanScalingLawsNeural2020,weiEmergentAbilitiesLarge2022}. If explainability and safety research is to address the largest systems which are likely to be deployed by large technology companies like OpenAI, Google, and Microsoft, researchers cannot naively hope that research focused on smaller open-access models will always generalize. Future researchers will have to learn how to strike a balance between the scientific need for reproducibility and the practical need to understand the latest achievements of large, proprietary, closed research efforts.

\section{The status of LLMs as cognitive models} \label{sec:cognitive-models}

We started this two-part paper with a question: Are LLMs more than approximate lookup tables? In other words, do they merely memorize and regurgitate common patterns in their training data through content-specific transitions without inducing more sophisticated representations and computations? Our survey of behavioral and mechanistic evidence supports a broadly negative answer (with some qualifications). LLMs can and do induce complex mechanisms that enable them to perform challenging reasoning tasks better than any other domain-general computer program. As they get more capable, they tend to approach or match human performance and error patterns on many such tasks \citep{webbEmergentAnalogicalReasoning2023,dasguptaLanguageModelsShow2023,hanInductiveReasoningHumans2024,suriLargeLanguageModels2024}. This raises intriguing questions about the status of LLMs and their derivatives – including large multimodal models and modular architectures – as potential models of aspects of human cognition \citep{mcgrathHowCanDeep2023, millierePhilosophyCognitiveScienceforthcoming}. We will conclude this overview by briefly addressing these questions in light of the preceding discussion.

\subsection{The allure of `alien intelligence'}

A refrain that has recently appeared in many popular science articles, blogs, and social media posts holds that LLMs may reflect the discovery of an entirely new kind of `alien' intelligence that is fundamentally unlike our own. This may be seen as a mere statement of fact -- an optimistic appraisal that the linguistic behavior already exhibited by state-of-the-art LLMs is obviously intelligent, coupled with a pessimistic appraisal that the underlying processes producing this behavior are much at all like ours. With respect to most of the philosophical questions raised above, however, it reflects an almost complete dodge and a return to a naive behaviorism about intelligence by use of an eye-catching label. In any particular case, we need to ensure that before we accept the use of such a label as a descriptor of some scientifically-important LLM behavior that we have scientifically-respectable methods to assess that behavior along the important dimensions reviewed in Part I.

As \cite{nicklesAlienReasoningMajor2020} points out, this language first began to appear as a descriptor of deep learning performance well before the current boom in LLMs, especially regarding new machine-learning-based methods to analyze data in high-complexity sciences like biology and fundamental physics. There, the language of `alien' intelligence was conceptually tied to older philosophical dreams of rationalists like Descartes to create a science that was free from the shackles of human perspective and bias -- an objective science free from the foibles of human `powers and faculties', as Hume put it. The hope kindled by `alien intelligence' here is that by automating the process of data collection and analysis and relaxing requirements on transparency and intelligibility, ``nonhuman sensors and neutral algorithms can replace these human `powers and faculties'\,'' with alien agents that are wholly ``data-driven and increasingly are able to construct their own models and search strategies''. Indeed, there is some reason to think that breakthrough systems like AlphaFold, which have nearly solved the highly-complex protein folding problem that eluded human microbiologists for more than a century, achieved their results precisely by discovering inscrutable features that are too complex for humans to understand (Buckner 2020, Kieval 2023). Whether one buys the talk of alien intelligence and inscrutable features in this context, however, the concept derives its utility from pragmatic scientific goals that humans largely accept -- the ability to predict and perhaps control natural phenomena.

When this notion of `alien intelligence' is applied to the linguistic behavior of LLMs, however, the standards relevant to assessing its utility are far less clear. We suggested in Part I that there are many reasons to be skeptical of purely behavioral criteria for intelligence when applied to products of modern LLMs, for they may in any instance be mere Blockheads that, like a student who has merely memorized the answer key, are simply regurgitating information contained in their training sets. A purely behavioral criteria for intelligence that makes no mention of learning efficiency or representational properties should also be rejected, for it gives us no way to say that two systems that display the same behavior differ in intelligence, even when they use very different procedures to achieve the same goal. We also reviewed reasons in section 2 above to be skeptical that implicit assumptions on which benchmarks or other measures are based could be trusted, without evidence of deeper mechanistic or algorithmic correspondences with human subjects. As such, we should resist the application of `alien intelligence' to LLMs until this term can be explicated in a way that addresses these concerns and outlines a scientifically-responsible set of practices for evaluating internal performance differences amongst various LLMs.

\subsection{A plausible middle ground}

Summarizing the previous discussion leads to the following conclusions. Firstly, DNNs like Transformers have have an astounding capacity for memorization and learn from enormous training sets. As a result, any particular behavior they exhibit could be mere regurgitation or approximate retrieval, rather than demonstrating human-like processing or understanding of the task. This is what we characterized as our null hypothesis in Part I \citep{millierePhilosophicalIntroductionLanguage2024}. Secondly, however, not all LLM behaviors can be explained by this null hypothesis. Interventionist methods suggest that, at least in some cases, their behavior is caused by robust representations of task variables.

Thirdly, these models do not yet generalize as well out of distribution as humans or some classical models. Like many other ANNs trained by gradient descent, LLMs exhibit varying degrees of abstraction in their generalization behavior, depending on the inductive biases of their neural network architecture and the exploitable informational structure of their training data. They may exhibit weak systematicity or islands of strongly systematic behavior, but not fully classical systematic generalization to very different task stimuli. While some `non-content-specific' processing does occur in LLMs \citep{sheaMovingContentspecificComputation2023}, their behavior still falls short of the full consistency and generality exhibited by humans.

This limitation is likely due to the fact that most LLMs only model formal aspects of human linguistic competence \citep{mahowaldDissociatingLanguageThought2023}. Their designers make little attempt to model a human-like cognitive architecture, which might include multiple semi-independent modules corresponding to domain-general psychological faculties that enable and maintain consistency in humans – such as perception, memory, imagination, attention, social cognition, metacognition, reflection, and long-term planning. Without these additional components, LLMs remain limited in their ability to match human-level systematicity across diverse contexts. The lack of such a comprehensive agent architecture probably also partly explains why LLMs learn so much less efficiently than humans, requiring orders of magnitude more training data before achieving comparable performance.

Considering available evidence, a middle-ground position emerges regarding the relevance of LLMs to human cognition. Current-generation LLMs may serve as inefficient but useful tools for reverse-engineering partial models of the processes or algorithms that humans use to generate linguistic behavior \citep{zhouWhatAlgorithmsCan2023}. While it would be far-fetched to consider LLMs adequate computational models of the human mind, let alone human learning, they are not entirely irrelevant to human cognition and intelligence.

LLMs may capture partial models of human cognition by learning representations and computations similar to those used by humans, at least in specific domains. These learned representations and computations can be flexibly combined using operations on continuous vectors, enabling LLMs to generate novel behaviors. The resulting behaviors can reproduce not only the surface-level linguistic patterns of human language use but also, to a limited extent, the underlying systematicity and generalization ability characteristic of human cognition. It is worth noting that no previous AI technology has been able to achieve this level of fidelity to human linguistic and cognitive behavior. However, it is also important to acknowledge that LLMs often still fall short of human efficiency and flexibility when generalizing to novel stimuli. 

To be clear, even this middle-ground position remains speculative. We have reviewed evidence from intervention methods suggesting that LLMs can represent syntactic and semantic properties of their utterances, but the degree of success achieved by current interventions is difficult to interpret as definitive. Weaker evidence from interchange interventions can often be obtained for alternative representational interpretations of networks, and these various interpretations may be mutually exclusive. Simply regarding the strongest interpretation as the `correct' representational interpretation of a trained LLM requires further justification.

This issue reflects a more general fact about language and language use: philosophers of language and linguists have long noted that words and sentences in typical dialogues are inherently ambiguous and vague. It may be that only when sentences are produced and controlled by agents with stable viewpoints and communicative intentions do they acquire determinate meanings. Current generations of LLMs likely lack such agential stability, as readers can verify for themselves by observing the malleability of their responses. It is easy to prompt the current generation of chatbots to recant previously affirmed statements and defend inconsistent positions by simply suggesting that they are wrong, confused, or offensive. In short, while LLMs may accumulate information about the same syntactic and semantic properties as humans and even combine that information in flexible ways to create novel outputs, they nevertheless lack the agential stability required to imbue their utterances with determinate meanings that remain stable over time.

\section{Conclusion}

In this two-part paper, we have sought to provide a systematic overview of philosophical issues raised by the rapid progress of LLMs. In Part I, we discussed the significance of LLMs in relation to classical debates in the philosophy of artificial intelligence, mind and language. We argued that LLMs have made important headway on problems that challenged previous connectionist approaches, such as modeling compositional and systematic aspects of language use. However, we also highlighted limitations that preclude hasty comparisons between LLMs and human competence.

In Part II, we turned to newer philosophical questions raised by the current state of the art in language modeling research. We reviewed philosophically-grounded interventionist methods that aim to uncover the causal mechanisms underlying LLM performance, such as the existence of algorithmically meaningful internal representations and computations. This growing body of work suggests that state-of-the-art LLMs implement important aspects of the abstractness, systematicity, and generalizability of human cognitive processes, although they still fall short in terms of efficiency, completeness, and agency. We also discussed emerging trends in LLM research, including the development of multimodal models and the integration of LLMs into broader `agent' architectures. These approaches attempt to address some of the weaknesses of traditional LLMs by creating AI systems that can learn from self-exploration, maintain internal memories, and formulate plans based on their interactions with real or virtual environments. While it remains unclear whether current LLMs satisfy proposed computational markers of consciousness, these advancements open up the possibility of future systems exhibiting increasingly sophisticated forms of intelligence.

Finally, we proposed a nuanced perspective on the status of LLMs as partial models of human cognition. While current LLMs are far from full-fledged computational models of the human mind, carefully designed experiments informed by cognitive science and philosophy may provide genuine insights into some of the representations and algorithms that enable specific aspects of natural language use in humans. For example, training LLMs on ecologically valid data in developmentally plausible learning scenarios, rather than internet-scale corpora, could constrain hypotheses about the mechanisms underlying language acquisition and processing. However, it is crucial to stress the limitations of current LLM-based systems as models of human cognition as a whole. They lack the architectural components that enable the consistency and generality of human cognition across diverse domains. Moreover, while LLMs can flexibly recombine semantic and syntactic information to produce novel outputs, they plausibly lack the agential stability and communicative intentions to reliably anchor the meaning of their utterances. Nevertheless, the fidelity with which they can emulate specific aspects of human linguistic and cognitive behavior, when trained and evaluated using scientifically rigorous methods, is already far beyond previous AI approaches. This suggests that LLMs could serve as valuable tools for investigating targeted research questions in cognitive science, provided that their limitations are carefully considered and their performance is not overinterpreted.

Research on LLMs is a highly active and fast-paced endeavor at the intersection of artificial intelligence, cognitive science, and philosophy. Our overview suggests that this line of research is not `hitting a wall' \citep{marcusDeepLearningHitting2022}, but rather that its prospects as a tool to study distinctively human forms of cognition continue to grow. Philosophy can provide valuable conceptual clarity and theoretical guidance in interpreting the behavior and capabilities of LLMs. We hope that this two-part paper will not only bring attention to the philosophical significance of LLMs, but also convince researchers across disciplines – from computer science to cognitive science – of the value of engaging with philosophical perspectives in their work. At the same time, we have cautioned against the trend towards increasingly closed, irreproducible, and proprietary research on LLMs, which risks compromising the scientific integrity and social responsibility of this field. Sustaining genuine progress in understanding LLMs will require a commitment to open and interdisciplinary research practices. Only then can we hope to fully grasp the implications of LLMs and their successors for the long-term project of illuminating the nature of intelligence – both artificial and natural.\footnote{We are grateful to Jim Garson and Charles Rathkopf for their comments on previous drafts of this work.}

\phantomsection
\label{glossary}
\printglossaries

\bibliography{references}

@misc{ahnCanNotSay2022,
  title = {Do {{As I Can}}, {{Not As I Say}}: {{Grounding Language}} in {{Robotic Affordances}}},
  shorttitle = {Do {{As I Can}}, {{Not As I Say}}},
  author = {Ahn, Michael and Brohan, Anthony and Brown, Noah and Chebotar, Yevgen and Cortes, Omar and David, Byron and Finn, Chelsea and Fu, Chuyuan and Gopalakrishnan, Keerthana and Hausman, Karol and Herzog, Alex and Ho, Daniel and Hsu, Jasmine and Ibarz, Julian and Ichter, Brian and Irpan, Alex and Jang, Eric and Ruano, Rosario Jauregui and Jeffrey, Kyle and Jesmonth, Sally and Joshi, Nikhil J. and Julian, Ryan and Kalashnikov, Dmitry and Kuang, Yuheng and Lee, Kuang-Huei and Levine, Sergey and Lu, Yao and Luu, Linda and Parada, Carolina and Pastor, Peter and Quiambao, Jornell and Rao, Kanishka and Rettinghouse, Jarek and Reyes, Diego and Sermanet, Pierre and Sievers, Nicolas and Tan, Clayton and Toshev, Alexander and Vanhoucke, Vincent and Xia, Fei and Xiao, Ted and Xu, Peng and Xu, Sichun and Yan, Mengyuan and Zeng, Andy},
  year = {2022},
  month = aug,
  number = {arXiv:2204.01691},
  eprint = {2204.01691},
  primaryclass = {cs},
  publisher = {arXiv},
  doi = {10.48550/arXiv.2204.01691},
  urldate = {2023-04-02},
  archiveprefix = {arxiv}
}

@misc{alainUnderstandingIntermediateLayers2018,
  title = {Understanding Intermediate Layers Using Linear Classifier Probes},
  author = {Alain, Guillaume and Bengio, Yoshua},
  year = {2018},
  month = nov,
  number = {arXiv:1610.01644},
  eprint = {1610.01644},
  primaryclass = {cs, stat},
  publisher = {arXiv},
  doi = {10.48550/arXiv.1610.01644},
  urldate = {2023-08-29},
  archiveprefix = {arxiv}
}

@article{alayracFlamingoVisualLanguage2022,
  title = {Flamingo: A {{Visual Language Model}} for {{Few-Shot Learning}}},
  shorttitle = {Flamingo},
  author = {Alayrac, Jean-Baptiste and Donahue, Jeff and Luc, Pauline and Miech, Antoine and Barr, Iain and Hasson, Yana and Lenc, Karel and Mensch, Arthur and Millican, Katie and Reynolds, Malcolm and Ring, Roman and Rutherford, Eliza and Cabi, Serkan and Han, Tengda and Gong, Zhitao and Samangooei, Sina and Monteiro, Marianne and Menick, Jacob and Borgeaud, Sebastian and Brock, Andrew and Nematzadeh, Aida and Sharifzadeh, Sahand and Binkowski, Mikolaj and Barreira, Ricardo and Vinyals, Oriol and Zisserman, Andrew and Simonyan, Karen},
  year = {2022},
  month = nov,
  journal = {Advances in Neural Information Processing Systems},
  volume = {35},
  eprint = {2204.14198},
  primaryclass = {cs},
  pages = {23716--23736},
  doi = {10.48550/arXiv.2204.14198},
  urldate = {2023-03-28},
  archiveprefix = {arxiv},
  langid = {english}
}

@article{aruFeasibilityArtificialConsciousness2023,
  title = {The Feasibility of Artificial Consciousness through the Lens of Neuroscience},
  author = {Aru, Jaan and Larkum, Matthew E. and Shine, James M.},
  year = {2023},
  month = oct,
  journal = {Trends in Neurosciences},
  issn = {0166-2236},
  doi = {10.1016/j.tins.2023.09.009},
  urldate = {2023-10-19}
}

@book{baarsCognitiveTheoryConsciousness1993,
  title = {A {{Cognitive Theory}} of {{Consciousness}}},
  author = {Baars, Bernard J.},
  year = {1993},
  month = jul,
  publisher = {Cambridge University Press},
  googlebooks = {7w6IYeJRqyoC},
  isbn = {978-0-521-42743-2},
  langid = {english}
}

@article{barronWhatInsectsCan2016,
  title = {What Insects Can Tell Us about the Origins of Consciousness},
  author = {Barron, Andrew B. and Klein, Colin},
  year = {2016},
  month = may,
  journal = {Proceedings of the National Academy of Sciences},
  volume = {113},
  number = {18},
  pages = {4900--4908},
  publisher = {Proceedings of the National Academy of Sciences},
  doi = {10.1073/pnas.1520084113},
  urldate = {2023-10-20}
}

@article{barwichValueFailureScience2019,
  title = {The {{Value}} of {{Failure}} in {{Science}}: {{The Story}} of {{Grandmother Cells}} in {{Neuroscience}}},
  shorttitle = {The {{Value}} of {{Failure}} in {{Science}}},
  author = {Barwich, Ann-Sophie},
  year = {2019},
  journal = {Frontiers in Neuroscience},
  volume = {13},
  issn = {1662-453X},
  urldate = {2023-10-18}
}

@inproceedings{beckersApproximateCausalAbstractions2020,
  title = {Approximate {{Causal Abstractions}}},
  booktitle = {Proceedings of {{The}} 35th {{Uncertainty}} in {{Artificial Intelligence Conference}}},
  author = {Beckers, Sander and Eberhardt, Frederick and Halpern, Joseph Y.},
  year = {2020},
  month = aug,
  pages = {606--615},
  publisher = {PMLR},
  issn = {2640-3498},
  urldate = {2023-10-04},
  langid = {english}
}

@article{belinkovProbingClassifiersPromises2022,
  title = {Probing {{Classifiers}}: {{Promises}}, {{Shortcomings}}, and {{Advances}}},
  shorttitle = {Probing {{Classifiers}}},
  author = {Belinkov, Yonatan},
  year = {2022},
  month = apr,
  journal = {Computational Linguistics},
  volume = {48},
  number = {1},
  pages = {207--219},
  issn = {0891-2017},
  doi = {10.1162/coli_a_00422},
  urldate = {2023-08-09}
}

@article{beranReplicationPreRegistrationComparative2018,
  title = {Replication and {{Pre-Registration}} in {{Comparative Psychology}}},
  author = {Beran, Michael},
  year = {2018},
  journal = {International Journal of Comparative Psychology},
  volume = {31},
  number = {0},
  issn = {0889-3675},
  doi = {10.46867/ijcp.2018.31.01.09},
  urldate = {2023-10-18},
  langid = {english}
}

@article{brownUnderstandingHigherOrderApproach2019,
  title = {Understanding the {{Higher-Order Approach}} to {{Consciousness}}},
  author = {Brown, Richard and Lau, Hakwan and LeDoux, Joseph E.},
  year = {2019},
  month = sep,
  journal = {Trends in Cognitive Sciences},
  volume = {23},
  number = {9},
  pages = {754--768},
  publisher = {Elsevier},
  issn = {1364-6613, 1879-307X},
  doi = {10.1016/j.tics.2019.06.009},
  urldate = {2023-10-21},
  langid = {english},
  pmid = {31375408}
}

@misc{bubeckSparksArtificialGeneral2023,
  title = {Sparks of {{Artificial General Intelligence}}: {{Early}} Experiments with {{GPT-4}}},
  shorttitle = {Sparks of {{Artificial General Intelligence}}},
  author = {Bubeck, S{\'e}bastien and Chandrasekaran, Varun and Eldan, Ronen and Gehrke, Johannes and Horvitz, Eric and Kamar, Ece and Lee, Peter and Lee, Yin Tat and Li, Yuanzhi and Lundberg, Scott and Nori, Harsha and Palangi, Hamid and Ribeiro, Marco Tulio and Zhang, Yi},
  year = {2023},
  month = mar,
  number = {arXiv:2303.12712},
  eprint = {2303.12712},
  primaryclass = {cs},
  publisher = {arXiv},
  doi = {10.48550/arXiv.2303.12712},
  urldate = {2023-03-28},
  archiveprefix = {arxiv}
}

@article{bucknerDeepLearningPhilosophical2019,
  title = {Deep Learning: {{A}} Philosophical Introduction},
  shorttitle = {Deep Learning},
  author = {Buckner, Cameron},
  year = {2019},
  journal = {Philosophy Compass},
  volume = {14},
  number = {10},
  pages = {e12625},
  issn = {1747-9991},
  doi = {10.1111/phc3.12625},
  urldate = {2021-03-15},
  copyright = {{\copyright} 2019 The Author(s) Philosophy Compass {\copyright} 2019 John Wiley \& Sons Ltd},
  langid = {english}
}

@book{bucknerDeepLearningRational2023,
  title = {From {{Deep Learning}} to {{Rational Machines}}: {{What}} the {{History}} of {{Philosophy Can Teach Us}} about the {{Future}} of {{Artificial Intelligence}}},
  shorttitle = {From {{Deep Learning}} to {{Rational Machines}}},
  author = {Buckner, Cameron},
  year = {2023},
  month = dec,
  publisher = {Oxford University Press},
  address = {Oxford, New York},
  isbn = {978-0-19-765330-2}
}

@article{bulatovRecurrentMemoryTransformer2022,
  title = {Recurrent {{Memory Transformer}}},
  author = {Bulatov, Aydar and Kuratov, Yury and Burtsev, Mikhail},
  year = {2022},
  month = dec,
  journal = {Advances in Neural Information Processing Systems},
  volume = {35},
  pages = {11079--11091},
  urldate = {2023-10-21},
  langid = {english}
}

@misc{butlinConsciousnessArtificialIntelligence2023,
  title = {Consciousness in {{Artificial Intelligence}}: {{Insights}} from the {{Science}} of {{Consciousness}}},
  shorttitle = {Consciousness in {{Artificial Intelligence}}},
  author = {Butlin, Patrick and Long, Robert and Elmoznino, Eric and Bengio, Yoshua and Birch, Jonathan and Constant, Axel and Deane, George and Fleming, Stephen M. and Frith, Chris and Ji, Xu and Kanai, Ryota and Klein, Colin and Lindsay, Grace and Michel, Matthias and Mudrik, Liad and Peters, Megan A. K. and Schwitzgebel, Eric and Simon, Jonathan and VanRullen, Rufin},
  year = {2023},
  month = aug,
  number = {arXiv:2308.08708},
  eprint = {2308.08708},
  primaryclass = {cs, q-bio},
  publisher = {arXiv},
  doi = {10.48550/arXiv.2308.08708},
  urldate = {2023-10-17},
  archiveprefix = {arxiv}
}

@article{cabanacEmergenceConsciousnessPhylogeny2009,
  title = {The Emergence of Consciousness in Phylogeny},
  author = {Cabanac, Michel and Cabanac, Arnaud J. and Parent, Andr{\'e}},
  year = {2009},
  month = mar,
  journal = {Behavioural Brain Research},
  volume = {198},
  number = {2},
  pages = {267--272},
  issn = {0166-4328},
  doi = {10.1016/j.bbr.2008.11.028},
  urldate = {2023-10-20}
}

@article{caoMultipleRealizabilitySpirit2022,
  title = {Multiple Realizability and the Spirit of Functionalism},
  author = {Cao, Rosa},
  year = {2022},
  month = dec,
  journal = {Synthese},
  volume = {200},
  number = {6},
  pages = {506},
  issn = {1573-0964},
  doi = {10.1007/s11229-022-03524-1},
  urldate = {2023-10-20},
  langid = {english}
}

@article{carruthersCognitiveFunctionsLanguage2002,
  title = {The Cognitive Functions of Language},
  author = {Carruthers, Peter},
  year = {2002},
  month = dec,
  journal = {Behavioral and Brain Sciences},
  volume = {25},
  number = {6},
  pages = {657--674},
  publisher = {Cambridge University Press},
  issn = {1469-1825, 0140-525X},
  doi = {10.1017/S0140525X02000122},
  urldate = {2023-10-18},
  langid = {english}
}

@incollection{carruthersHigherOrderTheoriesConsciousness2023,
  title = {Higher-{{Order Theories}} of {{Consciousness}}},
  booktitle = {The {{Stanford Encyclopedia}} of {{Philosophy}}},
  author = {Carruthers, Peter and Gennaro, Rocco},
  editor = {Zalta, Edward N. and Nodelman, Uri},
  year = {2023},
  edition = {Fall 2023},
  publisher = {Metaphysics Research Lab, Stanford University},
  urldate = {2023-10-21}
}

@misc{caucheteuxGPT2ActivationsPredict2021,
  title = {{{GPT-2}}'s Activations Predict the Degree of Semantic Comprehension in the Human Brain},
  author = {Caucheteux, Charlotte and Gramfort, Alexandre and King, Jean-R{\'e}mi},
  year = {2021},
  month = sep,
  primaryclass = {New Results},
  pages = {2021.04.20.440622},
  publisher = {bioRxiv},
  doi = {10.1101/2021.04.20.440622},
  urldate = {2023-10-18},
  archiveprefix = {bioRxiv},
  chapter = {New Results},
  copyright = {{\copyright} 2021, Posted by Cold Spring Harbor Laboratory. This pre-print is available under a Creative Commons License (Attribution 4.0 International), CC BY 4.0, as described at http://creativecommons.org/licenses/by/4.0/},
  langid = {english}
}

@incollection{chalmersAbsentQualiaFading1995,
  title = {Absent {{Qualia}}, {{Fading Qualia}}, {{Dancing Qualia}}},
  booktitle = {Conscious {{Experience}}},
  author = {Chalmers, David J.},
  editor = {Metzinger, Thomas},
  year = {1995},
  pages = {309--328},
  publisher = {Ferdinand Schoningh}
}

@inproceedings{cheferTransformerInterpretabilityAttention2021,
  title = {Transformer {{Interpretability Beyond Attention Visualization}}},
  booktitle = {Proceedings of the {{IEEE}}/{{CVF Conference}} on {{Computer Vision}} and {{Pattern Recognition}}},
  author = {Chefer, Hila and Gur, Shir and Wolf, Lior},
  year = {2021},
  pages = {782--791},
  urldate = {2024-02-13},
  langid = {english}
}

@book{chomskyAspectsTheorySyntax1965,
  title = {Aspects of the {{Theory}} of {{Syntax}}},
  author = {Chomsky, Noam},
  year = {1965},
  publisher = {Cambridge, MA, USA: MIT Press}
}

@book{christiansenCreatingLanguageIntegrating2016,
  title = {Creating {{Language}}: {{Integrating Evolution}}, {{Acquisition}}, and {{Processing}}},
  shorttitle = {Creating {{Language}}},
  author = {Christiansen, Morten H. and Chater, Nick},
  year = {2016},
  month = mar,
  publisher = {MIT Press},
  isbn = {978-0-262-03431-9},
  langid = {english}
}

@incollection{clarkMagicWordsHow1998,
  title = {Magic {{Words}}: {{How Language Augments Human Computation}}},
  shorttitle = {Magic {{Words}}},
  booktitle = {Language and {{Thought}}: {{Interdisciplinary Themes}}},
  author = {Clark, Andy},
  editor = {Carruthers, Peter and Boucher, Jill},
  year = {1998},
  pages = {162--183},
  publisher = {Cambridge University Press}
}

@article{colasAutotelicAgentsIntrinsically2022,
  title = {Autotelic {{Agents}} with {{Intrinsically Motivated Goal-Conditioned Reinforcement Learning}}: {{A Short Survey}}},
  shorttitle = {Autotelic {{Agents}} with {{Intrinsically Motivated Goal-Conditioned Reinforcement Learning}}},
  author = {Colas, C{\'e}dric and Karch, Tristan and Sigaud, Olivier and Oudeyer, Pierre-Yves},
  year = {2022},
  month = jul,
  journal = {Journal of Artificial Intelligence Research},
  volume = {74},
  pages = {1159--1199},
  issn = {1076-9757},
  doi = {10.1613/jair.1.13554},
  urldate = {2024-04-30},
  copyright = {Copyright (c)},
  langid = {english}
}

@article{colasLanguageCognitiveTool2021,
  title = {Language as a {{Cognitive Tool}}: {{Dall-E}}, {{Humans}} and {{Vygotskian RL Agents}}},
  shorttitle = {Language as a {{Cognitive Tool}}},
  author = {Colas, C{\'e}dric and Karch, Tristan and {Moulin-Frier}, Cl{\'e}ment and Oudeyer, Pierre-Yves},
  year = {2021},
  month = mar,
  urldate = {2023-10-18},
  langid = {english}
}

@misc{conmyAutomatedCircuitDiscovery2023,
  title = {Towards {{Automated Circuit Discovery}} for {{Mechanistic Interpretability}}},
  author = {Conmy, Arthur and {Mavor-Parker}, Augustine N. and Lynch, Aengus and Heimersheim, Stefan and {Garriga-Alonso}, Adri{\`a}},
  year = {2023},
  month = jul,
  number = {arXiv:2304.14997},
  eprint = {2304.14997},
  primaryclass = {cs},
  publisher = {arXiv},
  doi = {10.48550/arXiv.2304.14997},
  urldate = {2023-09-12},
  archiveprefix = {arxiv}
}

@book{craverExplainingBrainMechanisms2007,
  title = {Explaining the {{Brain}}: {{Mechanisms}} and the {{Mosaic Unity}} of {{Neuroscience}}},
  shorttitle = {Explaining the {{Brain}}},
  author = {Craver, Carl F.},
  year = {2007},
  publisher = {Oxford University Press, Clarendon Press},
  address = {New York : Oxford University Press,}
}

@misc{dasguptaLanguageModelsShow2023,
  title = {Language Models Show Human-like Content Effects on Reasoning Tasks},
  author = {Dasgupta, Ishita and Lampinen, Andrew K. and Chan, Stephanie C. Y. and Sheahan, Hannah R. and Creswell, Antonia and Kumaran, Dharshan and McClelland, James L. and Hill, Felix},
  year = {2023},
  month = oct,
  number = {arXiv:2207.07051},
  eprint = {2207.07051},
  primaryclass = {cs},
  publisher = {arXiv},
  doi = {10.48550/arXiv.2207.07051},
  urldate = {2024-04-30},
  archiveprefix = {arxiv}
}

@misc{davisTestingGPT4Wolfram2023,
  title = {Testing {{GPT-4}} with {{Wolfram Alpha}} and {{Code Interpreter}} Plug-Ins on Math and Science Problems},
  author = {Davis, Ernest and Aaronson, Scott},
  year = {2023},
  month = aug,
  number = {arXiv:2308.05713},
  eprint = {2308.05713},
  primaryclass = {physics},
  publisher = {arXiv},
  doi = {10.48550/arXiv.2308.05713},
  urldate = {2023-10-23},
  archiveprefix = {arxiv}
}

@article{dehaeneCognitiveNeuroscienceConsciousness2001,
  title = {Towards a Cognitive Neuroscience of Consciousness: Basic Evidence and a Workspace Framework},
  shorttitle = {Towards a Cognitive Neuroscience of Consciousness},
  author = {Dehaene, Stanislas and Naccache, Lionel},
  year = {2001},
  month = apr,
  journal = {Cognition},
  series = {The {{Cognitive Neuroscience}} of {{Consciousness}}},
  volume = {79},
  number = {1},
  pages = {1--37},
  issn = {0010-0277},
  doi = {10.1016/S0010-0277(00)00123-2},
  urldate = {2017-11-20}
}

@misc{dosovitskiyImageWorth16x162021,
  title = {An {{Image}} Is {{Worth}} 16x16 {{Words}}: {{Transformers}} for {{Image Recognition}} at {{Scale}}},
  shorttitle = {An {{Image}} Is {{Worth}} 16x16 {{Words}}},
  author = {Dosovitskiy, Alexey and Beyer, Lucas and Kolesnikov, Alexander and Weissenborn, Dirk and Zhai, Xiaohua and Unterthiner, Thomas and Dehghani, Mostafa and Minderer, Matthias and Heigold, Georg and Gelly, Sylvain and Uszkoreit, Jakob and Houlsby, Neil},
  year = {2021},
  month = jun,
  number = {arXiv:2010.11929},
  eprint = {2010.11929},
  primaryclass = {cs},
  publisher = {arXiv},
  doi = {10.48550/arXiv.2010.11929},
  urldate = {2024-02-21},
  archiveprefix = {arxiv}
}

@article{dupreRealismObservationView2022,
  title = {Realism and {{Observation}}: {{The View}} from {{Generative Grammar}}},
  shorttitle = {Realism and {{Observation}}},
  author = {Dupre, Gabe},
  year = {2022},
  month = jul,
  journal = {Philosophy of Science},
  volume = {89},
  number = {3},
  pages = {565--584},
  publisher = {Cambridge University Press},
  issn = {0031-8248, 1539-767X},
  doi = {10.1017/psa.2022.2},
  urldate = {2023-07-27},
  langid = {english}
}

@article{dupreWhatCanDeep2021,
  title = {({{What}}) {{Can Deep Learning Contribute}} to {{Theoretical Linguistics}}?},
  author = {Dupre, Gabe},
  year = {2021},
  month = dec,
  journal = {Minds and Machines},
  volume = {31},
  number = {4},
  pages = {617--635},
  issn = {1572-8641},
  doi = {10.1007/s11023-021-09571-w},
  urldate = {2022-11-11},
  langid = {english}
}

@article{elhage2021mathematical,
  title = {A Mathematical Framework for Transformer Circuits},
  author = {Elhage, Nelson and Nanda, Neel and Olsson, Catherine and Henighan, Tom and Joseph, Nicholas and Mann, Ben and Askell, Amanda and Bai, Yuntao and Chen, Anna and Conerly, Tom and DasSarma, Nova and Drain, Dawn and Ganguli, Deep and {Hatfield-Dodds}, Zac and Hernandez, Danny and Jones, Andy and Kernion, Jackson and Lovitt, Liane and Ndousse, Kamal and Amodei, Dario and Brown, Tom and Clark, Jack and Kaplan, Jared and McCandlish, Sam and Olah, Chris},
  year = {2021},
  journal = {Transformer Circuits Thread}
}

@article{elhage2022superposition,
  title = {Toy Models of Superposition},
  author = {Elhage, Nelson and Hume, Tristan and Olsson, Catherine and Schiefer, Nicholas and Henighan, Tom and Kravec, Shauna and {Hatfield-Dodds}, Zac and Lasenby, Robert and Drain, Dawn and Chen, Carol and Grosse, Roger and McCandlish, Sam and Kaplan, Jared and Amodei, Dario and Wattenberg, Martin and Olah, Christopher},
  year = {2022},
  journal = {Transformer Circuits Thread}
}

@inproceedings{esserTamingTransformersHighResolution2021,
  title = {Taming {{Transformers}} for {{High-Resolution Image Synthesis}}},
  booktitle = {Proceedings of the {{IEEE}}/{{CVF Conference}} on {{Computer Vision}} and {{Pattern Recognition}}},
  author = {Esser, Patrick and Rombach, Robin and Ommer, Bjorn},
  year = {2021},
  pages = {12873--12883},
  urldate = {2023-10-18},
  langid = {english}
}

@article{firestonePerformanceVsCompetence2020,
  title = {Performance vs. Competence in Human--Machine Comparisons},
  author = {Firestone, Chaz},
  year = {2020},
  month = oct,
  journal = {Proceedings of the National Academy of Sciences},
  volume = {117},
  number = {43},
  pages = {26562--26571},
  publisher = {National Academy of Sciences},
  issn = {0027-8424, 1091-6490},
  doi = {10.1073/pnas.1905334117},
  urldate = {2021-03-15},
  chapter = {Perspective},
  copyright = {{\copyright} 2020 . https://www.pnas.org/site/aboutpnas/licenses.xhtmlPublished under the PNAS license.},
  isbn = {9781905334117},
  langid = {english},
  pmid = {33051296}
}

@article{frankCollaborativeApproachInfant2017,
  title = {A {{Collaborative Approach}} to {{Infant Research}}: {{Promoting Reproducibility}}, {{Best Practices}}, and {{Theory-Building}}},
  shorttitle = {A {{Collaborative Approach}} to {{Infant Research}}},
  author = {Frank, Michael C. and Bergelson, Elika and Bergmann, Christina and Cristia, Alejandrina and Floccia, Caroline and Gervain, Judit and Hamlin, J. Kiley and Hannon, Erin E. and Kline, Melissa and Levelt, Claartje and {Lew-Williams}, Casey and Nazzi, Thierry and Panneton, Robin and Rabagliati, Hugh and Soderstrom, Melanie and Sullivan, Jessica and Waxman, Sandra and Yurovsky, Daniel},
  year = {2017},
  journal = {Infancy},
  volume = {22},
  number = {4},
  pages = {421--435},
  issn = {1532-7078},
  doi = {10.1111/infa.12182},
  urldate = {2023-10-18},
  copyright = {Copyright {\copyright} International Congress of Infant Studies (ICIS)},
  langid = {english}
}

@misc{frankLargeLanguageModels2023,
  title = {Large Language Models as Models of Human Cognition},
  author = {Frank, Michael C.},
  year = {2023},
  month = jul,
  publisher = {PsyArXiv},
  doi = {10.31234/osf.io/wxt69},
  urldate = {2023-08-01},
  langid = {american}
}

@article{franksExplanationCognitiveSciences1995,
  title = {On {{Explanation}} in the {{Cognitive Sciences}}: {{Competence}}, {{Idealization}}, and the {{Failure}} of the {{Classical Cascade}}},
  shorttitle = {On {{Explanation}} in the {{Cognitive Sciences}}},
  author = {Franks, Bradley},
  year = {1995},
  month = dec,
  journal = {The British Journal for the Philosophy of Science},
  volume = {46},
  number = {4},
  pages = {475--502},
  publisher = {The University of Chicago Press},
  issn = {0007-0882},
  doi = {10.1093/bjps/46.4.475},
  urldate = {2024-03-06}
}

@misc{friedmanLearningTransformerPrograms2023,
  title = {Learning {{Transformer Programs}}},
  author = {Friedman, Dan and Wettig, Alexander and Chen, Danqi},
  year = {2023},
  month = oct,
  number = {arXiv:2306.01128},
  eprint = {2306.01128},
  primaryclass = {cs},
  publisher = {arXiv},
  doi = {10.48550/arXiv.2306.01128},
  urldate = {2024-02-06},
  archiveprefix = {arxiv}
}

@book{gallistelMemoryComputationalBrain2011,
  title = {Memory and the {{Computational Brain}}: {{Why Cognitive Science}} Will {{Transform Neuroscience}}},
  shorttitle = {Memory and the {{Computational Brain}}},
  author = {Gallistel, Randy C. and King, Adam Philip},
  year = {2011},
  month = sep,
  publisher = {John Wiley \& Sons},
  googlebooks = {o0jpHcgwkEoC},
  isbn = {978-1-4443-5976-3},
  langid = {english}
}

@misc{geigerCausalAbstractionFaithful2023,
  title = {Causal {{Abstraction}} for {{Faithful Model Interpretation}}},
  author = {Geiger, Atticus and Potts, Chris and Icard, Thomas},
  year = {2023},
  month = jan,
  number = {arXiv:2301.04709},
  eprint = {2301.04709},
  primaryclass = {cs},
  publisher = {arXiv},
  doi = {10.48550/arXiv.2301.04709},
  urldate = {2023-10-04},
  archiveprefix = {arxiv}
}

@inproceedings{geigerCausalAbstractionsNeural2021,
  title = {Causal {{Abstractions}} of {{Neural Networks}}},
  booktitle = {Advances in {{Neural Information Processing Systems}}},
  author = {Geiger, Atticus and Lu, Hanson and Icard, Thomas and Potts, Christopher},
  year = {2021},
  volume = {34},
  pages = {9574--9586},
  publisher = {Curran Associates, Inc.},
  urldate = {2023-10-18}
}

@misc{giannouLoopedTransformersProgrammable2023,
  title = {Looped {{Transformers}} as {{Programmable Computers}}},
  author = {Giannou, Angeliki and Rajput, Shashank and Sohn, Jy-yong and Lee, Kangwook and Lee, Jason D. and Papailiopoulos, Dimitris},
  year = {2023},
  month = jan,
  number = {arXiv:2301.13196},
  eprint = {2301.13196},
  primaryclass = {cs},
  publisher = {arXiv},
  doi = {10.48550/arXiv.2301.13196},
  urldate = {2023-10-05},
  archiveprefix = {arxiv}
}

@inproceedings{giulianelliHoodUsingDiagnostic2018,
  title = {Under the {{Hood}}: {{Using Diagnostic Classifiers}} to {{Investigate}} and {{Improve}} How {{Language Models Track Agreement Information}}},
  shorttitle = {Under the {{Hood}}},
  booktitle = {Proceedings of the 2018 {{EMNLP Workshop BlackboxNLP}}: {{Analyzing}} and {{Interpreting Neural Networks}} for {{NLP}}},
  author = {Giulianelli, Mario and Harding, Jack and Mohnert, Florian and Hupkes, Dieuwke and Zuidema, Willem},
  year = {2018},
  month = nov,
  pages = {240--248},
  publisher = {Association for Computational Linguistics},
  address = {Brussels, Belgium},
  doi = {10.18653/v1/W18-5426},
  urldate = {2023-08-10}
}

@book{godfrey-smithMetazoaAnimalLife2021,
  title = {Metazoa: {{Animal Life}} and the {{Birth}} of the {{Mind}}},
  shorttitle = {Metazoa},
  author = {{Godfrey-Smith}, Peter},
  year = {2021},
  month = oct,
  publisher = {Picador},
  googlebooks = {gp0bzgEACAAJ},
  isbn = {978-1-250-80026-8},
  langid = {english}
}

@article{godfrey-smithMindMatterMetabolism2016,
  title = {Mind, {{Matter}}, and {{Metabolism}}},
  author = {{Godfrey-Smith}, Peter},
  year = {2016},
  journal = {The Journal of Philosophy},
  volume = {113},
  number = {10},
  eprint = {48568275},
  eprinttype = {jstor},
  pages = {481--506},
  publisher = {Journal of Philosophy, Inc.},
  issn = {0022-362X},
  urldate = {2023-10-20}
}

@article{goodhart1975problems,
  title = {Problems of Monetary Management: {{The U}}.{{K}}. Experience},
  author = {Goodhart, Charles},
  year = {1975},
  journal = {Papers in Monetary Economics},
  volume = {1},
  pages = {1--20},
  publisher = {Reserve Bank of Australia},
  address = {Sydney}
}

@article{goyalInductiveBiasesDeep2022,
  title = {Inductive Biases for Deep Learning of Higher-Level Cognition},
  author = {Goyal, Anirudh and Bengio, Yoshua},
  year = {2022},
  month = oct,
  journal = {Proceedings of the Royal Society A: Mathematical, Physical and Engineering Sciences},
  volume = {478},
  number = {2266},
  pages = {20210068},
  publisher = {Royal Society},
  doi = {10.1098/rspa.2021.0068},
  urldate = {2023-10-18}
}

@misc{gozalo-brizuelaSurveyGenerativeAI2023,
  title = {A Survey of {{Generative AI Applications}}},
  author = {{Gozalo-Brizuela}, Roberto and {Garrido-Merch{\'a}n}, Eduardo C.},
  year = {2023},
  month = jun,
  number = {arXiv:2306.02781},
  eprint = {2306.02781},
  primaryclass = {cs},
  publisher = {arXiv},
  doi = {10.48550/arXiv.2306.02781},
  urldate = {2023-10-18},
  archiveprefix = {arxiv}
}

@misc{gururanganAnnotationArtifactsNatural2018,
  title = {Annotation {{Artifacts}} in {{Natural Language Inference Data}}},
  author = {Gururangan, Suchin and Swayamdipta, Swabha and Levy, Omer and Schwartz, Roy and Bowman, Samuel R. and Smith, Noah A.},
  year = {2018},
  month = apr,
  number = {arXiv:1803.02324},
  eprint = {1803.02324},
  primaryclass = {cs},
  publisher = {arXiv},
  doi = {10.48550/arXiv.1803.02324},
  urldate = {2023-10-11},
  archiveprefix = {arxiv}
}

@article{hanInductiveReasoningHumans2024,
  title = {Inductive Reasoning in Humans and Large Language Models},
  author = {Han, Simon Jerome and Ransom, Keith J. and Perfors, Andrew and Kemp, Charles},
  year = {2024},
  month = jan,
  journal = {Cognitive Systems Research},
  volume = {83},
  pages = {101155},
  issn = {1389-0417},
  doi = {10.1016/j.cogsys.2023.101155},
  urldate = {2024-04-30}
}

@misc{hardingOperationalisingRepresentationNatural2023,
  title = {Operationalising {{Representation}} in {{Natural Language Processing}}},
  author = {Harding, Jacqueline},
  year = {2023},
  month = jun,
  number = {arXiv:2306.08193},
  eprint = {2306.08193},
  primaryclass = {cs},
  publisher = {arXiv},
  doi = {10.48550/arXiv.2306.08193},
  urldate = {2023-08-10},
  archiveprefix = {arxiv}
}

@misc{hazinehLinearLatentWorld2023,
  title = {Linear {{Latent World Models}} in {{Simple Transformers}}: {{A Case Study}} on {{Othello-GPT}}},
  shorttitle = {Linear {{Latent World Models}} in {{Simple Transformers}}},
  author = {Hazineh, Dean S. and Zhang, Zechen and Chiu, Jeffery},
  year = {2023},
  month = oct,
  number = {arXiv:2310.07582},
  eprint = {2310.07582},
  primaryclass = {cs},
  publisher = {arXiv},
  doi = {10.48550/arXiv.2310.07582},
  urldate = {2023-12-11},
  archiveprefix = {arxiv}
}

@article{hendersonDeepReinforcementLearning2018,
  title = {Deep {{Reinforcement Learning That Matters}}},
  author = {Henderson, Peter and Islam, Riashat and Bachman, Philip and Pineau, Joelle and Precup, Doina and Meger, David},
  year = {2018},
  month = apr,
  journal = {Proceedings of the AAAI Conference on Artificial Intelligence},
  volume = {32},
  number = {1},
  issn = {2374-3468},
  doi = {10.1609/aaai.v32i1.11694},
  urldate = {2024-04-30},
  copyright = {Copyright (c)},
  langid = {english}
}

@article{henighan2023superposition,
  title = {Superposition, Memorization, and Double Descent},
  author = {Henighan, Tom and Carter, Shan and Hume, Tristan and Elhage, Nelson and Lasenby, Robert and Fort, Stanislav and Schiefer, Nicholas and Olah, Christopher},
  year = {2023},
  month = jan,
  journal = {Transformer Circuits Thread}
}

@inproceedings{hewittDesigningInterpretingProbes2019,
  title = {Designing and {{Interpreting Probes}} with {{Control Tasks}}},
  booktitle = {Proceedings of the 2019 {{Conference}} on {{Empirical Methods}} in {{Natural Language Processing}} and the 9th {{International Joint Conference}} on {{Natural Language Processing}} ({{EMNLP-IJCNLP}})},
  author = {Hewitt, John and Liang, Percy},
  year = {2019},
  month = nov,
  pages = {2733--2743},
  publisher = {Association for Computational Linguistics},
  address = {Hong Kong, China},
  doi = {10.18653/v1/D19-1275},
  urldate = {2023-08-05}
}

@article{hsiehSugarCrepeFixingHackable2023,
  title = {{{SugarCrepe}}: {{Fixing Hackable Benchmarks}} for {{Vision-Language Compositionality}}},
  shorttitle = {{{SugarCrepe}}},
  author = {Hsieh, Cheng-Yu and Zhang, Jieyu and Ma, Zixian and Kembhavi, Aniruddha and Krishna, Ranjay},
  year = {2023},
  month = dec,
  journal = {Advances in Neural Information Processing Systems},
  volume = {36},
  urldate = {2024-02-26},
  langid = {english}
}

@misc{huangInnerMonologueEmbodied2022,
  title = {Inner {{Monologue}}: {{Embodied Reasoning}} through {{Planning}} with {{Language Models}}},
  shorttitle = {Inner {{Monologue}}},
  author = {Huang, Wenlong and Xia, Fei and Xiao, Ted and Chan, Harris and Liang, Jacky and Florence, Pete and Zeng, Andy and Tompson, Jonathan and Mordatch, Igor and Chebotar, Yevgen and Sermanet, Pierre and Brown, Noah and Jackson, Tomas and Luu, Linda and Levine, Sergey and Hausman, Karol and Ichter, Brian},
  year = {2022},
  month = jul,
  number = {arXiv:2207.05608},
  eprint = {2207.05608},
  primaryclass = {cs},
  publisher = {arXiv},
  doi = {10.48550/arXiv.2207.05608},
  urldate = {2023-10-18},
  archiveprefix = {arxiv}
}

@article{hupkesVisualisationDiagnosticClassifiers2018,
  title = {Visualisation and '{{Diagnostic Classifiers}}' {{Reveal How Recurrent}} and {{Recursive Neural Networks Process Hierarchical Structure}}},
  author = {Hupkes, Dieuwke and Veldhoen, Sara and Zuidema, Willem},
  year = {2018},
  month = apr,
  journal = {Journal of Artificial Intelligence Research},
  volume = {61},
  pages = {907--926},
  issn = {1076-9757},
  doi = {10.1613/jair.1.11196},
  urldate = {2023-08-10},
  copyright = {Copyright (c) 2018},
  langid = {english}
}

@article{hutchinsBlockRecurrentTransformers2022,
  title = {Block-{{Recurrent Transformers}}},
  author = {Hutchins, DeLesley and Schlag, Imanol and Wu, Yuhuai and Dyer, Ethan and Neyshabur, Behnam},
  year = {2022},
  month = dec,
  journal = {Advances in Neural Information Processing Systems},
  volume = {35},
  pages = {33248--33261},
  urldate = {2023-10-21},
  langid = {english}
}

@inproceedings{icardProgramsCausalModels2017,
  title = {From Programs to Causal Models},
  booktitle = {Proceedings of the 21st {{Amsterdam}} Colloquium},
  author = {Icard, Thomas F.},
  year = {2017},
  pages = {35--44},
  urldate = {2023-10-04}
}

@article{irvineDevelopingDarkPessimism2021,
  title = {Developing {{Dark Pessimism Towards}} the {{Justificatory Role}} of {{Introspective Reports}}},
  author = {Irvine, Elizabeth},
  year = {2021},
  month = dec,
  journal = {Erkenntnis},
  volume = {86},
  number = {6},
  pages = {1319--1344},
  issn = {1572-8420},
  doi = {10.1007/s10670-019-00156-9},
  urldate = {2023-10-21},
  langid = {english}
}

@article{jonasCouldNeuroscientistUnderstand2017,
  title = {Could a {{Neuroscientist Understand}} a {{Microprocessor}}?},
  author = {Jonas, Eric and Kording, Konrad Paul},
  year = {2017},
  month = jan,
  journal = {PLOS Computational Biology},
  volume = {13},
  number = {1},
  pages = {e1005268},
  publisher = {Public Library of Science},
  issn = {1553-7358},
  doi = {10.1371/journal.pcbi.1005268},
  urldate = {2023-10-18},
  langid = {english}
}

@inproceedings{kamathTextEncodersBottleneck2023,
  title = {Text Encoders Bottleneck Compositionality in Contrastive Vision-Language Models},
  booktitle = {The 2023 {{Conference}} on {{Empirical Methods}} in {{Natural Language Processing}}},
  author = {Kamath, Amita and Hessel, Jack and Chang, Kai-Wei},
  year = {2023},
  month = dec,
  urldate = {2024-02-26},
  langid = {english}
}

@misc{kamathWhatVisionlanguageModels2023,
  title = {What's "up" with Vision-Language Models? {{Investigating}} Their Struggle with Spatial Reasoning},
  shorttitle = {What's "up" with Vision-Language Models?},
  author = {Kamath, Amita and Hessel, Jack and Chang, Kai-Wei},
  year = {2023},
  month = oct,
  number = {arXiv:2310.19785},
  eprint = {2310.19785},
  primaryclass = {cs},
  publisher = {arXiv},
  doi = {10.48550/arXiv.2310.19785},
  urldate = {2024-02-26},
  archiveprefix = {arxiv}
}

@misc{kaplanScalingLawsNeural2020,
  title = {Scaling {{Laws}} for {{Neural Language Models}}},
  author = {Kaplan, Jared and McCandlish, Sam and Henighan, Tom and Brown, Tom B. and Chess, Benjamin and Child, Rewon and Gray, Scott and Radford, Alec and Wu, Jeffrey and Amodei, Dario},
  year = {2020},
  month = jan,
  number = {arXiv:2001.08361},
  eprint = {2001.08361},
  primaryclass = {cs, stat},
  publisher = {arXiv},
  doi = {10.48550/arXiv.2001.08361},
  urldate = {2023-10-22},
  archiveprefix = {arxiv}
}

@misc{karvonenChessGPTInternalWorld2024,
  title = {Chess-{{GPT}}'s {{Internal World Model}}},
  author = {Karvonen, Adam},
  year = {2024},
  month = jan,
  journal = {Adam Karvonen's Blog},
  urldate = {2024-02-06},
  howpublished = {https://adamkarvonen.github.io/machine\_learning/2024/01/03/chess-world-models.html},
  langid = {english}
}

@misc{kielaDynabenchRethinkingBenchmarking2021,
  title = {Dynabench: {{Rethinking Benchmarking}} in {{NLP}}},
  shorttitle = {Dynabench},
  author = {Kiela, Douwe and Bartolo, Max and Nie, Yixin and Kaushik, Divyansh and Geiger, Atticus and Wu, Zhengxuan and Vidgen, Bertie and Prasad, Grusha and Singh, Amanpreet and Ringshia, Pratik and Ma, Zhiyi and Thrush, Tristan and Riedel, Sebastian and Waseem, Zeerak and Stenetorp, Pontus and Jia, Robin and Bansal, Mohit and Potts, Christopher and Williams, Adina},
  year = {2021},
  month = apr,
  number = {arXiv:2104.14337},
  eprint = {2104.14337},
  primaryclass = {cs},
  publisher = {arXiv},
  doi = {10.48550/arXiv.2104.14337},
  urldate = {2023-03-28},
  archiveprefix = {arxiv}
}

@misc{kosinskiTheoryMindMight2023,
  title = {Theory of {{Mind Might Have Spontaneously Emerged}} in {{Large Language Models}}},
  author = {Kosinski, Michal},
  year = {2023},
  month = aug,
  number = {arXiv:2302.02083},
  eprint = {2302.02083},
  primaryclass = {cs},
  publisher = {arXiv},
  doi = {10.48550/arXiv.2302.02083},
  urldate = {2023-10-11},
  archiveprefix = {arxiv}
}

@article{lammeTrueNeuralStance2006,
  title = {Towards a True Neural Stance on Consciousness},
  author = {Lamme, Victor A. F.},
  year = {2006},
  month = nov,
  journal = {Trends in Cognitive Sciences},
  volume = {10},
  number = {11},
  pages = {494--501},
  publisher = {Elsevier},
  issn = {1364-6613, 1879-307X},
  doi = {10.1016/j.tics.2006.09.001},
  urldate = {2023-10-21},
  langid = {english},
  pmid = {16997611}
}

@article{ledouxConsciousnessHumanCase2023,
  title = {Consciousness beyond the Human Case},
  author = {LeDoux, Joseph and Birch, Jonathan and Andrews, Kristin and Clayton, Nicola S. and Daw, Nathaniel D. and Frith, Chris and Lau, Hakwan and Peters, Megan A. K. and Schneider, Susan and Seth, Anil and Suddendorf, Thomas and Vandekerckhove, Marie M. P.},
  year = {2023},
  month = aug,
  journal = {Current Biology},
  volume = {33},
  number = {16},
  pages = {R832-R840},
  issn = {0960-9822},
  doi = {10.1016/j.cub.2023.06.067},
  urldate = {2023-10-17}
}

@misc{lewisDoesCLIPBind2023,
  title = {Does {{CLIP Bind Concepts}}? {{Probing Compositionality}} in {{Large Image Models}}},
  shorttitle = {Does {{CLIP Bind Concepts}}?},
  author = {Lewis, Martha and Nayak, Nihal V. and Yu, Peilin and Yu, Qinan and Merullo, Jack and Bach, Stephen H. and Pavlick, Ellie},
  year = {2023},
  month = mar,
  number = {arXiv:2212.10537},
  eprint = {2212.10537},
  primaryclass = {cs},
  publisher = {arXiv},
  doi = {10.48550/arXiv.2212.10537},
  urldate = {2024-03-23},
  archiveprefix = {arxiv}
}

@misc{liEmergentWorldRepresentations2023,
  title = {Emergent {{World Representations}}: {{Exploring}} a {{Sequence Model Trained}} on a {{Synthetic Task}}},
  shorttitle = {Emergent {{World Representations}}},
  author = {Li, Kenneth and Hopkins, Aspen K. and Bau, David and Vi{\'e}gas, Fernanda and Pfister, Hanspeter and Wattenberg, Martin},
  year = {2023},
  month = feb,
  number = {arXiv:2210.13382},
  eprint = {2210.13382},
  primaryclass = {cs},
  publisher = {arXiv},
  doi = {10.48550/arXiv.2210.13382},
  urldate = {2023-04-02},
  archiveprefix = {arxiv}
}

@inproceedings{liImplicitRepresentationsMeaning2021,
  title = {Implicit {{Representations}} of {{Meaning}} in {{Neural Language Models}}},
  booktitle = {Proceedings of the 59th {{Annual Meeting}} of the {{Association}} for {{Computational Linguistics}} and the 11th {{International Joint Conference}} on {{Natural Language Processing}} ({{Volume}} 1: {{Long Papers}})},
  author = {Li, Belinda Z. and Nye, Maxwell and Andreas, Jacob},
  year = {2021},
  month = aug,
  pages = {1813--1827},
  publisher = {Association for Computational Linguistics},
  address = {Online},
  doi = {10.18653/v1/2021.acl-long.143},
  urldate = {2022-11-11}
}

@article{lindsayTestingMethodsNeural2023,
  title = {Testing Methods of Neural Systems Understanding},
  author = {Lindsay, Grace W. and Bau, David},
  year = {2023},
  month = dec,
  journal = {Cognitive Systems Research},
  volume = {82},
  pages = {101156},
  issn = {1389-0417},
  doi = {10.1016/j.cogsys.2023.101156},
  urldate = {2023-09-21}
}

@article{liptonMythosModelInterpretability2018,
  title = {The Mythos of Model Interpretability},
  author = {Lipton, Zachary C.},
  year = {2018},
  month = sep,
  journal = {Communications of the ACM},
  volume = {61},
  number = {10},
  pages = {36--43},
  issn = {0001-0782},
  doi = {10.1145/3233231},
  urldate = {2024-01-11}
}

@incollection{lupyanChapterSevenWhat2012,
  title = {Chapter {{Seven}} - {{What Do Words Do}}? {{Toward}} a {{Theory}} of {{Language-Augmented Thought}}},
  shorttitle = {Chapter {{Seven}} - {{What Do Words Do}}?},
  booktitle = {Psychology of {{Learning}} and {{Motivation}}},
  author = {Lupyan, Gary},
  editor = {Ross, Brian H.},
  year = {2012},
  month = jan,
  series = {The {{Psychology}} of {{Learning}} and {{Motivation}}},
  volume = {57},
  pages = {255--297},
  publisher = {Academic Press},
  doi = {10.1016/B978-0-12-394293-7.00007-8},
  urldate = {2023-10-18}
}

@article{machamerThinkingMechanisms2000,
  title = {Thinking about {{Mechanisms}}},
  author = {Machamer, Peter and Darden, Lindley and Craver, Carl F.},
  year = {2000},
  month = mar,
  journal = {Philosophy of Science},
  volume = {67},
  number = {1},
  pages = {1--25},
  publisher = {Cambridge University Press},
  issn = {0031-8248, 1539-767X},
  doi = {10.1086/392759},
  urldate = {2023-09-20},
  langid = {english}
}

@misc{mahowaldDissociatingLanguageThought2023,
  title = {Dissociating Language and Thought in Large Language Models: A Cognitive Perspective},
  shorttitle = {Dissociating Language and Thought in Large Language Models},
  author = {Mahowald, Kyle and Ivanova, Anna A. and Blank, Idan A. and Kanwisher, Nancy and Tenenbaum, Joshua B. and Fedorenko, Evelina},
  year = {2023},
  month = jan,
  number = {arXiv:2301.06627},
  eprint = {2301.06627},
  primaryclass = {cs},
  publisher = {arXiv},
  doi = {10.48550/arXiv.2301.06627},
  urldate = {2023-08-01},
  archiveprefix = {arxiv}
}

@misc{manheimCategorizingVariantsGoodhart2018,
  title = {Categorizing {{Variants}} of {{Goodhart}}'s {{Law}}},
  author = {Manheim, David and Garrabrant, Scott},
  year = {2018},
  month = mar,
  journal = {arXiv.org},
  urldate = {2023-10-08},
  howpublished = {https://arxiv.org/abs/1803.04585v4},
  langid = {english}
}

@misc{marcusDeepLearningHitting2022,
  title = {Deep {{Learning Is Hitting}} a {{Wall}}},
  author = {Marcus, Gary},
  year = {2022},
  month = mar,
  journal = {Nautilus},
  urldate = {2024-05-01},
  langid = {american}
}

@book{marrVisionComputationalApproach1982,
  title = {Vision: {{A}} Computational Approach},
  shorttitle = {Vision},
  author = {Marr, David},
  year = {1982},
  publisher = {Freeman \& Co}
}

@misc{mccoyEmbersAutoregressionUnderstanding2023,
  title = {Embers of {{Autoregression}}: {{Understanding Large Language Models Through}} the {{Problem They}} Are {{Trained}} to {{Solve}}},
  shorttitle = {Embers of {{Autoregression}}},
  author = {McCoy, R. Thomas and Yao, Shunyu and Friedman, Dan and Hardy, Matthew and Griffiths, Thomas L.},
  year = {2023},
  month = sep,
  number = {arXiv:2309.13638},
  eprint = {2309.13638},
  primaryclass = {cs},
  publisher = {arXiv},
  doi = {10.48550/arXiv.2309.13638},
  urldate = {2023-09-27},
  archiveprefix = {arxiv}
}

@misc{mcgrathHowCanDeep2023,
  title = {How {{Can Deep Neural Networks Inform Theory}} in {{Psychological Science}}?},
  author = {McGrath, Sam and Russin, Jacob and Pavlick, Ellie and Feiman, Roman},
  year = {2023},
  month = nov,
  publisher = {OSF},
  doi = {10.31234/osf.io/j5ckf},
  urldate = {2024-03-27},
  archiveprefix = {OSF},
  langid = {american}
}

@misc{mengLocatingEditingFactual2023,
  title = {Locating and {{Editing Factual Associations}} in {{GPT}}},
  author = {Meng, Kevin and Bau, David and Andonian, Alex and Belinkov, Yonatan},
  year = {2023},
  month = jan,
  number = {arXiv:2202.05262},
  eprint = {2202.05262},
  primaryclass = {cs},
  publisher = {arXiv},
  doi = {10.48550/arXiv.2202.05262},
  urldate = {2023-12-11},
  archiveprefix = {arxiv}
}

@article{merkerLiabilitiesMobilitySelection2005,
  title = {The Liabilities of Mobility: {{A}} Selection Pressure for the Transition to Consciousness in Animal Evolution},
  shorttitle = {The Liabilities of Mobility},
  author = {Merker, Bjorn},
  year = {2005},
  month = mar,
  journal = {Consciousness and Cognition},
  series = {Neurobiology of {{Animal Consciousness}}},
  volume = {14},
  number = {1},
  pages = {89--114},
  issn = {1053-8100},
  doi = {10.1016/S1053-8100(03)00002-3},
  urldate = {2023-10-20}
}

@misc{meyesAblationStudiesArtificial2019,
  title = {Ablation {{Studies}} in {{Artificial Neural Networks}}},
  author = {Meyes, Richard and Lu, Melanie and {de Puiseau}, Constantin Waubert and Meisen, Tobias},
  year = {2019},
  month = feb,
  number = {arXiv:1901.08644},
  eprint = {1901.08644},
  primaryclass = {cs, q-bio},
  publisher = {arXiv},
  doi = {10.48550/arXiv.1901.08644},
  urldate = {2023-10-18},
  archiveprefix = {arxiv}
}

@incollection{milliereLanguageModelsModelsforthcoming,
  title = {Language {{Models}} as {{Models}} of {{Language}}},
  booktitle = {The {{Oxford Handbook}} of the {{Philosophy}} of {{Linguistics}}},
  author = {Milli{\`e}re, Rapha{\"e}l},
  editor = {Nefdt, Ryan and Dupre, Gabe and Stanton, Kate},
  pubstate = {forthcoming},
  publisher = {Oxford University Press},
  address = {Oxford}
}

@misc{millierePhilosophicalIntroductionLanguage2024,
  title = {A {{Philosophical Introduction}} to {{Language Models}} -- {{Part I}}: {{Continuity With Classic Debates}}},
  shorttitle = {A {{Philosophical Introduction}} to {{Language Models}} -- {{Part I}}},
  author = {Milli{\`e}re, Rapha{\"e}l and Buckner, Cameron},
  year = {2024},
  month = jan,
  number = {arXiv:2401.03910},
  eprint = {2401.03910},
  primaryclass = {cs},
  publisher = {arXiv},
  doi = {10.48550/arXiv.2401.03910},
  urldate = {2024-01-19},
  archiveprefix = {arxiv},
  copyright = {All rights reserved}
}

@article{millierePhilosophyCognitiveScienceforthcoming,
  title = {Philosophy of {{Cognitive Science}} in the {{Age}} of {{Deep Learning}}},
  author = {Milli{\`e}re, Rapha{\"e}l},
  year = {forthcoming},
  journal = {WIREs Cognitive Science}
}

@misc{mirchandaniLargeLanguageModels2023,
  title = {Large {{Language Models}} as {{General Pattern Machines}}},
  author = {Mirchandani, Suvir and Xia, Fei and Florence, Pete and Ichter, Brian and Driess, Danny and Arenas, Montserrat Gonzalez and Rao, Kanishka and Sadigh, Dorsa and Zeng, Andy},
  year = {2023},
  month = jul,
  number = {arXiv:2307.04721},
  eprint = {2307.04721},
  primaryclass = {cs},
  publisher = {arXiv},
  doi = {10.48550/arXiv.2307.04721},
  urldate = {2023-07-27},
  archiveprefix = {arxiv}
}

@misc{molloVectorGroundingProblem2023,
  title = {The {{Vector Grounding Problem}}},
  author = {Mollo, Dimitri Coelho and Milli{\`e}re, Rapha{\"e}l},
  year = {2023},
  month = apr,
  number = {arXiv:2304.01481},
  eprint = {2304.01481},
  primaryclass = {cs},
  publisher = {arXiv},
  doi = {10.48550/arXiv.2304.01481},
  urldate = {2023-08-04},
  archiveprefix = {arxiv}
}

@misc{murphyComparativeInvestigationCompositional2024,
  title = {A {{Comparative Investigation}} of {{Compositional Syntax}} and {{Semantics}} in {{DALL-E}} 2},
  author = {Murphy, Elliot and {de Villiers}, Jill and Morales, Sofia Lucero},
  year = {2024},
  month = mar,
  number = {arXiv:2403.12294},
  eprint = {2403.12294},
  primaryclass = {cs},
  doi = {10.48550/arXiv.2403.12294},
  urldate = {2024-04-30},
  archiveprefix = {arxiv}
}

@misc{nandaEmergentLinearRepresentations2023,
  title = {Emergent {{Linear Representations}} in {{World Models}} of {{Self-Supervised Sequence Models}}},
  author = {Nanda, Neel and Lee, Andrew and Wattenberg, Martin},
  year = {2023},
  month = sep,
  number = {arXiv:2309.00941},
  eprint = {2309.00941},
  primaryclass = {cs},
  publisher = {arXiv},
  doi = {10.48550/arXiv.2309.00941},
  urldate = {2024-02-14},
  archiveprefix = {arxiv}
}

@inproceedings{nandaProgressMeasuresGrokking2022,
  title = {Progress Measures for Grokking via Mechanistic Interpretability},
  booktitle = {The {{Eleventh International Conference}} on {{Learning Representations}}},
  author = {Nanda, Neel and Chan, Lawrence and Lieberum, Tom and Smith, Jess and Steinhardt, Jacob},
  year = {2022},
  month = sep,
  urldate = {2023-08-11},
  langid = {english}
}

@article{nicklesAlienReasoningMajor2020,
  title = {Alien {{Reasoning}}: {{Is}} a {{Major Change}} in {{Scientific Research Underway}}?},
  shorttitle = {Alien {{Reasoning}}},
  author = {Nickles, Thomas},
  year = {2020},
  month = sep,
  journal = {Topoi},
  volume = {39},
  number = {4},
  pages = {901--914},
  issn = {1572-8749},
  doi = {10.1007/s11245-018-9557-1},
  urldate = {2023-10-18},
  langid = {english}
}

@article{olsson2022context,
  title = {In-Context Learning and Induction Heads},
  author = {Olsson, Catherine and Elhage, Nelson and Nanda, Neel and Joseph, Nicholas and DasSarma, Nova and Henighan, Tom and Mann, Ben and Askell, Amanda and Bai, Yuntao and Chen, Anna and Conerly, Tom and Drain, Dawn and Ganguli, Deep and {Hatfield-Dodds}, Zac and Hernandez, Danny and Johnston, Scott and Jones, Andy and Kernion, Jackson and Lovitt, Liane and Ndousse, Kamal and Amodei, Dario and Brown, Tom and Clark, Jack and Kaplan, Jared and McCandlish, Sam and Olah, Chris},
  year = {2022},
  journal = {Transformer Circuits Thread}
}

@misc{openaiGPT4TechnicalReport2023,
  title = {{{GPT-4 Technical Report}}},
  author = {OpenAI},
  year = {2023},
  month = mar,
  number = {arXiv:2303.08774},
  eprint = {2303.08774},
  primaryclass = {cs},
  publisher = {arXiv},
  doi = {10.48550/arXiv.2303.08774},
  urldate = {2023-03-28},
  archiveprefix = {arxiv}
}

@misc{openaiGPT4VIsionSystem2023,
  title = {{{GPT-4V}}(Ision) {{System Card}}},
  author = {{OpenAI}},
  year = {2023},
  month = sep
}

@article{ottMappingGlobalDynamics2022,
  title = {Mapping Global Dynamics of Benchmark Creation and Saturation in Artificial Intelligence},
  author = {Ott, Simon and {Barbosa-Silva}, Adriano and Blagec, Kathrin and Brauner, Jan and Samwald, Matthias},
  year = {2022},
  month = oct,
  journal = {Nature Communications},
  volume = {13},
  number = {1},
  pages = {6793},
  publisher = {Nature Publishing Group},
  issn = {2041-1723},
  doi = {10.1038/s41467-022-34591-0},
  urldate = {2022-11-11},
  copyright = {2022 The Author(s)},
  langid = {english}
}

@misc{parkGenerativeAgentsInteractive2023,
  title = {Generative {{Agents}}: {{Interactive Simulacra}} of {{Human Behavior}}},
  shorttitle = {Generative {{Agents}}},
  author = {Park, Joon Sung and O'Brien, Joseph C. and Cai, Carrie J. and Morris, Meredith Ringel and Liang, Percy and Bernstein, Michael S.},
  year = {2023},
  month = aug,
  number = {arXiv:2304.03442},
  eprint = {2304.03442},
  primaryclass = {cs},
  publisher = {arXiv},
  doi = {10.48550/arXiv.2304.03442},
  urldate = {2023-10-18},
  archiveprefix = {arxiv}
}

@misc{parkLinearRepresentationHypothesis2023,
  title = {The {{Linear Representation Hypothesis}} and the {{Geometry}} of {{Large Language Models}}},
  author = {Park, Kiho and Choe, Yo Joong and Veitch, Victor},
  year = {2023},
  month = nov,
  number = {arXiv:2311.03658},
  eprint = {2311.03658},
  primaryclass = {cs, stat},
  publisher = {arXiv},
  doi = {10.48550/arXiv.2311.03658},
  urldate = {2023-11-12},
  archiveprefix = {arxiv}
}

@article{plautLocatingObjectKnowledge2010,
  title = {Locating Object Knowledge in the Brain: {{Comment}} on {{Bowers}}'s (2009) Attempt to Revive the Grandmother Cell Hypothesis},
  shorttitle = {Locating Object Knowledge in the Brain},
  author = {Plaut, David C. and McClelland, James L.},
  year = {2010},
  journal = {Psychological Review},
  volume = {117},
  number = {1},
  pages = {284--288},
  publisher = {American Psychological Association},
  address = {US},
  issn = {1939-1471},
  doi = {10.1037/a0017101}
}

@inproceedings{racaniereImaginationAugmentedAgentsDeep2017,
  title = {Imagination-{{Augmented Agents}} for {{Deep Reinforcement Learning}}},
  booktitle = {Advances in {{Neural Information Processing Systems}}},
  author = {Racani{\`e}re, S{\'e}bastien and Weber, Theophane and Reichert, David and Buesing, Lars and Guez, Arthur and Jimenez Rezende, Danilo and Puigdom{\`e}nech Badia, Adri{\`a} and Vinyals, Oriol and Heess, Nicolas and Li, Yujia and Pascanu, Razvan and Battaglia, Peter and Hassabis, Demis and Silver, David and Wierstra, Daan},
  year = {2017},
  volume = {30},
  publisher = {Curran Associates, Inc.},
  urldate = {2023-10-18}
}

@inproceedings{radfordLearningTransferableVisual2021,
  title = {Learning {{Transferable Visual Models From Natural Language Supervision}}},
  booktitle = {Proceedings of the 38th {{International Conference}} on {{Machine Learning}}},
  author = {Radford, Alec and Kim, Jong Wook and Hallacy, Chris and Ramesh, Aditya and Goh, Gabriel and Agarwal, Sandhini and Sastry, Girish and Askell, Amanda and Mishkin, Pamela and Clark, Jack and Krueger, Gretchen and Sutskever, Ilya},
  year = {2021},
  month = feb,
  eprint = {2103.00020},
  pages = {8748--8763},
  publisher = {PMLR},
  urldate = {2021-07-28},
  archiveprefix = {arxiv},
  langid = {english}
}

@article{rahwanMachineBehaviour2019,
  title = {Machine Behaviour},
  author = {Rahwan, Iyad and Cebrian, Manuel and Obradovich, Nick and Bongard, Josh and Bonnefon, Jean-Fran{\c c}ois and Breazeal, Cynthia and Crandall, Jacob W. and Christakis, Nicholas A. and Couzin, Iain D. and Jackson, Matthew O. and Jennings, Nicholas R. and Kamar, Ece and Kloumann, Isabel M. and Larochelle, Hugo and Lazer, David and McElreath, Richard and Mislove, Alan and Parkes, David C. and Pentland, Alex `Sandy' and Roberts, Margaret E. and Shariff, Azim and Tenenbaum, Joshua B. and Wellman, Michael},
  year = {2019},
  month = apr,
  journal = {Nature},
  volume = {568},
  number = {7753},
  pages = {477--486},
  publisher = {Nature Publishing Group},
  issn = {1476-4687},
  doi = {10.1038/s41586-019-1138-y},
  urldate = {2023-10-18},
  copyright = {2019 Springer Nature Limited},
  langid = {english}
}

@misc{rameshHierarchicalTextConditionalImage2022,
  title = {Hierarchical {{Text-Conditional Image Generation}} with {{CLIP Latents}}},
  author = {Ramesh, Aditya and Dhariwal, Prafulla and Nichol, Alex and Chu, Casey and Chen, Mark},
  year = {2022},
  month = apr,
  number = {arXiv:2204.06125},
  eprint = {2204.06125},
  primaryclass = {cs},
  institution = {arXiv},
  doi = {10.48550/arXiv.2204.06125},
  urldate = {2022-06-02},
  archiveprefix = {arxiv}
}

@inproceedings{ravfogelCounterfactualInterventionsReveal2021,
  title = {Counterfactual {{Interventions Reveal}} the {{Causal Effect}} of {{Relative Clause Representations}} on {{Agreement Prediction}}},
  booktitle = {Proceedings of the 25th {{Conference}} on {{Computational Natural Language Learning}}},
  author = {Ravfogel, Shauli and Prasad, Grusha and Linzen, Tal and Goldberg, Yoav},
  year = {2021},
  month = nov,
  pages = {194--209},
  publisher = {Association for Computational Linguistics},
  address = {Online},
  doi = {10.18653/v1/2021.conll-1.15},
  urldate = {2023-08-05}
}

@misc{ravfogelNullItOut2020,
  title = {Null {{It Out}}: {{Guarding Protected Attributes}} by {{Iterative Nullspace Projection}}},
  shorttitle = {Null {{It Out}}},
  author = {Ravfogel, Shauli and Elazar, Yanai and Gonen, Hila and Twiton, Michael and Goldberg, Yoav},
  year = {2020},
  month = apr,
  number = {arXiv:2004.07667},
  eprint = {2004.07667},
  primaryclass = {cs},
  publisher = {arXiv},
  doi = {10.48550/arXiv.2004.07667},
  urldate = {2024-02-06},
  archiveprefix = {arxiv}
}

@article{reedGeneralistAgent2022,
  title = {A {{Generalist Agent}}},
  author = {Reed, Scott and Zolna, Konrad and Parisotto, Emilio and Colmenarejo, Sergio G{\'o}mez and Novikov, Alexander and {Barth-maron}, Gabriel and Gim{\'e}nez, Mai and Sulsky, Yury and Kay, Jackie and Springenberg, Jost Tobias and Eccles, Tom and Bruce, Jake and Razavi, Ali and Edwards, Ashley and Heess, Nicolas and Chen, Yutian and Hadsell, Raia and Vinyals, Oriol and Bordbar, Mahyar and de Freitas, Nando},
  year = {2022},
  month = nov,
  journal = {Transactions on Machine Learning Research},
  eprint = {2205.06175},
  primaryclass = {cs},
  issn = {2835-8856},
  doi = {10.48550/arXiv.2205.06175},
  urldate = {2023-03-31},
  archiveprefix = {arxiv},
  langid = {english}
}

@misc{robertsDataContaminationLens2023,
  title = {Data {{Contamination Through}} the {{Lens}} of {{Time}}},
  author = {Roberts, Manley and Thakur, Himanshu and Herlihy, Christine and White, Colin and Dooley, Samuel},
  year = {2023},
  month = oct,
  number = {arXiv:2310.10628},
  eprint = {2310.10628},
  primaryclass = {cs},
  publisher = {arXiv},
  doi = {10.48550/arXiv.2310.10628},
  urldate = {2024-01-24},
  archiveprefix = {arxiv}
}

@article{rogersPrimerBERTologyWhat2020,
  title = {A {{Primer}} in {{BERTology}}: {{What We Know About How BERT Works}}},
  shorttitle = {A {{Primer}} in {{BERTology}}},
  author = {Rogers, Anna and Kovaleva, Olga and Rumshisky, Anna},
  year = {2020},
  journal = {Transactions of the Association for Computational Linguistics},
  volume = {8},
  pages = {842--866},
  publisher = {MIT Press},
  address = {Cambridge, MA},
  doi = {10.1162/tacl_a_00349},
  urldate = {2022-11-11}
}

@inproceedings{rombachHighResolutionImageSynthesis2022,
  title = {High-{{Resolution Image Synthesis}} with {{Latent Diffusion Models}}},
  booktitle = {Proceedings of the {{IEEE}}/{{CVF Conference}} on {{Computer Vision}} and {{Pattern Recognition}}},
  author = {Rombach, Robin and Blattmann, Andreas and Lorenz, Dominik and Esser, Patrick and Ommer, Bj{\"o}rn},
  year = {2022},
  month = apr,
  eprint = {2112.10752},
  primaryclass = {cs},
  pages = {10684--10695},
  publisher = {arXiv},
  doi = {10.48550/arXiv.2112.10752},
  urldate = {2022-09-09},
  archiveprefix = {arxiv},
  langid = {english}
}

@book{rumelhartParallelDistributedProcessing1987,
  title = {Parallel {{Distributed Processing}}, {{Volume}} 1: {{Explorations}} in the {{Microstructure}} of {{Cognition}}: {{Foundations}}},
  shorttitle = {Parallel {{Distributed Processing}}, {{Volume}} 1},
  author = {Rumelhart, David E. and Mcclelland, James L. and Group, PDP Research},
  year = {1987},
  month = jul,
  publisher = {MIT Press},
  googlebooks = {lOSMEAAAQBAJ},
  isbn = {978-0-262-68053-0},
  langid = {english}
}

@misc{schickToolformerLanguageModels2023,
  title = {Toolformer: {{Language Models Can Teach Themselves}} to {{Use Tools}}},
  shorttitle = {Toolformer},
  author = {Schick, Timo and {Dwivedi-Yu}, Jane and Dess{\`i}, Roberto and Raileanu, Roberta and Lomeli, Maria and Zettlemoyer, Luke and Cancedda, Nicola and Scialom, Thomas},
  year = {2023},
  month = feb,
  number = {arXiv:2302.04761},
  eprint = {2302.04761},
  primaryclass = {cs},
  publisher = {arXiv},
  doi = {10.48550/arXiv.2302.04761},
  urldate = {2023-10-17},
  archiveprefix = {arxiv}
}

@article{schrimpfNeuralArchitectureLanguage2021,
  title = {The Neural Architecture of Language: {{Integrative}} Modeling Converges on Predictive Processing},
  shorttitle = {The Neural Architecture of Language},
  author = {Schrimpf, Martin and Blank, Idan Asher and Tuckute, Greta and Kauf, Carina and Hosseini, Eghbal A. and Kanwisher, Nancy and Tenenbaum, Joshua B. and Fedorenko, Evelina},
  year = {2021},
  month = nov,
  journal = {Proceedings of the National Academy of Sciences},
  volume = {118},
  number = {45},
  pages = {e2105646118},
  publisher = {Proceedings of the National Academy of Sciences},
  doi = {10.1073/pnas.2105646118},
  urldate = {2023-10-18}
}

@article{sejnowskiParallelNetworksThat1987,
  title = {Parallel {{Networks}} That {{Learn}} to {{Pronounce English Text}}},
  author = {Sejnowski, T. J. and Rosenberg, C. R.},
  year = {1987},
  journal = {Complex System},
  volume = {1},
  pages = {145--168},
  urldate = {2023-10-18}
}

@misc{sellamMultiBERTsBERTReproductions2022,
  title = {The {{MultiBERTs}}: {{BERT Reproductions}} for {{Robustness Analysis}}},
  shorttitle = {The {{MultiBERTs}}},
  author = {Sellam, Thibault and Yadlowsky, Steve and Wei, Jason and Saphra, Naomi and D'Amour, Alexander and Linzen, Tal and Bastings, Jasmijn and Turc, Iulia and Eisenstein, Jacob and Das, Dipanjan and Tenney, Ian and Pavlick, Ellie},
  year = {2022},
  month = mar,
  number = {arXiv:2106.16163},
  eprint = {2106.16163},
  primaryclass = {cs},
  publisher = {arXiv},
  doi = {10.48550/arXiv.2106.16163},
  urldate = {2024-04-30},
  archiveprefix = {arxiv}
}

@article{sethCriteriaConsciousnessHumans2005,
  title = {Criteria for Consciousness in Humans and Other Mammals},
  author = {Seth, Anil K. and Baars, Bernard J. and Edelman, David B.},
  year = {2005},
  month = mar,
  journal = {Consciousness and Cognition},
  series = {Neurobiology of {{Animal Consciousness}}},
  volume = {14},
  number = {1},
  pages = {119--139},
  issn = {1053-8100},
  doi = {10.1016/j.concog.2004.08.006},
  urldate = {2023-10-20}
}

@article{sheaMovingContentspecificComputation2023,
  title = {Moving beyond Content-Specific Computation in Artificial Neural Networks},
  author = {Shea, Nicholas},
  year = {2023},
  journal = {Mind \& Language},
  volume = {38},
  number = {1},
  pages = {156--177},
  issn = {1468-0017},
  doi = {10.1111/mila.12387},
  urldate = {2023-10-05},
  copyright = {{\copyright} 2021 The Author. Mind \& Language published by John Wiley \& Sons Ltd.},
  langid = {english}
}

@misc{shinnReflexionLanguageAgents2023,
  title = {Reflexion: {{Language Agents}} with {{Verbal Reinforcement Learning}}},
  shorttitle = {Reflexion},
  author = {Shinn, Noah and Cassano, Federico and Berman, Edward and Gopinath, Ashwin and Narasimhan, Karthik and Yao, Shunyu},
  year = {2023},
  month = oct,
  number = {arXiv:2303.11366},
  eprint = {2303.11366},
  primaryclass = {cs},
  publisher = {arXiv},
  doi = {10.48550/arXiv.2303.11366},
  urldate = {2023-10-23},
  archiveprefix = {arxiv}
}

@incollection{smolenskyNeuralConceptualInterpretation1986,
  title = {Neural and Conceptual Interpretation of {{PDP}} Models},
  booktitle = {Parallel Distributed Processing: Explorations in the Microstructure, Vol. 2: Psychological and Biological Models},
  author = {Smolensky, P.},
  year = {1986},
  month = jan,
  pages = {390--431},
  publisher = {MIT Press},
  address = {Cambridge, MA, USA},
  urldate = {2024-02-13},
  isbn = {978-0-262-63110-5}
}

@article{smolenskyProperTreatmentConnectionism1988,
  title = {On the Proper Treatment of Connectionism},
  author = {Smolensky, Paul},
  year = {1988},
  month = mar,
  journal = {Behavioral and Brain Sciences},
  volume = {11},
  number = {1},
  pages = {1--23},
  publisher = {Cambridge University Press},
  issn = {1469-1825, 0140-525X},
  doi = {10.1017/S0140525X00052432},
  urldate = {2023-08-21},
  langid = {english}
}

@article{srivastavaImitationGameQuantifying2023,
  title = {Beyond the {{Imitation Game}}: {{Quantifying}} and Extrapolating the Capabilities of Language Models},
  shorttitle = {Beyond the {{Imitation Game}}},
  author = {Srivastava, Aarohi and Rastogi, Abhinav and Rao, Abhishek and Shoeb, Abu Awal Md and Abid, Abubakar and Fisch, Adam and Brown, Adam R. and Santoro, Adam and Gupta, Aditya and {Garriga-Alonso}, Adri{\`a} and Kluska, Agnieszka and Lewkowycz, Aitor and Agarwal, Akshat and Power, Alethea and Ray, Alex and Warstadt, Alex and Kocurek, Alexander W. and Safaya, Ali and Tazarv, Ali and Xiang, Alice and Parrish, Alicia and Nie, Allen and Hussain, Aman and Askell, Amanda and Dsouza, Amanda and Slone, Ambrose and Rahane, Ameet and Iyer, Anantharaman S. and Andreassen, Anders Johan and Madotto, Andrea and Santilli, Andrea and Stuhlm{\"u}ller, Andreas and Dai, Andrew M. and La, Andrew and Lampinen, Andrew and Zou, Andy and Jiang, Angela and Chen, Angelica and Vuong, Anh and Gupta, Animesh and Gottardi, Anna and Norelli, Antonio and Venkatesh, Anu and Gholamidavoodi, Arash and Tabassum, Arfa and Menezes, Arul and Kirubarajan, Arun and Mullokandov, Asher and Sabharwal, Ashish and Herrick, Austin and Efrat, Avia and Erdem, Aykut and Karaka{\c s}, Ayla and Roberts, B. Ryan and Loe, Bao Sheng and Zoph, Barret and Bojanowski, Bart{\l}omiej and {\"O}zyurt, Batuhan and Hedayatnia, Behnam and Neyshabur, Behnam and Inden, Benjamin and Stein, Benno and Ekmekci, Berk and Lin, Bill Yuchen and Howald, Blake and Orinion, Bryan and Diao, Cameron and Dour, Cameron and Stinson, Catherine and Argueta, Cedrick and Ferri, Cesar and Singh, Chandan and Rathkopf, Charles and Meng, Chenlin and Baral, Chitta and Wu, Chiyu and {Callison-Burch}, Chris and Waites, Christopher and Voigt, Christian and Manning, Christopher D. and Potts, Christopher and Ramirez, Cindy and Rivera, Clara E. and Siro, Clemencia and Raffel, Colin and Ashcraft, Courtney and Garbacea, Cristina and Sileo, Damien and Garrette, Dan and Hendrycks, Dan and Kilman, Dan and Roth, Dan and Freeman, C. Daniel and Khashabi, Daniel and Levy, Daniel and Gonz{\'a}lez, Daniel Mosegu{\'i} and Perszyk, Danielle and Hernandez, Danny and Chen, Danqi and Ippolito, Daphne and Gilboa, Dar and Dohan, David and Drakard, David and Jurgens, David and Datta, Debajyoti and Ganguli, Deep and Emelin, Denis and Kleyko, Denis and Yuret, Deniz and Chen, Derek and Tam, Derek and Hupkes, Dieuwke and Misra, Diganta and Buzan, Dilyar and Mollo, Dimitri Coelho and Yang, Diyi and Lee, Dong-Ho and Schrader, Dylan and Shutova, Ekaterina and Cubuk, Ekin Dogus and Segal, Elad and Hagerman, Eleanor and Barnes, Elizabeth and Donoway, Elizabeth and Pavlick, Ellie and Rodol{\`a}, Emanuele and Lam, Emma and Chu, Eric and Tang, Eric and Erdem, Erkut and Chang, Ernie and Chi, Ethan A. and Dyer, Ethan and Jerzak, Ethan and Kim, Ethan and Manyasi, Eunice Engefu and Zheltonozhskii, Evgenii and Xia, Fanyue and Siar, Fatemeh and {Mart{\'i}nez-Plumed}, Fernando and Happ{\'e}, Francesca and Chollet, Francois and Rong, Frieda and Mishra, Gaurav and Winata, Genta Indra and de Melo, Gerard and Kruszewski, Germ{\'a}n and Parascandolo, Giambattista and Mariani, Giorgio and Wang, Gloria Xinyue and {Jaimovitch-Lopez}, Gonzalo and Betz, Gregor and {Gur-Ari}, Guy and Galijasevic, Hana and Kim, Hannah and Rashkin, Hannah and Hajishirzi, Hannaneh and Mehta, Harsh and Bogar, Hayden and Shevlin, Henry Francis Anthony and Schuetze, Hinrich and Yakura, Hiromu and Zhang, Hongming and Wong, Hugh Mee and Ng, Ian and Noble, Isaac and Jumelet, Jaap and Geissinger, Jack and Kernion, Jackson and Hilton, Jacob and Lee, Jaehoon and Fisac, Jaime Fern{\'a}ndez and Simon, James B. and Koppel, James and Zheng, James and Zou, James and Kocon, Jan and Thompson, Jana and Wingfield, Janelle and Kaplan, Jared and Radom, Jarema and {Sohl-Dickstein}, Jascha and Phang, Jason and Wei, Jason and Yosinski, Jason and Novikova, Jekaterina and Bosscher, Jelle and Marsh, Jennifer and Kim, Jeremy and Taal, Jeroen and Engel, Jesse and Alabi, Jesujoba and Xu, Jiacheng and Song, Jiaming and Tang, Jillian and Waweru, Joan and Burden, John and Miller, John and Balis, John U. and Batchelder, Jonathan and Berant, Jonathan and Frohberg, J{\"o}rg and Rozen, Jos and {Hernandez-Orallo}, Jose and Boudeman, Joseph and Guerr, Joseph and Jones, Joseph and Tenenbaum, Joshua B. and Rule, Joshua S. and Chua, Joyce and Kanclerz, Kamil and Livescu, Karen and Krauth, Karl and Gopalakrishnan, Karthik and Ignatyeva, Katerina and Markert, Katja and Dhole, Kaustubh and Gimpel, Kevin and Omondi, Kevin and Mathewson, Kory Wallace and Chiafullo, Kristen and Shkaruta, Ksenia and Shridhar, Kumar and McDonell, Kyle and Richardson, Kyle and Reynolds, Laria and Gao, Leo and Zhang, Li and Dugan, Liam and Qin, Lianhui and {Contreras-Ochando}, Lidia and Morency, Louis-Philippe and Moschella, Luca and Lam, Lucas and Noble, Lucy and Schmidt, Ludwig and He, Luheng and {Oliveros-Col{\'o}n}, Luis and Metz, Luke and Senel, L{\"u}tfi Kerem and Bosma, Maarten and Sap, Maarten and Hoeve, Maartje Ter and Farooqi, Maheen and Faruqui, Manaal and Mazeika, Mantas and Baturan, Marco and Marelli, Marco and Maru, Marco and {Ramirez-Quintana}, Maria Jose and Tolkiehn, Marie and Giulianelli, Mario and Lewis, Martha and Potthast, Martin and Leavitt, Matthew L. and Hagen, Matthias and Schubert, M{\'a}ty{\'a}s and Baitemirova, Medina Orduna and Arnaud, Melody and McElrath, Melvin and Yee, Michael Andrew and Cohen, Michael and Gu, Michael and Ivanitskiy, Michael and Starritt, Michael and Strube, Michael and Sw{\k e}drowski, Micha{\l} and Bevilacqua, Michele and Yasunaga, Michihiro and Kale, Mihir and Cain, Mike and Xu, Mimee and Suzgun, Mirac and Walker, Mitch and Tiwari, Mo and Bansal, Mohit and Aminnaseri, Moin and Geva, Mor and Gheini, Mozhdeh and T, Mukund Varma and Peng, Nanyun and Chi, Nathan Andrew and Lee, Nayeon and Krakover, Neta Gur-Ari and Cameron, Nicholas and Roberts, Nicholas and Doiron, Nick and Martinez, Nicole and Nangia, Nikita and Deckers, Niklas and Muennighoff, Niklas and Keskar, Nitish Shirish and Iyer, Niveditha S. and Constant, Noah and Fiedel, Noah and Wen, Nuan and Zhang, Oliver and Agha, Omar and Elbaghdadi, Omar and Levy, Omer and Evans, Owain and Casares, Pablo Antonio Moreno and Doshi, Parth and Fung, Pascale and Liang, Paul Pu and Vicol, Paul and Alipoormolabashi, Pegah and Liao, Peiyuan and Liang, Percy and Chang, Peter W. and Eckersley, Peter and Htut, Phu Mon and Hwang, Pinyu and Mi{\l}kowski, Piotr and Patil, Piyush and Pezeshkpour, Pouya and Oli, Priti and Mei, Qiaozhu and Lyu, Qing and Chen, Qinlang and Banjade, Rabin and Rudolph, Rachel Etta and Gabriel, Raefer and Habacker, Rahel and Risco, Ramon and Milli{\`e}re, Rapha{\"e}l and Garg, Rhythm and Barnes, Richard and Saurous, Rif A. and Arakawa, Riku and Raymaekers, Robbe and Frank, Robert and Sikand, Rohan and Novak, Roman and Sitelew, Roman and Bras, Ronan Le and Liu, Rosanne and Jacobs, Rowan and Zhang, Rui and Salakhutdinov, Russ and Chi, Ryan Andrew and Lee, Seungjae Ryan and Stovall, Ryan and Teehan, Ryan and Yang, Rylan and Singh, Sahib and Mohammad, Saif M. and Anand, Sajant and Dillavou, Sam and Shleifer, Sam and Wiseman, Sam and Gruetter, Samuel and Bowman, Samuel R. and Schoenholz, Samuel Stern and Han, Sanghyun and Kwatra, Sanjeev and Rous, Sarah A. and Ghazarian, Sarik and Ghosh, Sayan and Casey, Sean and Bischoff, Sebastian and Gehrmann, Sebastian and Schuster, Sebastian and Sadeghi, Sepideh and Hamdan, Shadi and Zhou, Sharon and Srivastava, Shashank and Shi, Sherry and Singh, Shikhar and Asaadi, Shima and Gu, Shixiang Shane and Pachchigar, Shubh and Toshniwal, Shubham and Upadhyay, Shyam and Debnath, Shyamolima Shammie and Shakeri, Siamak and Thormeyer, Simon and Melzi, Simone and Reddy, Siva and Makini, Sneha Priscilla and Lee, Soo-Hwan and Torene, Spencer and Hatwar, Sriharsha and Dehaene, Stanislas and Divic, Stefan and Ermon, Stefano and Biderman, Stella and Lin, Stephanie and Prasad, Stephen and Piantadosi, Steven and Shieber, Stuart and Misherghi, Summer and Kiritchenko, Svetlana and Mishra, Swaroop and Linzen, Tal and Schuster, Tal and Li, Tao and Yu, Tao and Ali, Tariq and Hashimoto, Tatsunori and Wu, Te-Lin and Desbordes, Th{\'e}o and Rothschild, Theodore and Phan, Thomas and Wang, Tianle and Nkinyili, Tiberius and Schick, Timo and Kornev, Timofei and Tunduny, Titus and Gerstenberg, Tobias and Chang, Trenton and Neeraj, Trishala and Khot, Tushar and Shultz, Tyler and Shaham, Uri and Misra, Vedant and Demberg, Vera and Nyamai, Victoria and Raunak, Vikas and Ramasesh, Vinay Venkatesh and Prabhu, Vinay Uday and Padmakumar, Vishakh and Srikumar, Vivek and Fedus, William and Saunders, William and Zhang, William and Vossen, Wout and Ren, Xiang and Tong, Xiaoyu and Zhao, Xinran and Wu, Xinyi and Shen, Xudong and Yaghoobzadeh, Yadollah and Lakretz, Yair and Song, Yangqiu and Bahri, Yasaman and Choi, Yejin and Yang, Yichi and Hao, Yiding and Chen, Yifu and Belinkov, Yonatan and Hou, Yu and Hou, Yufang and Bai, Yuntao and Seid, Zachary and Zhao, Zhuoye and Wang, Zijian and Wang, Zijie J. and Wang, Zirui and Wu, Ziyi},
  year = {2023},
  month = jan,
  journal = {Transactions on Machine Learning Research},
  issn = {2835-8856},
  doi = {10.48550/arXiv.2206.04615},
  urldate = {2023-08-09},
  copyright = {All rights reserved},
  langid = {english}
}

@article{suriLargeLanguageModels2024,
  title = {Do Large Language Models Show Decision Heuristics Similar to Humans? {{A}} Case Study Using {{GPT-3}}.5.},
  shorttitle = {Do Large Language Models Show Decision Heuristics Similar to Humans?},
  author = {Suri, Gaurav and Slater, Lily R. and Ziaee, Ali and Nguyen, Morgan},
  year = {2024},
  journal = {Journal of Experimental Psychology: General},
  volume = {153},
  number = {4},
  pages = {1066--1075},
  publisher = {American Psychological Association},
  address = {US},
  issn = {1939-2222},
  doi = {10.1037/xge0001547}
}

@misc{syedAttributionPatchingOutperforms2023,
  title = {Attribution {{Patching Outperforms Automated Circuit Discovery}}},
  author = {Syed, Aaquib and Rager, Can and Conmy, Arthur},
  year = {2023},
  month = oct,
  number = {arXiv:2310.10348},
  eprint = {2310.10348},
  primaryclass = {cs},
  publisher = {arXiv},
  doi = {10.48550/arXiv.2310.10348},
  urldate = {2023-10-18},
  archiveprefix = {arxiv}
}

@article{theakstonRolePerformanceLimitations2001,
  title = {The Role of Performance Limitations in the Acquisition of Verb-Argument Structure: An Alternative Account},
  shorttitle = {The Role of Performance Limitations in the Acquisition of Verb-Argument Structure},
  author = {Theakston, Anna L. and Lieven, Elena V. M. and Pine, Julian M. and Rowland, Caroline F.},
  year = {2001},
  month = feb,
  journal = {Journal of Child Language},
  volume = {28},
  number = {1},
  pages = {127--152},
  issn = {1469-7602, 0305-0009},
  doi = {10.1017/S0305000900004608},
  urldate = {2024-04-30},
  langid = {english}
}

@incollection{thomasLloydMorganCanon1998,
  title = {Lloyd {{Morgan}}'s {{Canon}}},
  booktitle = {Comparative Psychology: {{A}} Handbook},
  author = {Thomas, Roger K.},
  editor = {Greenberg, G. and Haraway, M. M.},
  year = {1998},
  volume = {894},
  pages = {156--163},
  publisher = {Garland Publishing Co},
  address = {New York},
  urldate = {2023-10-18}
}

@book{tomaselloConstructingLanguage2009,
  title = {Constructing a {{Language}}},
  author = {Tomasello, Michael},
  year = {2009},
  month = jun,
  publisher = {Harvard University Press},
  googlebooks = {7M\_lSEfzTQoC},
  isbn = {978-0-674-04439-5},
  langid = {english}
}

@article{tongMassProducingFailuresMultimodal2023,
  title = {Mass-{{Producing Failures}} of {{Multimodal Systems}} with {{Language Models}}},
  author = {Tong, Shengbang and Jones, Erik and Steinhardt, Jacob},
  year = {2023},
  month = dec,
  journal = {Advances in Neural Information Processing Systems},
  volume = {36},
  pages = {29292--29322},
  urldate = {2024-03-23},
  langid = {english}
}

@article{tschannenImageCaptionersAre2023,
  title = {Image {{Captioners Are Scalable Vision Learners Too}}},
  author = {Tschannen, Michael and Kumar, Manoj and Steiner, Andreas and Zhai, Xiaohua and Houlsby, Neil and Beyer, Lucas},
  year = {2023},
  month = dec,
  journal = {Advances in Neural Information Processing Systems},
  volume = {36},
  urldate = {2024-02-26},
  langid = {english}
}

@misc{ullmanLargeLanguageModels2023,
  title = {Large {{Language Models Fail}} on {{Trivial Alterations}} to {{Theory-of-Mind Tasks}}},
  author = {Ullman, Tomer},
  year = {2023},
  month = mar,
  number = {arXiv:2302.08399},
  eprint = {2302.08399},
  primaryclass = {cs},
  publisher = {arXiv},
  doi = {10.48550/arXiv.2302.08399},
  urldate = {2023-10-11},
  archiveprefix = {arxiv}
}

@article{vygotskyThinkingSpeech1987,
  title = {Thinking and {{Speech}}},
  author = {Vygotsky, L. S.},
  year = {1987},
  journal = {The Collected Works of L. S. Vygotsky},
  volume = {1},
  pages = {39--285},
  publisher = {Plenum},
  urldate = {2023-10-18}
}

@misc{wangSurveyLargeLanguage2023,
  title = {A {{Survey}} on {{Large Language Model}} Based {{Autonomous Agents}}},
  author = {Wang, Lei and Ma, Chen and Feng, Xueyang and Zhang, Zeyu and Yang, Hao and Zhang, Jingsen and Chen, Zhiyuan and Tang, Jiakai and Chen, Xu and Lin, Yankai and Zhao, Wayne Xin and Wei, Zhewei and Wen, Ji-Rong},
  year = {2023},
  month = sep,
  number = {arXiv:2308.11432},
  eprint = {2308.11432},
  primaryclass = {cs},
  publisher = {arXiv},
  doi = {10.48550/arXiv.2308.11432},
  urldate = {2023-10-18},
  archiveprefix = {arxiv}
}

@misc{wangVoyagerOpenEndedEmbodied2023,
  title = {Voyager: {{An Open-Ended Embodied Agent}} with {{Large Language Models}}},
  shorttitle = {Voyager},
  author = {Wang, Guanzhi and Xie, Yuqi and Jiang, Yunfan and Mandlekar, Ajay and Xiao, Chaowei and Zhu, Yuke and Fan, Linxi and Anandkumar, Anima},
  year = {2023},
  month = oct,
  number = {arXiv:2305.16291},
  eprint = {2305.16291},
  primaryclass = {cs},
  publisher = {arXiv},
  doi = {10.48550/arXiv.2305.16291},
  urldate = {2023-10-23},
  archiveprefix = {arxiv}
}

@article{webbEmergentAnalogicalReasoning2023,
  title = {Emergent Analogical Reasoning in Large Language Models},
  author = {Webb, Taylor and Holyoak, Keith J. and Lu, Hongjing},
  year = {2023},
  month = jul,
  journal = {Nature Human Behaviour},
  pages = {1--16},
  publisher = {Nature Publishing Group},
  issn = {2397-3374},
  doi = {10.1038/s41562-023-01659-w},
  urldate = {2023-08-29},
  copyright = {2023 The Author(s), under exclusive licence to Springer Nature Limited},
  langid = {english}
}

@misc{weiEmergentAbilitiesLarge2022,
  title = {Emergent {{Abilities}} of {{Large Language Models}}},
  author = {Wei, Jason and Tay, Yi and Bommasani, Rishi and Raffel, Colin and Zoph, Barret and Borgeaud, Sebastian and Yogatama, Dani and Bosma, Maarten and Zhou, Denny and Metzler, Donald and Chi, Ed H. and Hashimoto, Tatsunori and Vinyals, Oriol and Liang, Percy and Dean, Jeff and Fedus, William},
  year = {2022},
  month = oct,
  number = {arXiv:2206.07682},
  eprint = {2206.07682},
  primaryclass = {cs},
  publisher = {arXiv},
  doi = {10.48550/arXiv.2206.07682},
  urldate = {2023-10-05},
  archiveprefix = {arxiv}
}

@inproceedings{weissThinkingTransformers2021,
  title = {Thinking {{Like Transformers}}},
  booktitle = {Proceedings of the 38th {{International Conference}} on {{Machine Learning}}},
  author = {Weiss, Gail and Goldberg, Yoav and Yahav, Eran},
  year = {2021},
  month = jul,
  pages = {11080--11090},
  publisher = {PMLR},
  issn = {2640-3498},
  urldate = {2023-10-05},
  langid = {english}
}

@misc{whittingtonRelatingTransformersModels2022,
  title = {Relating Transformers to Models and Neural Representations of the Hippocampal Formation},
  author = {Whittington, James C. R. and Warren, Joseph and Behrens, Timothy E. J.},
  year = {2022},
  month = mar,
  number = {arXiv:2112.04035},
  eprint = {2112.04035},
  primaryclass = {cs, q-bio},
  publisher = {arXiv},
  doi = {10.48550/arXiv.2112.04035},
  urldate = {2023-10-18},
  archiveprefix = {arxiv}
}

@article{wigginsReplicationCrisisPsychology2019,
  title = {The Replication Crisis in Psychology: {{An}} Overview for Theoretical and Philosophical Psychology},
  shorttitle = {The Replication Crisis in Psychology},
  author = {Wiggins, Bradford J. and Christopherson, Cody D.},
  year = {2019},
  journal = {Journal of Theoretical and Philosophical Psychology},
  volume = {39},
  number = {4},
  pages = {202--217},
  publisher = {Educational Publishing Foundation},
  address = {US},
  issn = {2151-3341},
  doi = {10.1037/teo0000137}
}

@book{woodwardMakingThingsHappen2005,
  title = {Making {{Things Happen}}: {{A Theory}} of {{Causal Explanation}}},
  shorttitle = {Making {{Things Happen}}},
  author = {Woodward, James},
  year = {2005},
  month = oct,
  publisher = {Oxford University Press, USA},
  googlebooks = {IFVJAbgySmEC},
  isbn = {978-0-19-518953-7},
  langid = {english}
}

@misc{wuEarlyEvaluationGPT4V2023,
  title = {An {{Early Evaluation}} of {{GPT-4V}}(Ision)},
  author = {Wu, Yang and Wang, Shilong and Yang, Hao and Zheng, Tian and Zhang, Hongbo and Zhao, Yanyan and Qin, Bing},
  year = {2023},
  month = oct,
  number = {arXiv:2310.16534},
  eprint = {2310.16534},
  primaryclass = {cs},
  publisher = {arXiv},
  doi = {10.48550/arXiv.2310.16534},
  urldate = {2024-02-20},
  archiveprefix = {arxiv}
}

@misc{wuInterpretabilityScaleIdentifying2023,
  title = {Interpretability at {{Scale}}: {{Identifying Causal Mechanisms}} in {{Alpaca}}},
  shorttitle = {Interpretability at {{Scale}}},
  author = {Wu, Zhengxuan and Geiger, Atticus and Potts, Christopher and Goodman, Noah D.},
  year = {2023},
  month = may,
  number = {arXiv:2305.08809},
  eprint = {2305.08809},
  primaryclass = {cs},
  publisher = {arXiv},
  doi = {10.48550/arXiv.2305.08809},
  urldate = {2023-09-12},
  archiveprefix = {arxiv}
}

@misc{yangDawnLMMsPreliminary2023,
  title = {The {{Dawn}} of {{LMMs}}: {{Preliminary Explorations}} with {{GPT-4V}}(Ision)},
  shorttitle = {The {{Dawn}} of {{LMMs}}},
  author = {Yang, Zhengyuan and Li, Linjie and Lin, Kevin and Wang, Jianfeng and Lin, Chung-Ching and Liu, Zicheng and Wang, Lijuan},
  year = {2023},
  month = oct,
  number = {arXiv:2309.17421},
  eprint = {2309.17421},
  primaryclass = {cs},
  publisher = {arXiv},
  doi = {10.48550/arXiv.2309.17421},
  urldate = {2024-02-20},
  archiveprefix = {arxiv}
}

@misc{yangRethinkingBenchmarkContamination2023,
  title = {Rethinking {{Benchmark}} and {{Contamination}} for {{Language Models}} with {{Rephrased Samples}}},
  author = {Yang, Shuo and Chiang, Wei-Lin and Zheng, Lianmin and Gonzalez, Joseph E. and Stoica, Ion},
  year = {2023},
  month = nov,
  number = {arXiv:2311.04850},
  eprint = {2311.04850},
  primaryclass = {cs},
  publisher = {arXiv},
  doi = {10.48550/arXiv.2311.04850},
  urldate = {2024-01-24},
  archiveprefix = {arxiv}
}

@misc{yaoReActSynergizingReasoning2023,
  title = {{{ReAct}}: {{Synergizing Reasoning}} and {{Acting}} in {{Language Models}}},
  shorttitle = {{{ReAct}}},
  author = {Yao, Shunyu and Zhao, Jeffrey and Yu, Dian and Du, Nan and Shafran, Izhak and Narasimhan, Karthik and Cao, Yuan},
  year = {2023},
  month = mar,
  number = {arXiv:2210.03629},
  eprint = {2210.03629},
  primaryclass = {cs},
  publisher = {arXiv},
  doi = {10.48550/arXiv.2210.03629},
  urldate = {2023-10-23},
  archiveprefix = {arxiv}
}

@article{yildirimTaskStructuresWorld2023,
  title = {From Task Structures to World Models: {{What}} Do {{LLMs}} Know?},
  shorttitle = {From Task Structures to World Models},
  author = {Yildirim, Ilker and Paul, L. A.},
  year = {2023},
  month = oct,
  journal = {Trends in Cognitive Sciences},
  eprint = {2310.04276},
  primaryclass = {cs, q-bio},
  doi = {10.48550/arXiv.2310.04276},
  urldate = {2023-10-16},
  archiveprefix = {arxiv}
}

@inproceedings{yuksekgonulWhenWhyVisionLanguage2022,
  title = {When and {{Why Vision-Language Models Behave}} like {{Bags-Of-Words}}, and {{What}} to {{Do About It}}?},
  booktitle = {The {{Eleventh International Conference}} on {{Learning Representations}}},
  author = {Yuksekgonul, Mert and Bianchi, Federico and Kalluri, Pratyusha and Jurafsky, Dan and Zou, James},
  year = {2022},
  month = sep,
  urldate = {2024-04-30},
  langid = {english}
}

@misc{zengSocraticModelsComposing2022,
  title = {Socratic {{Models}}: {{Composing Zero-Shot Multimodal Reasoning}} with {{Language}}},
  shorttitle = {Socratic {{Models}}},
  author = {Zeng, Andy and Attarian, Maria and Ichter, Brian and Choromanski, Krzysztof and Wong, Adrian and Welker, Stefan and Tombari, Federico and Purohit, Aveek and Ryoo, Michael and Sindhwani, Vikas and Lee, Johnny and Vanhoucke, Vincent and Florence, Pete},
  year = {2022},
  month = may,
  number = {arXiv:2204.00598},
  eprint = {2204.00598},
  primaryclass = {cs},
  publisher = {arXiv},
  doi = {10.48550/arXiv.2204.00598},
  urldate = {2023-12-19},
  archiveprefix = {arxiv}
}

@misc{zhangBestPracticesActivation2023,
  title = {Towards {{Best Practices}} of {{Activation Patching}} in {{Language Models}}: {{Metrics}} and {{Methods}}},
  shorttitle = {Towards {{Best Practices}} of {{Activation Patching}} in {{Language Models}}},
  author = {Zhang, Fred and Nanda, Neel},
  year = {2023},
  month = sep,
  number = {arXiv:2309.16042},
  eprint = {2309.16042},
  primaryclass = {cs},
  publisher = {arXiv},
  doi = {10.48550/arXiv.2309.16042},
  urldate = {2023-09-29},
  archiveprefix = {arxiv}
}

@inproceedings{zhangEvaluatingCLIPUnderstanding2023,
  title = {Evaluating {{CLIP}}'s {{Understanding}} on {{Relationships}} in a {{Blocks World}}},
  booktitle = {2023 {{IEEE International Conference}} on {{Big Data}} ({{BigData}})},
  author = {Zhang, Kairui and Lewis, Martha},
  year = {2023},
  month = dec,
  pages = {2257--2264},
  doi = {10.1109/BigData59044.2023.10386915},
  urldate = {2024-03-23}
}

@misc{zhouSolvingChallengingMath2023,
  title = {Solving {{Challenging Math Word Problems Using GPT-4 Code Interpreter}} with {{Code-based Self-Verification}}},
  author = {Zhou, Aojun and Wang, Ke and Lu, Zimu and Shi, Weikang and Luo, Sichun and Qin, Zipeng and Lu, Shaoqing and Jia, Anya and Song, Linqi and Zhan, Mingjie and Li, Hongsheng},
  year = {2023},
  month = aug,
  number = {arXiv:2308.07921},
  eprint = {2308.07921},
  primaryclass = {cs},
  publisher = {arXiv},
  doi = {10.48550/arXiv.2308.07921},
  urldate = {2023-10-23},
  archiveprefix = {arxiv}
}

@misc{zhouWhatAlgorithmsCan2023,
  title = {What {{Algorithms}} Can {{Transformers Learn}}? {{A Study}} in {{Length Generalization}}},
  shorttitle = {What {{Algorithms}} Can {{Transformers Learn}}?},
  author = {Zhou, Hattie and Bradley, Arwen and Littwin, Etai and Razin, Noam and Saremi, Omid and Susskind, Josh and Bengio, Samy and Nakkiran, Preetum},
  year = {2023},
  month = oct,
  number = {arXiv:2310.16028},
  eprint = {2310.16028},
  primaryclass = {cs, stat},
  publisher = {arXiv},
  doi = {10.48550/arXiv.2310.16028},
  urldate = {2023-11-28},
  archiveprefix = {arxiv}
}

@inproceedings{zitkovichRT2VisionLanguageActionModels2023,
  title = {{{RT-2}}: {{Vision-Language-Action Models Transfer Web Knowledge}} to {{Robotic Control}}},
  shorttitle = {{{RT-2}}},
  booktitle = {7th {{Annual Conference}} on {{Robot Learning}}},
  author = {Zitkovich, Brianna and Yu, Tianhe and Xu, Sichun and Xu, Peng and Xiao, Ted and Xia, Fei and Wu, Jialin and Wohlhart, Paul and Welker, Stefan and Wahid, Ayzaan and Vuong, Quan and Vanhoucke, Vincent and Tran, Huong and Soricut, Radu and Singh, Anikait and Singh, Jaspiar and Sermanet, Pierre and Sanketi, Pannag R. and Salazar, Grecia and Ryoo, Michael S. and Reymann, Krista and Rao, Kanishka and Pertsch, Karl and Mordatch, Igor and Michalewski, Henryk and Lu, Yao and Levine, Sergey and Lee, Lisa and Lee, Tsang-Wei Edward and Leal, Isabel and Kuang, Yuheng and Kalashnikov, Dmitry and Julian, Ryan and Joshi, Nikhil J. and Irpan, Alex and Ichter, Brian and Hsu, Jasmine and Herzog, Alexander and Hausman, Karol and Gopalakrishnan, Keerthana and Fu, Chuyuan and Florence, Pete and Finn, Chelsea and Dubey, Kumar Avinava and Driess, Danny and Ding, Tianli and Choromanski, Krzysztof Marcin and Chen, Xi and Chebotar, Yevgen and Carbajal, Justice and Brown, Noah and Brohan, Anthony and Arenas, Montserrat Gonzalez and Han, Kehang},
  year = {2023},
  month = aug,
  urldate = {2023-10-23},
  langid = {english}
}

\end{document}